# A Unified Framework for Comprehensive Cardiac CT Segmentation and Phenotyping: Human-in-the-Loop Data Annotation, Vision Foundation Model Development, Multicenter Evaluation and Clinical Validation

Pooya Mohammadi Kazaj[1,2,3], Leo Fridolin Weber[1,2†], Wen Xie[1,2†], Seyed Amir Ahmad Safavi-Naini[†1,2], Anselm Stark[1,2], Giovanni Baj[1,2], Ali Mokhtari[1,2], Toshiya Yoshida[4], Christoph Ryffel[1,2], Taishi Okuno[4], Yoshihiro Akashi[4], Ronny R Buechel[5], Thomas Pilgrim[1], Waldo Valenzuela[6], George CM Siontis[1], Xiaowei Xu[7], Moritz Hundertmark[1], Stephan Windecker[1], Christoph Gräni[1,2*], Isaac Shiri[1,2*]

1-Department of Cardiology, Inselspital, Bern University Hospital, University of Bern, Bern, Switzerland

2-Department of Digital Medicine, University of Bern, Bern, Switzerland

3-Graduate School for Cellular and Biomedical Sciences, University of Bern, Bern, Switzerland

4-Department of Cardiology, St. Marianna University School of Medicine, 2-16-1, Sugao, Miyamae-ku, Kawasaki, 216-8511, Japan

5-Department of Nuclear Medicine, Cardiac Imaging, University Hospital Zurich, Zurich, Switzerland

6- Institute for Diagnostic and Interventional Neuroradiology, Inselspital Bern, Bern, Switzerland

7-Guangdong Provincial People's Hospital, Guangzhou, China

[†] Leo Fridolin Weber, Wen Xie, Seyed Amir Ahmad Safavi-Naini contributed equally to this work.

[*] Dr. Gräni and Dr. Shiri jointly supervised this work.

*** Corresponding Author:**
Isaac Shiri, PhD
Department of Cardiology
Inselspital, Bern University Hospital
University of Bern,
Freiburgstrasse CH - 3010 Bern, Switzerland
Email: isaac.shirilord@unibe.ch

## Abstract

Comprehensive quantification of cardiac structures from computed tomography (CT) remains limited not by data availability but by the scalability of measurements, which makes routine use impractical. Here we present a unified framework for comprehensive cardiac CT segmentation and phenotyping that combines a human-in-the-loop annotation pipeline, a cardiac CT augmentation technique, and a self-supervised foundation model pre-trained on 60,000 unlabeled cardiac CT scans. Using this approach, we assembled the largest and most comprehensive expert-annotated cardiac CT segmentation dataset to date, comprising 1598 cases and 14 distinct cardiac structures (1000 for training, 598 for the external test set). Across five external datasets, the framework segmented all structures more accurately and comprehensively than existing open-source tools. Self-supervised pre-training improved labeling efficiency, with the most significant gains observed during external evaluation in the low-data regime. Benchmarking across convolutional, transformer, and state-space architectures showed comparable performance, indicating that data quality and pre-training, rather than architecture, drove accuracy. The framework was scaled to population-level phenotyping, with segmented anatomy that carries functionally relevant information about ventricular function and disease severity beyond demographic variables. By openly releasing the largest dataset with human labels, code, model weights, a CT augmentation library, and software, this work provides a reproducible foundation for opportunistic cardiac phenotyping from routinely acquired CT scans.

**Keywords**

Cardiac, Segmentation, Computed Tomography, Foundation Model, Human-in-the-Loop

# 1. Introduction

Computed tomography (CT) provides high-resolution detailed anatomical imaging of the human body, and cardiac CT (CCT) acquisitions are specifically optimized to capture cardiac structure and function throughout the cardiac cycle. As the first-line test for numerous cardiovascular indications, CCT is now performed more than 1 million times annually in the United States[1] alone, spanning coronary artery disease, valvular heart disease, and congenital heart disease, as well as pre-procedural planning and post-procedural follow-up of surgical and transcatheter heart interventions[2]. However, a single CCT study contains far more information than is typically extracted during standard clinical interpretation for a predefined indication. The same CCT acquisition that addresses a specific clinical question simultaneously captures all cardiac chambers, left and right ventricular mass, both atria, heart valves, the great vessels, coronary arteries, pericardial and epicardial fat, implanted cardiac devices as well as indirect indicators of active pathophysiology[3].

Despite the breadth of available data, routine workflows quantify only a limited subset of these structures in these scans. Manual and semi-automated measurements are time-intensive, demand specialized expertise, and are subject to intra-observer, inter-observer, and inter-method variability[4,5]. The manual quantification of every structure in every scan is therefore impractical on a clinical scale, and workflow scalability becomes the main limitation rather than data acquisition. Beyond dedicated CCT studies, tens of millions of chest and abdominal CT scans are performed annually for non-cardiac indications, each incidentally capturing the heart yet rarely analyzed for cardiac-specific information[6,7]. Overcoming this limitation may enable opportunistic, comprehensive, and reproducible cardiac phenotyping from scans already performed in routine care, supporting individualized cardiovascular care.

Early artificial intelligence (AI) tools for medical imaging were built for single, narrowly defined tasks and generalized poorly beyond the institution and population where they were developed[8]. To overcome these limitations, generalist or foundation models were developed to accommodate multiple tasks, disease entities, and imaging phenotypes within a single framework. CT-specific segmentation tools[9–11] and general medical imaging foundation models[12–14], offer broad coverage but lack cardiac-specific depth. Developing automated multi-structure analysis of CCT depends on several elements that rarely coexist: data that adequately represent the relevant cardiac conditions, high-quality expert annotations for supervised training, and a learning approach that generalizes beyond narrow datasets[15]. Expert-annotated CCT datasets are scarce because delineating numerous distinct structures in each case requires significant manual effort and specialized expertise[16]. Self-supervised learning could potentially ease this bottleneck by leveraging the large volumes of unlabeled CCT already acquired in clinical practice, requiring only a limited number of expert labels for downstream tasks[17–19]. A human-in-the-loop process complements this process by enabling experts to correct model-generated segmentations rather than annotate from scratch and to focus the effort on the most informative cases, increasing the volume of labeled data while reducing manual effort[20]. However, existing public datasets remain limited in scale and cardiac anatomical coverage, with none offering comprehensive multi-structure cardiac segmentation. Addressing this limitation requires a data-centric strategy that assembles large-scale samples and enables external validation across diverse populations, demonstrating clinical generalizability in real-world settings rather than under laboratory conditions alone[21].

Here we present a unified framework for comprehensive CCT segmentation and quantification that enables phenotyping at both the patient and population level. Using a human-in-the-loop pipeline, we curated expert-corrected segmentations of multiple cardiac structures across a large training cohort and five external test datasets. To improve robustness on different scans and conditions, we developed CTAug, a CCT augmentation library. To improve label efficiency, we introduce CCT foundation model (CCT-FM), a foundation model pre-trained in a self-supervised manner on large-scale unlabeled CCT volumes. For model evaluation, we tested the models on five external datasets, benchmarking different architectures (convolutional, transformer, and Mamba) and comparing them against existing open-source CCT segmentation models on the same datasets. For clinical validation, we use additional independent datasets in which CT-derived ejection fraction from 4D CCT and myocardial mass are compared with echocardiographic measurements for agreement. We deploy the framework at population scale across multiple tasks to assess the added value of cardiac substructures and to characterize variation across clinical parameters. CCT-FM improves segmentation performance across diverse labels and external validation, particularly in low-data regimes. Our framework outperforms all existing open-source algorithms across all shared structures and provides the most comprehensive segmentation model for CCT to date. The benchmark shows that different architectures yield comparable performance. We release the largest labeled and most comprehensive CCT segmentation dataset, along with model weights, open-source code, and software for CCT augmentation, segmentation, and quantification.

# 2. Results

## *2.1. Study Overview*

**Figure 1** provides the complete overview of the study pipeline, dataset sources, human-in-the-loop diagram, and evaluation framework. **Table 1** summarizes all dataset characteristics, with no overlap between the pre-training, training, and external test cohorts. To develop, test, and validate a comprehensive CCT segmentation framework, we combined several complementary components.

We first established a data-centric, human-in-the-loop labeling pipeline to create six high-quality reference datasets of CCT, one for training (1000 patients) and five for external testing (598 patients), with expert-validated segmentations across 14 cardiac structures. To improve robustness during training, we developed CTAug, a CCT augmentation library that simulates different characteristic imaging artifacts. Next, we performed self-supervised pre-training on 60,331 unlabeled CCT volumes followed by supervised fine-tuning to develop CCT-FM, which we evaluated on downstream tasks including CCT segmentation, coronary anomaly segmentation, and non-contrast CT domain adaptation.

To contextualize performance, we conducted a systematic benchmark of model architectures (convolutional, transformer, and Mamba U-Net[22] designs) using identical data splits and assessed generalization across five independent external test datasets. We further compared three open-source models with cardiac segmentation capabilities (TotalSegmentator[9], Atlas[10], and MOOSE[11]) against our trained models across the same five external cohorts. Finally, we deployed a final model for clinical evaluation, encompassing agreement with transthoracic echocardiography and population-scale cardiac phenotyping using 3D and 4D phase-resolved protocols.

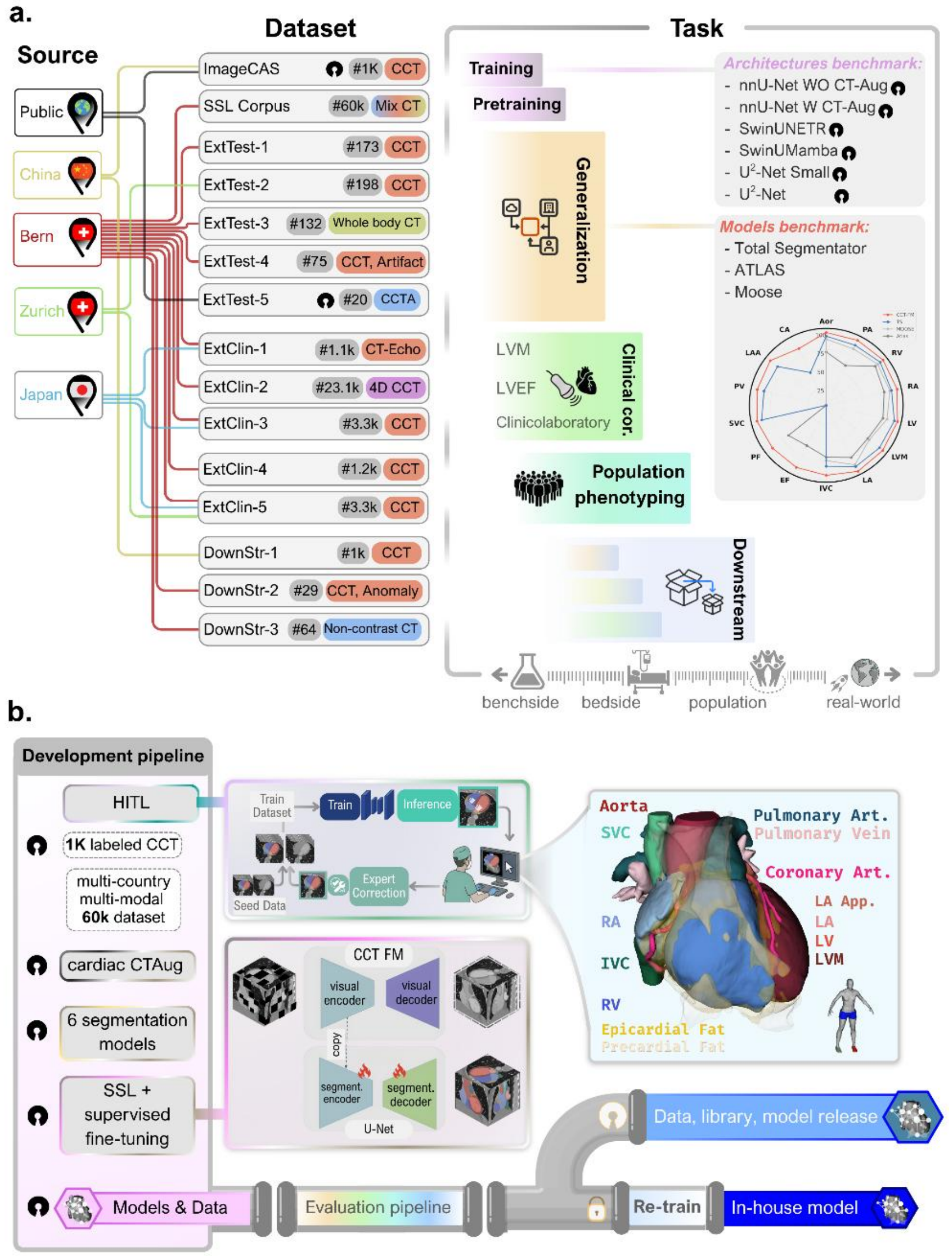

**Figure 1. CCT-FM development and evaluation pipeline. (a)** Dataset and task overview. The left columns map each dataset to its geographic source: the labeled training set, the self-supervised pre training corpus (SSL Corpus), five external test sets (ExtTest-1 to -5), five clinical correlation cohorts (ExtClin-1 to -5), and three downstream task datasets (DownStr-1 to -3), each annotated with case count and modality. The right column maps these to tasks spanning the translational spectrum from benchside to real-world: architecture benchmarking, open access model benchmarking, generalization, clinical correlation, and population phenotyping. **(b)** Development pipeline. A human-in-the-loop process trains and refines segmentation through iterative training, inference, and expert correction; self-supervised pre training of a masked autoencoder, followed by supervised fine-tuning on 1K labeled CCT with CTAug within a U-Net, yields CCT-FM. The three-dimensional rendering shows the 14 segmented cardiac structures color-coded by anatomical category. Models and datasets feed an evaluation pipeline, after which the datasets, libraries, and models are publicly released and an in-house model is retrained to support clinical correlation and population phenotyping. Datasets released as open access and open-sourced models are marked with an open-source icon. Structure abbreviations: Aor, aorta; CA/Coronary Art, coronary arteries; EF, epicardial fat; IVC, inferior vena cava; LA, left atrium; LAA/LA App, left atrial appendage; LV, left ventricle; LVM, left ventricular myocardium; PA, pulmonary arteries; PF, pericardial fat; PV, pulmonary vein; RA, right atrium; RV, right ventricle; SVC, superior vena cava.

**Table 1.** Details of datasets used for model development, validation, and population-scale analysis. HITL: human-in-the-loop labeling

| Dataset | Role | Source | Patients | Modality | Labels | Key Characteristics |
|---|---|---|---|---|---|---|
| Train & DownStr-1 | Training & Transfer learning | Open-Access China | 1,000 | CCT | 14-structure HITL-corrected | Multi-center, corrected via iterative HITL |
| FM-Pretrain | Pre Training Foundation Model | Bern + open Access | ~60,331 | Mixed CT (cardiac) | None (unlabeled) | Multi-protocol, multi-indication |
| ExtTest-1 | External test | Inselspital Bern | 173 | CCT | 14-structure expert | Standard diagnostic CCT |
| ExtTest-2 | External test | Zurich | 198 | CCT | 14-structure expert | Independent institution |
| ExtTest-3 | External test | Inselspital Bern | 132 | Whole-body CT | 14-structure expert | Pre-procedural imaging |
| ExtTest-4 | External test | Inselspital Bern | 75 | CCT (artifact) | 14-structure expert | Metal implants, motion degradation |
| ExtTest-5 | External test | Open-Access | 20 | CCT | 14-structure expert | Whole heart CT images |
| ExtClin-1 | Clinical validation | Inselspital Bern & Japan | 953 | CCT + paired echo | Echo-derived LVM | Paired CT and echo |
| ExtClin-2 | Clinical validation | Inselspital Bern | 1091 | 4D CCT + paired echo | Echo-derived LVEF | Paired CT and echo |
| ExtClin-3 | Clinical validation | Inselspital Bern & Japan | 1962 | CCT + demographics + clinical+ laboratory | CT-Pred Information Gain | CT volume information gain with demographics |
| ExtClin-4 | Population analysis | Inselspital Bern | 1,245 4D CT (23,142 3D CT) | Multi-phase CCT | - | ~21 cardiac phases per patient |
| ExtClin-5 | Population analysis | Inselspital Bern + Zurich + Japan | 3307 | CCT | - | Three countries, cross-sectional |
| DownStr-2 | Transfer learning | Inselspital Bern | ~29 | CCT | anomaly masks | Rare pathology, few-shot |
| DownStr-3 | Domain transfer | Inselspital Bern | 64 | Non-contrast (PCCT) | Automated | Non-contrast Photon-counting CT |

## *2.2. Data Augmentation*

Across all five external test datasets, training with CTAug increased the mean Dice score from 95.5 to 96.3 (Wilcoxon signed-rank test, all $p < 0.001$; **Supplementary Figure S1**). This benefit was most pronounced in the artifact-rich cohort (ExtTest-4, 94.81 vs 92.45), where baseline Dice was lowest and most variable. CTAug raised the lower tail of the distribution and narrowed its spread, indicating fewer low-Dice outliers on the most challenging cases.

## 2.3. Self-Supervised Pre-Training and Data Efficiency

Self-supervised pre-training on unlabeled CCT scans substantially improved label efficiency during downstream fine-tuning. Pooled across all five external test datasets, CCT-FM-pretrained models outperformed the same architecture trained from scratch at every training-set size, and the advantage was largest when labeled data were scarcest, adding 4.66 Dice points at the 1% fraction (82.18 versus 77.52) and narrowing to 0.82 points at 100% (**Figure 2a**). This pattern held in each cohort individually (Figure 2b-f), with the largest low-data gain in the artifact-rich ExtTest-4, where pre-training improved Dice by 13.35 points at the 5% fraction. The benefit extended to two downstream tasks, non-contrast photon-counting CT segmentation and coronary anomaly segmentation, where pretrained models achieved higher mean Dice across all training-set sizes, and the gap widened as cases became fewer, reaching 11.52 points with only 2 coronary training cases (**Figure 2g-h**). Sex-stratified analyses showed the same trend in female and male subgroups (**Supplementary Figure S2** and **Supplementary Figure S3**, respectively).

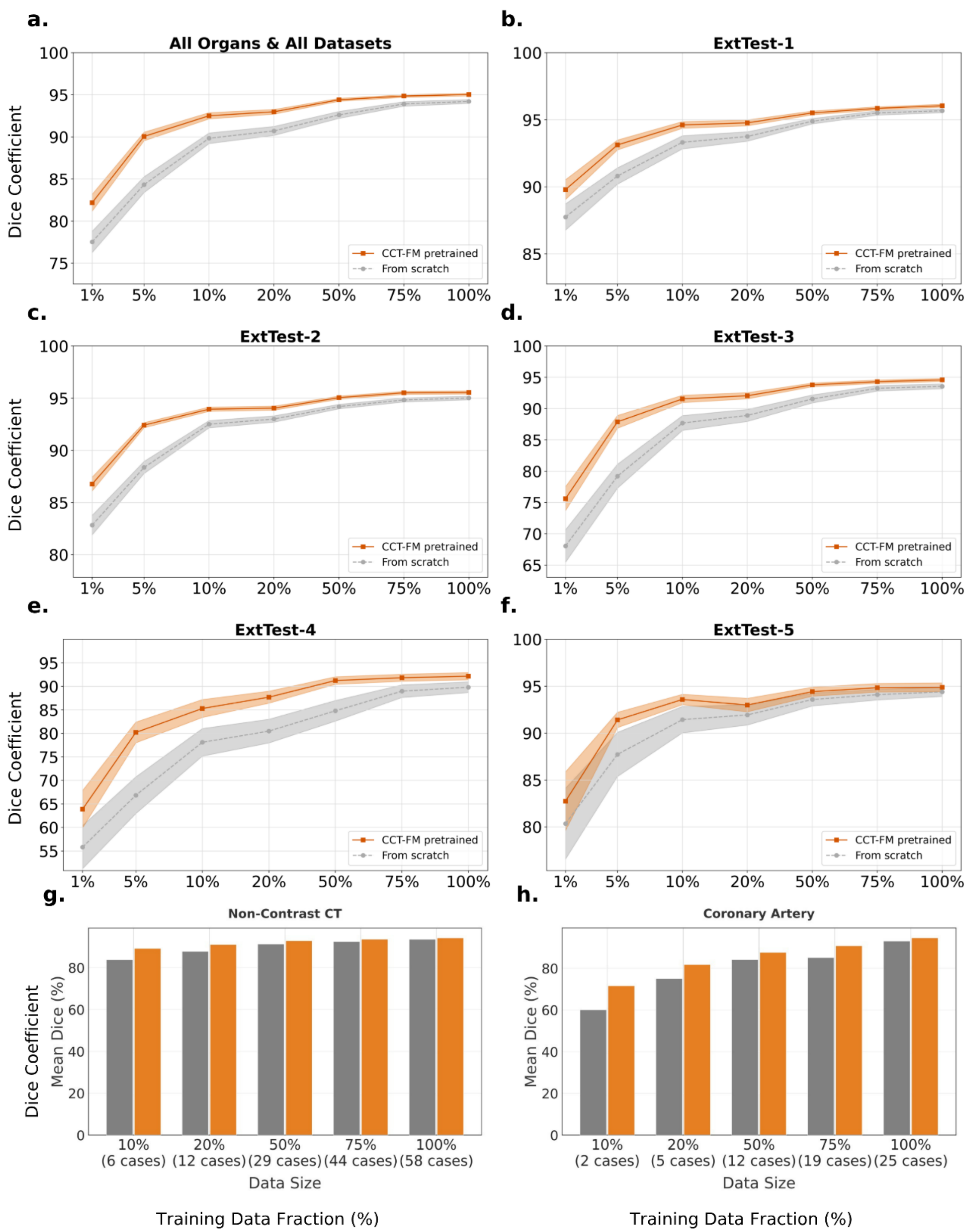

**Figure 2. Data efficiency of self-supervised pre-training across datasets and downstream tasks.** Mean Dice coefficient as a function of labeled training data fraction (1–100%) for models initialized from CCT-FM pretrained weights (orange) versus trained from scratch (gray), aggregated across datasets. Shaded bands denote 95% confidence intervals. (a) All organs across all datasets micro-average (N = 598) and (b–f) each external test dataset individually (ExtTest-1: N = 173; ExtTest-2: N = 198; ExtTest-3: N = 132; ExtTest-4: N = 75; ExtTest-5: N = 20). (g–h) Mean Dice on the downstream non-contrast CT and coronary anomaly tasks across increasing training-set sizes. Pre-training delivers its greatest benefit at low data fractions, enabling pretrained models to reach high segmentation accuracy with markedly fewer labeled cases.

At the structure level, the same pattern held for all fourteen cardiac structures, but its magnitude tracked how difficult each structure was to segment from scratch (**Figure 3**). Gains concentrated in thin, tubular, and low-contrast structures, where from-scratch models were weakest when labels were scarce. The most striking case was the pulmonary arteries, where pre-training raised Dice from 62.42 to 80.16 at the 1% fraction, a 17.74-point gain that was 18.37 points at 5% (**Figure 3k**). Similar low-data improvements appeared for the aorta, inferior vena cava, pulmonary vein, and left atrium, with the left atrium rising by 10.03 points at 1% (**Figure 3e, h, j, l**). By contrast, structures that from-scratch models already segmented well gained little: the left ventricle and the pericardial and epicardial fat compartments showed nearly overlapping curves at every fraction (**Figure 3b, n, o**). The coronary arteries remained the hardest target overall, with the lowest ceiling (83.76 at full data), yet pre-training preserved a small consistent edge there as well (**Figure 3m**). At 100% of the labeled data, the two curves converged for most structures, confirming that pre-training mattered most where labels were scarce and the target was hardest to delineate.

These structure-level patterns were reproduced in sex-stratified analyses pooled across all datasets (female, **Supplementary Figure S4**; male, **Supplementary Figure S5**) and within each individual cohort, both overall and stratified by sex: ExtTest-1 (**Supplementary Figure S6-8**), ExtTest-2 (**Supplementary Figure S9-11**), ExtTest-3 (**Supplementary Figure S12-14**), ExtTest-4 (**Supplementary Figure S15-S17**), and ExtTest-5 (**Supplementary Figure S18**).

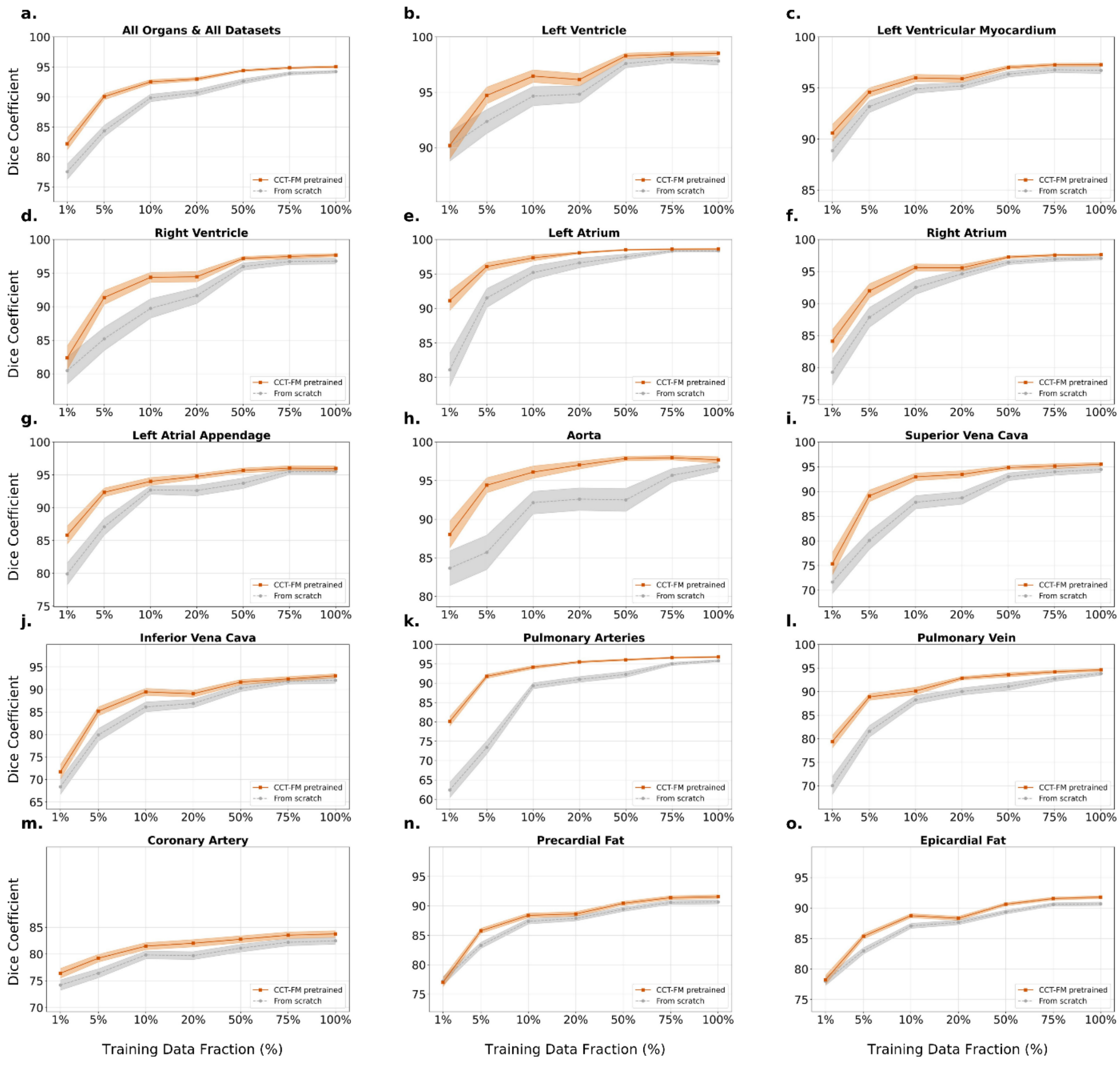

**Figure 3. Per-structure data efficiency of self-supervised pre-training.** Mean Dice coefficient, reported as a micro-average across external datasets (N = 598) and as a function of labeled training data fraction (1–100%), for CCT-FM pretrained (orange) versus from-scratch (gray) models. Results are shown for **(a)** all organs combined and **(b–o)** each of the 14 cardiac structures individually; shaded bands denote 95% confidence intervals. Pretrained models consistently achieve higher accuracy with substantially fewer labeled cases.

## 2.4. Segmentation Benchmark

Across the seven in-house models, full-data accuracy was uniformly high and tightly clustered (**Figure 4**). Mean Dice scores across all fourteen structures fell within a span of under one point for six of the seven models (95.6 to 96.3), with Swin-UNETR trailing slightly at 93.4 (**Figure 4b, 4d**). The ranking by structure mirrored the same per-structure analysis, with the cardiac chambers and the aorta approaching the ceiling (left atrium, ventricles, and aorta all near 98 to 99) and the coronary arteries remaining the hardest target for every model, plateauing at roughly 83 to 85 Dice. Given the full labeled set and augmentation, architecture choice therefore had little effect on whole-heart accuracy, and the coronary arteries set the performance floor.

The differences with existing open-source tools were notable. Taking CCT-FM as the representative model, it outperformed TotalSegmentator, MOOSE, and Atlas on every structure the tools shared (**Figure 4c, 4e**). The gaps were widest exactly where segmentation is hardest and clinically most consequential. On the coronary arteries, CCT-FM reached 85.0 Dice against 49.1 for TotalSegmentator, a margin of roughly 36 points, while MOOSE and Atlas did not produce a coronary label at all. On the aorta, which our model segmented at 98.3, Atlas fell to 72.0. Coverage differed as well as accuracy, with the left atrial appendage, pulmonary vein, and superior vena cava absent from all three open-source tools, leaving the whole-heart aggregate undefined for those models (**Figure 4e**). The qualitative example reflects these numbers, with CCT-FM closely matching the ground truth while the open-source tools showed missing or anatomically inconsistent labels (**Figure 4a**).

These benchmark comparisons were reproduced in sex-stratified analyses pooled across all datasets (female, **Supplementary Figure S19**; male, **Supplementary Figure S20**) and within each individual cohort, both overall and stratified by sex: ExtTest-1 (**Supplementary Figure S21-23**), ExtTest-2 (**Supplementary Figure S24-26**), ExtTest-3 (**Supplementary Figure S27-29**), ExtTest-4 (**Supplementary Figure S30-32**), and ExtTest-5 (**Supplementary Figure S35**).

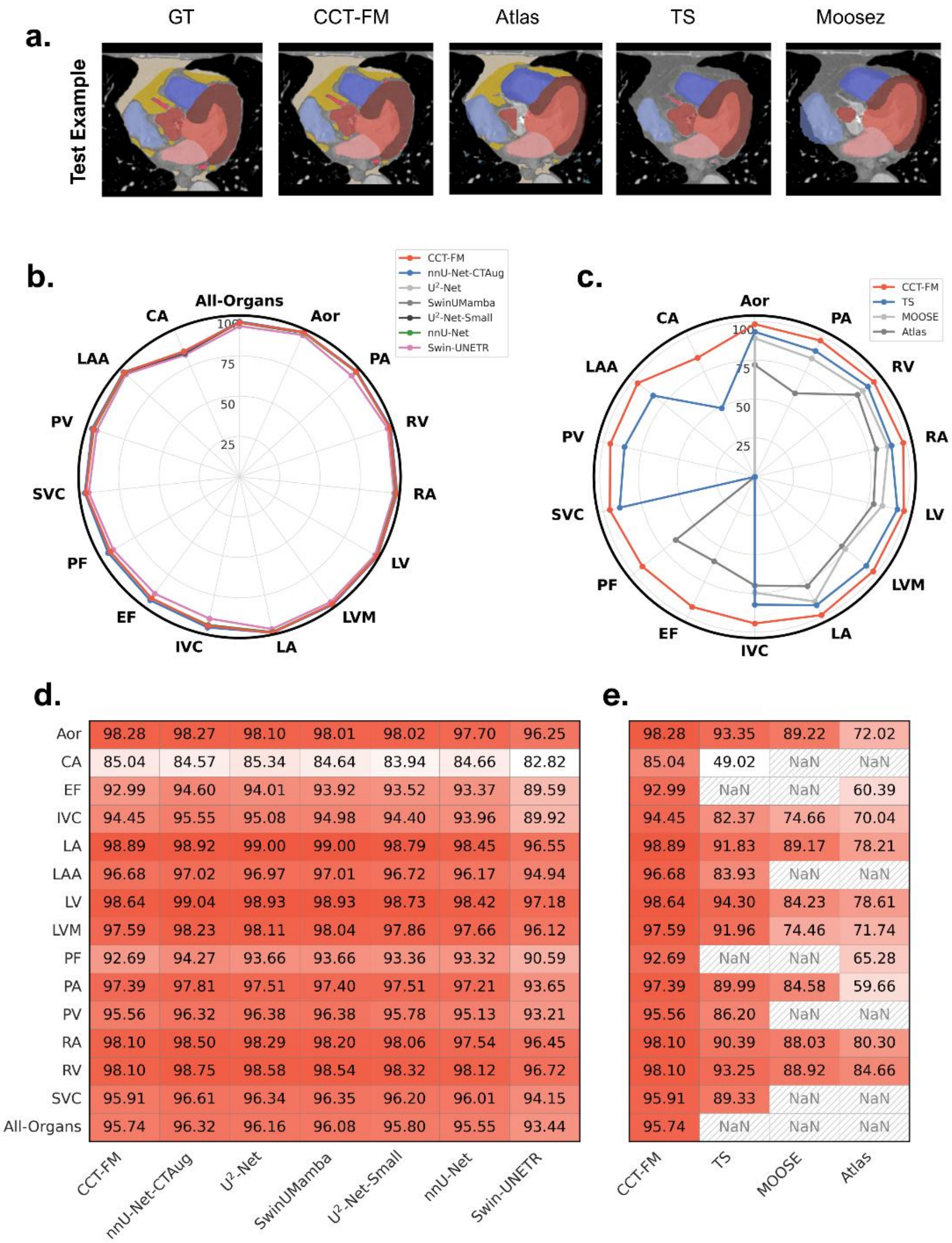

**Figure 4. Whole-heart segmentation benchmark across in-house models and open-source tools. (a)** Showcase of a representative external test case, showing the expert ground truth (GT) alongside CCT-FM and three open-source tools (Atlas, TotalSegmentator, and MOOSE). **(b)** Per-structure Dice coefficients for the seven in-house models (CCT-FM, nnU-Net, nnU-Net-CTAug, SwinUMamba, Swin-UNETR, U²-Net, and U²-Net-Small) across all fourteen cardiac structures, where All-Organs denotes the mean across structures. **(c)** Per-structure Dice coefficients comparing CCT-FM with the three open-source tools, restricted to the structures each tool produces. **(d)** Heatmap of mean Dice for every in-house model and structure, with the All-Organs row giving the across-structure mean. **(e)** Heatmap of mean Dice for CCT-FM against the three open-source tools, where hatched cells mark structures a tool does not produce and that are therefore excluded from its aggregate. All values are mean Dice computed across the external test datasets at the full labeled training set. Structure abbreviations: Aor, aorta; CA, coronary arteries; EF, epicardial fat; IVC, inferior vena cava; LA, left atrium; LAA, left atrial appendage; LV, left ventricle; LVM, left ventricular myocardium; PA, pulmonary arteries; PF, pericardial fat; PV, pulmonary vein; RA, right atrium; RV, right ventricle; SVC, superior vena cava.

## 2.5. Clinical-attribute correlation

To test the clinical relevance of the CCT-FM-derived volumes, we quantified their concordance with operator-derived echocardiographic measurements and assessed their information gain for predicting clinical and laboratory parameters. CT-derived left-ventricular myocardial mass (n = 953, ExtClin-1, **Figure 5a**) and ejection fraction (n = 1,091, ExtClin-2, **Figure 5b**) correlated with their echocardiographic references (Pearson r = 0.70 and 0.72, both $p < 0.001$). This analysis was reproduced in sex-stratified analyses (**Supplementary Figure S34, Supplementary Figure S3**5).

Across the 17 clinical and laboratory targets, adding the eight CT-derived substructure volumes to a demographic baseline (age, sex, and body-surface area) increased out-of-fold $R^2$ for five targets after Benjamini–Hochberg false-discovery-rate (FDR) control at q = 0.05 (ExtClin-3, **Figure 5c**). Two of the three pre-specified primary outcomes were confirmed and remained significant under both the FDR-adjusted bootstrap and the FDR-adjusted per-fold cross-check: left-ventricular ejection fraction ($\Delta R^2 = +0.27$; BCa 95 % CI [+0.23, +0.32]) and the mean trans-aortic gradient ($\Delta R^2 = +0.04$; [+0.03, +0.06]); the third primary, brain natriuretic peptide, cleared the false-discovery threshold but not the per-fold cross-check ($\Delta R^2 = +0.07$; [+0.01, +0.14]). In the exploratory family (aortic stenosis cohort), EuroSCORE II ($\Delta R^2 = +0.03$; [+0.02, +0.05]) and the STS Score ($\Delta R^2 = +0.02$; [+0.01, +0.03]) also cleared FDR control, albeit with minimal information gain.

Feature attribution localized the signal to the anatomically expected structures, and this localization remained stable whether each predictor was examined in isolation or within the joint model (**Supplementary Figure S36**). With each predictor fitted alone and adjusted for age and sex, the dominant contributor was the left-ventricular cavity for ejection fraction (SHAP-$\eta$ = 0.39), both ventricular cavities for natriuretic peptide ($\eta$ = 0.26 and 0.23), and left-ventricular myocardial mass for the trans-aortic gradient ($\eta$ = 0.14) (**Supplementary Figure S36**.b). Sex-stratified analysis preserved this anatomical localization, with the incremental signal generally more pronounced in males (**Supplementary Figure S38**) than in females (**Supplementary Figure S37**).

The eight CT-derived volumes alone, without any demographic input, recovered most of the Full-model $R^2$ for the three primary outcomes ($R^2_{\text{volume-only}}$ = 0.27 for ejection fraction, 0.07 for natriuretic peptide, 0.04 for the gradient), while body-mass index served as a successful negative control through its known body-surface-area-to-body-mass-index tautology (**Supplementary Figure S39**). Relationships were largely linear, except for ejection fraction (**Supplementary Figure S40**.a), and predictions for the primary outcomes were calibrated (**Supplementary Figure S40**.b).

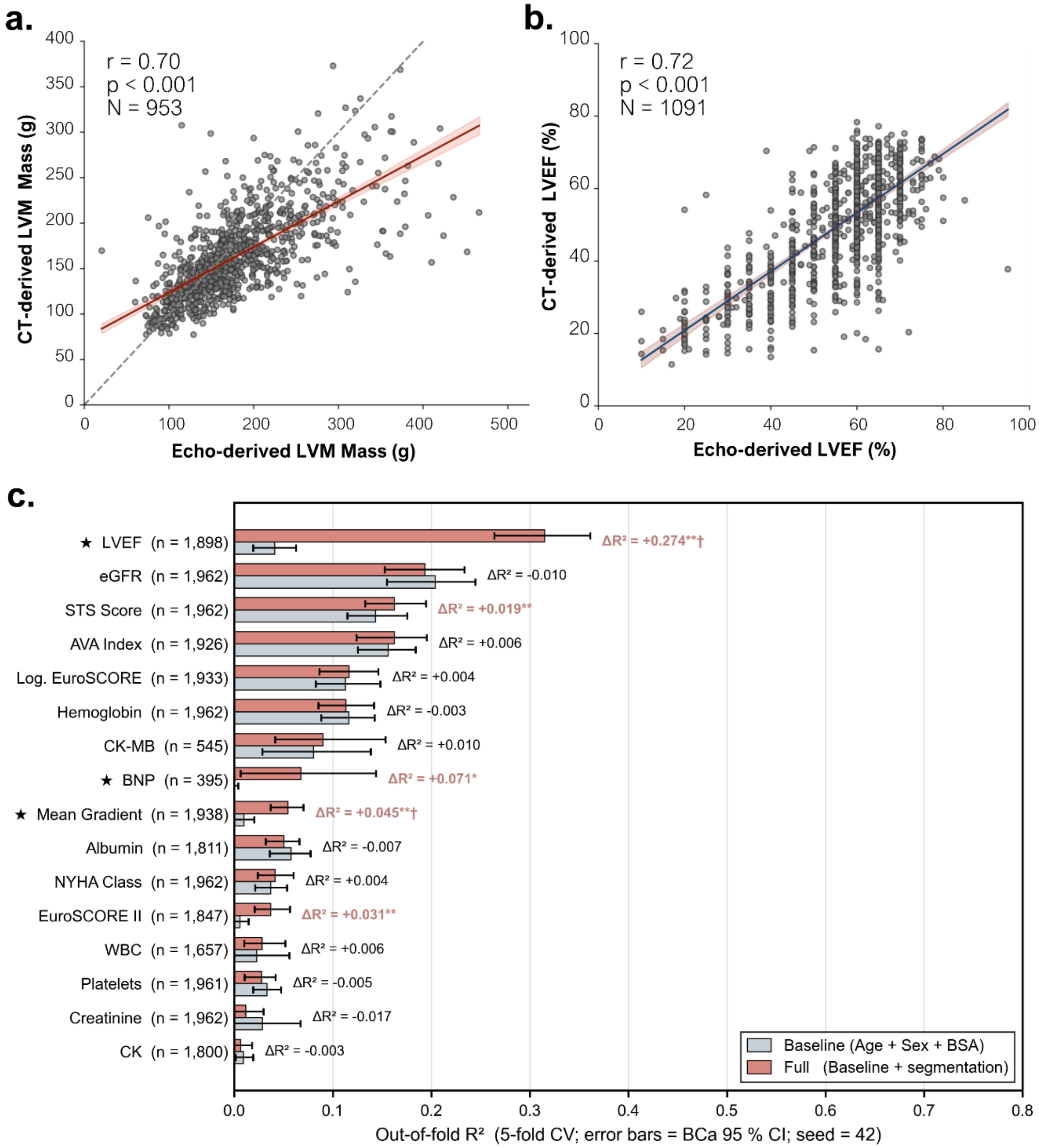

**Figure 5. Clinical validation of CT-derived cardiac measurements by concordance with echocardiography (a, b) and association with clinical parameters (c). (a, b)** CT-derived left-ventricular myocardial mass (a, ExtClin-1, N = 953) and LVEF (b, ExtClin-2, N = 1,091) plotted against the matched echocardiographic measurement, with OLS fit and shaded 95 % band (Pearson r = 0.70 and 0.72, both $p < 0.001$). **(c)** Out-of-fold $R^2$ (5-fold cross-validation) of a Baseline CatBoost model (age, sex, body-surface area; grey) and a Full model adding the eight CT-derived substructure volumes (red), ordered by full-model $R^2$ (ExtClin-3). $\Delta R^2$ (Full − Baseline) is annotated beside each pair. Error bars are bias-corrected accelerated (BCa) 95 % confidence intervals for each model's $R^2$ (patient-level bootstrap); the significance of $\Delta R^2$ is assessed from the paired bootstrap of the difference. The false discovery rate (FDR) was controlled at q = 0.05 with the Benjamini–Hochberg procedure, applied separately within the pre-specified primary outcomes (★; LVEF, BNP, mean trans-aortic gradient) and the exploratory family. Asterisks mark targets whose $\Delta R^2$ remains significant after FDR control (* adjusted $p \leq 0.05$, ** $\leq 0.01$); a dagger (†) marks those additionally corroborated by the FDR-adjusted per-fold paired t-test.

## 2.6. Population-level Phenotyping

Time-resolved 4D CT captured cardiac motion (ExtClin-4) in the aortic stenosis cohort. Per-phase volumes traced the mechanics of the cardiac cycle, with the chambers dividing into two reciprocal groups **(Figure 6)**. The ventricles showed the largest cyclic excursion, with the mean left ventricular volume rising from about 56.16% to 93.27% of its normalized maximum and the right ventricular volume from about 65.39% to 93.35%, each peaking near mid-cycle (**Figure 6a, 6c**). The atria moved in antiphase and reached their minima at the same phase, the mean left and right atrial volumes falling to roughly 79.38% and 73.22% before refilling (**Figure 6b, 6d**). The left ventricular myocardium and the epicardial and pericardial fat compartments stayed nearly constant throughout, each varying by only a few percent about its mean (**Figure 6e, 6g, 6h**).

The cross-sectional 3D cohort characterized cardiac aging (ExtClin-5) in a mixed population (normal, aortic stenosis, coronary artery disease, and coronary anomaly). Volumes varied systematically with age, and the direction of the age association differed across structures and between sexes **(Figure 7)**. Ventricular and myocardial volumes were lower at older ages, the median left ventricular volume measuring about 125.10 mL at age 46 and 84.95 mL at age 91 in women, and 164.21 mL and 115.86 mL in men, with the right ventricle and left ventricular myocardium following the same downward course (**Figure 7a, 7c, 7e**). The atria and left atrial appendage showed an opposite change in dimensions, the median left atrial volume rising from about 67.43 mL at age 46 to 101.48 mL at age 91 in women and from 76.91 mL to 111.65 mL in men, an increase that steepened after roughly age 65 (**Figure 7b, 7d, 7f**). Both fat compartments also enlarged with age (**Figure 7g, 7h**). Across all structures, men had larger absolute volumes than women, while the age trends were in the same direction in both sexes (**Figure 7**).

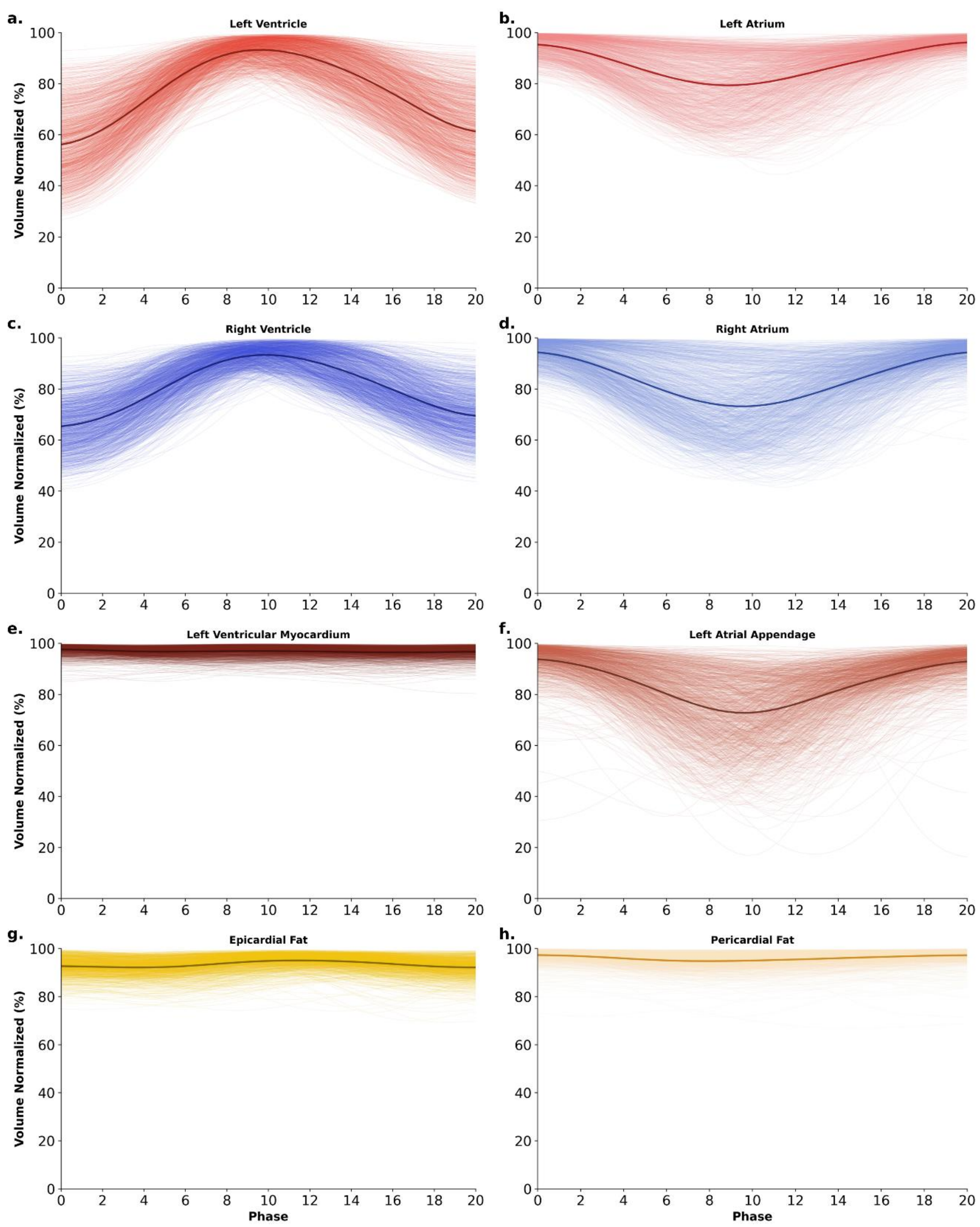

**Figure 6. Phase-resolved volumetric dynamics across the cardiac cycle (ExtClin-4, n = 1,245 patients with 4D CT and 23,142 3D CT).** Volume (mL) across cardiac phases for eight cardiac structures (a–h). Faint lines represent individual patients, and bold lines represent the cohort mean, illustrating the characteristic filling and emptying dynamics of each structure, automatically recovered across the longitudinal four-dimensional cohort.

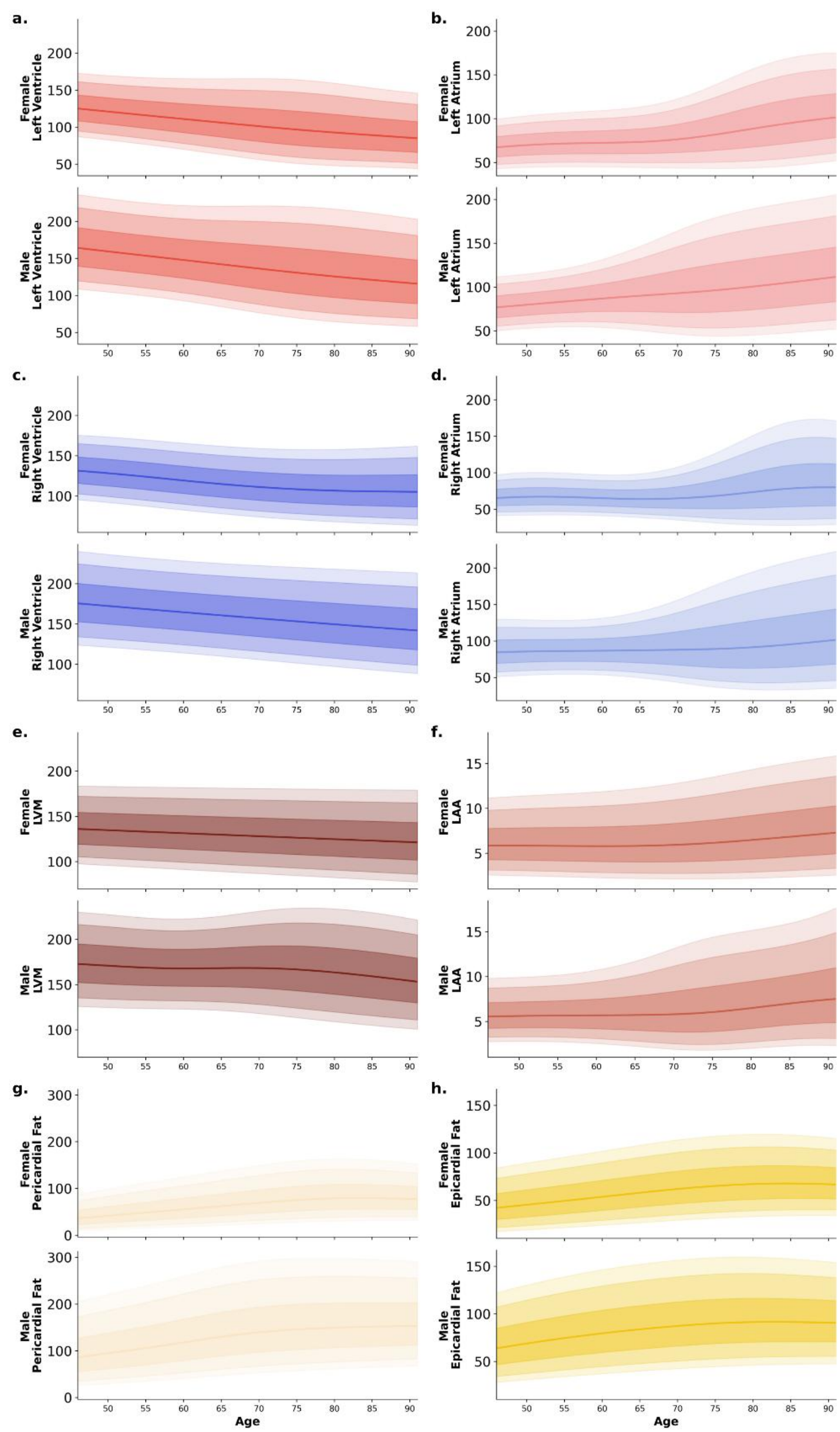

**Figure 7. Sex-stratified age percentile curves for cardiac structure volumes (ExtClin-5, n = 3307).** Population percentile curves modeling structure volume as a function of age, fitted separately for female and male subjects using Generalized Additive Models for Location, Scale and Shape, for eight cardiac structures **(a–h)**. Shaded bands denote nested percentile ranges across the population.

## 2.7. Carbon footprint estimate

On the low-carbon Switzerland grid used for development, pre-training and training all six benchmark models together would emit 11 kg $CO_2e$ (**Supplementary Figure S42**). Because $CO_2e$ scales almost linearly with grid carbon intensity, the model's footprint is influenced more by where it is run than by which architecture is chosen (full methods and results in **Supplementary Note S2**).

## 2.8 Open science

To support reproducibility and reusability, we release the resources generated in this study as open-access and open-source assets. These comprise the expert-validated 14-structure segmentations for the supervised training (train; n = 1000) dataset and our expert segmentations for the one external validation set (ExtTest-5; n = 20). We also release the coronary artery centerlines extracted from the training and MM-WHS data for 1,020 cases (train and ExtTest-5), which can be used to develop centerline detection models (an example image of a left coronary artery render is available in **Supplementary Figure S43**). In addition, we release 14,280 STL structures that could be used to develop shape-aware networks for cardiac imaging and to support cardiovascular simulation, 3D printing, and physical and digital twinning of the heart (**Supplementary Figure S44**). The trained model weights, including the pretrained foundation model, CCT-FM, and other benchmarking architectures; the CCT augmentation library (CTAug); and the nnUZoo23 framework with all new 3D segmentation model implementations were also made publicly available for further use and study. To facilitate clinical and research adoption, the model is integrated into our group's desktop application called HolOrama[24,] which is freely available for Linux and Windows, enabling CCT segmentation without programming expertise (**Supplementary Figures S45-49**).

## 3. Discussion

We developed a unified framework for comprehensive CCT segmentation and phenotyping that combines a human-in-the-loop annotation pipeline, a CCT augmentation library, and a self-supervised foundation model pre-trained on large-scale unlabeled CCT volumes. We benchmarked different architectures and compared against available open-source tools. Across five external datasets, the framework segmented 14 cardiac structures more accurately and comprehensively than existing open-source tools, and self-supervised pre-training improved label efficiency, most evident in the low-regime data scenario. Architecture benchmarking showed that different deep learning architectures performed comparably, indicating that data quality, rather than architecture, drove performance. The framework showed moderate agreement with echocardiographic measurement and scaled to population-level cardiac phenotyping, supporting reproducible and opportunistic cardiac quantification from scans already acquired in routine care.

Through the human-in-the-loop pipeline, we assembled one of the largest and most complete expert-annotated CCT segmentation dataset to date[20]. Existing open-access resources are limited in several respects. Many are not acquired using dedicated CCT protocols and therefore lack ECG gating and contrast enhancement, which blurs the borders between structures and limits the accuracy and certainty of models trained on them[9,10]. Moreover, datasets that use cardiac acquisitions typically annotate only the major chambers or a single additional structure, and comprise just a few dozen cases[16,25],[26]. Our dataset exceeds these in both case volume and anatomical coverage, and we release it alongside open model weights, an augmentation library, and a software tool for CCT segmentation.

Beyond the dataset, self-supervised (SSL) pre-training conferred its clearest advantage under external evaluation in the low-data regime, indicating improved generalizability and transferability across institutions and scanners. As labeled cases increased from hundreds to thousands, this advantage narrowed, and the SSL benefit also depended on the target structure[12,27]. Well-defined structures whose boundaries can be inferred from the surrounding context, such as the aorta, pulmonary arteries, and ventricles, benefited most, whereas the diffuse, low-contrast epicardial and pericardial fat gained little. These effects were clearest for contrast-enhanced datasets, the main task and data in this study, where external evaluation in the low-data regime showed significant gains from SSL. For non-contrast datasets and coronary artery anomaly segmentation, however, the absence of an external test set and the small size of the available data left the benefit of SSL less clearly established, even where pre-training still improved performance.

For CCT segmentation, previous studies have proposed foundation models development using different approaches. One study applied self-supervised pre-training on a large unlabeled dataset with an xLSTM backbone to segment cardiac structures across CT and MRI[28]. Although their pre-training corpus comprised roughly 30K CT scans, only about 1,000 were dedicated CCTs, with the remainder being non-cardiac CTs, including cardiac regions captured in a lung cancer cohort.

The downstream task was trained on only 20 cases from an open-source whole-heart dataset[16,25] (four main chambers, aorta, and pulmonary artery segmentation) and evaluated by five-fold cross-validation. Another study[18] proposed domain adaptation for CT subtraction using a DIstillation of No Labels (DINO) approach, employing no CCT during pre-training and only 69 cases for the downstream task, again evaluated by five-fold cross-validation. Both studies reported significant improvements from foundation-model pre-training, consistent with our finding that pre training helps most when labeled data are scarce. However, in both cases the pre-training data were not dedicated CCT, and the downstream tasks were evaluated on only a few dozen cases by cross-validation without external validation, which may limit the generalizability of their models. In our study, by contrast, the majority of pre-training data were acquired with dedicated CCT protocols, and the model was evaluated on CCT segmentation across multiple external test sets.

In our study, the importance of data, rather than model design, was further supported by our architecture benchmarking and new augmentation techniques, developed for this study. Across network families spanning convolutional, transformer, and state-space (Mamba) designs, $U^2$-Net variants, nnU-Net, and our foundation model, performance converged to comparable levels when training data were sufficiently large and clean[20]. The small differences between architectures suggest that, given adequate data quality and scale, performance depends more on the training data than on architectural choice, technique, or network generation, in line with prior work reporting that data quality outweighs architectural novelty[29]. Consistent with this data-centric view, the domain-specific augmentation library developed in this study improved accuracy, robustness, and generalizability while holding the architecture fixed, most notably for the challenging cases encountered in external validation.

Using the segmentation model developed in this study, we evaluated the measurement of clinical parameters as an opportunistic, added-value application on an unlabeled dataset and showed that automated ejection fraction and LV mass derived from CT were moderately correlated with echocardiographic measurements. This correlation should be interpreted cautiously, with several factors in mind. CT measurements rest on directly measured volumes rather than the geometric assumptions on which echocardiography depends, so part of the discordance may lie in the reference standard rather than in the model, although only a volumetric standard such as cardiac magnetic resonance (CMR) could arbitrate this[5,30]. In 4D imaging analysis, LV mass remained highly consistent across volumes irrespective of cardiac phase, demonstrating the reproducibility of CT-derived mass; in contrast, echocardiographic LV mass estimation is operator- and view-dependent and subject to high inter- and intra-observer variability. CMR provides the volumetric reference standard, albeit with its own contour-dependence.

Beyond direct quantification, the segmented volumes carried indirect information about ventricular function, aortic-stenosis severity, and a circulating biomarker of heart failure. However, the associations were modest and selective, and most laboratory targets showed none. These findings should therefore be viewed as evidence that segmented anatomy may carry functionally relevant information rather than being used as a basis for clinical prediction in cases without a structural correlation. Prior studies have shown prognostic value for cardiac substructures in chamber volumes, epicardial fat, and coronary calcium across several large cardiac and non-cardiac CT cohorts[6,31–33]. Our 4D analysis confirms that chamber size varies markedly throughout the cardiac cycle. Therefore, measurements from non-gated or non-cardiac CT should be interpreted with caution, bearing this variation in mind. By contrast, epicardial fat, pericardial fat, and LV myocardium, remained stable across cardiac phases, indicating that these measures are largely independent of cardiac phase and could serve as reproducible imaging biomarkers in contrast-enhanced CT. Future work should assess their reproducibility and agreement in non-contrast CT and relate the quantified measures to patient outcomes to establish the added value of this framework for cardiac disease.

The study has some limitations that provide clear direction for future work. First, while it provides the most comprehensive whole-heart segmentation yet, it leaves several clinically relevant structures uncovered, including pericoronary fat, coronary artery plaque, coronary artery calcification, and the valves. Future work building on our segmentation could incorporate these substructures. We also extended segmentation to non-contrast images by training on contrast-enhanced photon-counting CT and transferring the masks to virtual non-contrast reconstructions, but because these are virtual rather than true non-contrast images, and because structure borders are inherently less defined without contrast, the resulting segmentations are necessarily approximate; generative approaches that synthesize realistic non-contrast CT and transfer segmentations accordingly could address this. Finally, although we demonstrated clinical validation of the segmentation, future studies should incorporate a wider range of clinical outcomes to establish the added value of the derived parameters, and quantitative metrics should be applied to large-scale datasets to define CT-based reference thresholds and standards, comparable to those established for other modalities, enabling population-level analysis.

In conclusion, this framework reliably quantifies cardiac structures, with performance validated across multiple external test sets. It potentially opens several opportunistic applications: relating the shape or density of individual structures to clinical outcomes, enabling radiomic analyses, and supporting 4D assessment of functional metrics such as 3D strain. The resulting segmentations could further serve as a basis for disease-specific planning, three-dimensional visualization, patient-specific flow simulation, and physical and digital twinning of the heart. By releasing the dataset and model openly, we provide a foundation that the wider research community can build upon and extend.

# 4. Methods

## 4.1. Overview of Study Design

**Figure 1** provides the complete graphical overview of the study pipeline, dataset sources, human-in-the-loop diagram, and evaluation framework.

### 4.1.1. Primary Training Datasets

This study received institutional review board approval at all in-house participating centers for different datasets (Bern: KEK 2024-01345, KEK 2021-01738, KEK 2021-0058, and KEK 2020-00841; Zurich: KEK 2015-0235 and KEK 2014-0632; Japan: no. 6970). All datasets are summarized in **Table 1**. The primary training set comprised coronary CT volumes from the publicly available dataset[26], which underwent systematic human-in-the-loop correction to yield consistent segmentations of 14 cardiac structures, including the original coronary segmentation. The self-supervised pre-training corpus comprised unlabeled cardiac, non-contrast, and whole-body CT volumes from an in-house dataset, supplemented by public datasets including LUNA16[34]. No overlap exists between the pre-training, training, and external test cohorts.

### 4.1.2. External Test Datasets

Five independent external test datasets were recruited to assess segmentation accuracy and generalization across diverse CCT imaging contexts: standard diagnostic CCT from two independent institutions, Bern (ExtTest-1; n = 173) and Zurich (ExtTest-2; n = 198); whole-body photon counting CT Bern cropped to heart (ExtTest-3; n = 132); artifact-rich cases with metallic implants (ExtTest-4; n = 75); and the publicly available MM-WHS dataset (ExtTest-5; n = 20) [3]. Ground-truth segmentations for all five datasets were generated through manual labeling. Full dataset characteristics are summarized in **Table 1**.

### 4.1.3. Clinical Validation and Population Analysis Cohorts

Five cohorts were assembled for clinical validation. A cohort with paired CCT and transthoracic echocardiography (ExtClin-1; *n* = 953, from Bern and Japan) was used for agreement analysis against echo-derived LVM, and a four-dimensional cohort with paired echocardiography (ExtClin-2; *n* = 1,091, Bern) for agreement against echo-derived LVEF; A third cohort combining CCT with demographic, clinical, and laboratory data (ExtClin-3; *n* = 1,962, Bern and Japan) was used to assess the incremental information gain of CT-derived measures. Two further cohorts supported population-scale phenotyping: a longitudinal multi-phase four-dimensional dataset (ExtClin-4; *n* = 1,245 patients, 23,142 3D volumes at ~21 cardiac phases each, Bern) and a cross-sectional cohort spanning three countries (ExtClin-5; *n* = 3,307, Bern, Zurich, and Japan). Full cohort characteristics are summarized in **Table 1**.

## *4.2. Image Preprocessing*

All CCT images were resampled to dataset-specific isotropic voxel resolutions using third-order spline interpolation, while ground truth labels were resampled using nearest-neighbor interpolation to avoid interpolation artifacts in discrete anatomical boundaries. Image intensities were windowed to the 0.5 to 99.5 Hounsfield unit (HU) percentile range to suppress irrelevant anatomical structures outside the field of interest, followed by z-score normalization to accommodate deep neural network learning dynamics. All datasets underwent identical intensity normalization and standardization pipelines, ensuring that segmentation model outputs could be directly compared across datasets without corrections for imaging protocol or intensity scale differences.

## *4.3. Data-Centric Pipeline*

### 4.3.1. Human-in-the-Loop Labeling Process

At the beginning, five cases were labeled with a semi-automated approach for 14 target structures by two medical students under the direct supervision of cardiac imaging specialists using ITK-SNAP[35], with one annotating the coronary arteries and the other the remaining 13 structures. The model was applied iteratively to unlabeled cases: in each cycle, human observers prioritized correcting the most inaccurate and challenging cases, the model was retrained on the expanded set, and the process was repeated until all 1,000 cases were labeled. The model was trained to delineate 14 cardiac structures: left ventricle, right ventricle, left atrium, right atrium, left atrial appendage, left ventricular myocardium, ascending aorta, pulmonary arteries, pulmonary veins, superior vena cava, inferior vena cava, coronary arteries, epicardial fat, and pericardial fat. Anatomical definitions consisting of standard cardiac imaging conventions were provided to both annotators.

### 4.3.2. Cardiac CT Data Augmentation

To improve robustness to clinical image degradation, we developed **CTAug**, an open-source Python library that simulates various artifact categories. Augmentations were applied during training with probabilities of 15% (calcification, wire) and 10% (metal, step-and-shoot). Calcifications were simulated as small high HU deposits; metal artifacts as high HU regions with radiating bright and dark streaks; wires as linear or curvilinear high HU structures; and step-and-shoot motion as stepwise intensity discontinuities with 1 up to 2.5% spatial shifts. For metal, wire, and calcification augmentations, positioning was guided by the segmentation mask to ensure anatomically plausible localization. **Supplementary Figure S1.a** shows representative examples.

## *4.4. Model Architecture and Training*

### 4.4.1. Self-Supervised Pre-Training

To learn generalizable representations and reduce dependence on annotated data, the model was pretrained on 60,331 unlabeled CCT volumes prior to supervised fine-tuning, using a masked autoencoder (MAE)[17] strategy that reconstructs randomly masked regions of the input volume. The architecture used the nnU-Net 3D ResEnc backbone with skip connections removed, so the decoder was reconstructed solely from the encoder's latent representation (Figure 1b). The encoder processed 256 × 256 × 256 patches across six resolution stages (~192 million parameters), with a symmetric decoder. During pre-training, 75% of input patches were masked to zero, and the decoder predicted their original intensities, minimizing an equally weighted combination of mean squared error and mean absolute error. Pre-training used SGD with Nesterov momentum (learning rate 0.01, cosine annealing) over 1,000 epochs on a single NVIDIA H200 NVL 140GB GPU with batch size 2 (~50 GPU-hours); the checkpoint with the lowest validation reconstruction loss was carried forward.

#### 4.4.2. Supervised Fine-Tuning

For fine-tuning, multi-scale skip connections were reintroduced, the encoder was initialized from the self-supervised checkpoint, and the decoder was randomly reinitialized. The encoder was frozen for a warmup period of 10 epochs for labeled-data subsets ≤10% and 25 epochs for larger subsets, alongside a 5-epoch linear learning-rate warmup. Fine-tuning used AdamW[36] (initial learning rate 0.001, cosine annealing, batch size 2), whereas the scratch baseline used SGD with Nesterov momentum (initial learning rate 0.01, polynomial decay, batch size 2); both used a combined Dice and cross-entropy loss. To quantify the benefit of pre-training, the pretrained and scratch variants were trained under identical settings at labeled-data fractions of 1, 5, 10, 20, 50, 75, and 100%, with epochs scaled to subset size (50 at 1% up to 250 at full) across all three downstream tasks.

The first task was the primary whole-heart segmentation on the benchmarking train CCT dataset (DownStr-1), trained on the expert-corrected 14-structure labels and evaluated across the five external test datasets. The second task addressed coronary anomaly segmentation (DownStr-2) using a cohort of 29 cases, all containing coronary artery anomalies, with 4 held out for testing. All 14 structures were segmented, and evaluation focused on coronary artery anomalies. The third task addressed cross-domain adaptation to non-contrast CT (DownStr-3) using a cohort of 64 cases, with 6 held out for testing and all 14 structures segmented to evaluate generalization from contrast-enhanced CCT to non-contrast acquisitions.

#### 4.4.3. Segmentation Models Benchmarking

Six trainable 3D segmentation architectures were developed and evaluated in this study, spanning convolutional, transformer-based, and Mamba-based designs. All architectures are integrated into our open-source nnUZoo[23] framework, which we extended to support 3D models in this study. In particular, we developed three-dimensional extensions of the Mamba-based SwinUMamba and the multi-scale U²-Net architectures: the 3D SwinUMamba replaces the attention encoder in its transformers based counterparts with a volumetric state space model (SS3D) that scans features along all three spatial axes in both directions, capturing long-range anatomical context at linear computational cost, while the 3D U²-Net extends its nested residual U-blocks to volumetric convolutions for multi-scale feature extraction. Details of model architectures are provided in the supplementary dataset. Additionally, three open-source models (TotalSegmentator, Atlas, and MOOSE) were evaluated to establish an external reference benchmark.

#### 4.4.4. Training Configuration and Data Splits

All six trainable models used identical 800/200 train/validation splits, with checkpoints selected on best validation Dice and cross-entropy loss. Training ran on a single NVIDIA A100 80GB GPU (batch size 2, mixed precision) for 1,000 epochs, with the spatial and intensity augmentations of Section 4.3.2 and mirroring disabled to preserve cardiac orientation. Optimizer settings differed by architecture family: nnU-Net used SGD with Nesterov momentum (learning rate 0.01, weight decay 3e-5, polynomial decay), while the SwinUNETR, SwinUMamba, and U²-Net variants used AdamW (learning rate 1e-4, weight decay 5e-2, cosine annealing).

## *4.5. Model Evaluation*

### 4.5.1. Segmentation Benchmark

Segmentation was evaluated using the Dice similarity coefficient, computed per structure and averaged across the 14 structures. All six trainable models were assessed on the five external test datasets, per structure and per dataset, with results aggregated across datasets to compare architecture families. TotalSegmentator, Atlas, and MOOSE were evaluated using their published pretrained weights without fine-tuning under default inference, with aorta segmentations cropped to the ascending aorta extent of the reference labels for fair comparison. For TotalSegmentator, we used the cardiac-specific model and the coronary artery segmentation model for the structures they cover, and the default model for the remaining structures, so that the best available model was used for each comparison.

### 4.5.2. Data Efficiency Evaluation

The benefit of self-supervised pre-training was quantified by comparing pretrained and scratch-initialized models across all labeled data fractions for each downstream task. For the full cardiac structure segmentation task, models trained on DownStr-1 fractions were evaluated on the five external test datasets, while the non-contrast CT and coronary anomaly tasks were each evaluated on their respective held-out test sets. For each fraction, mean Dice scores were recorded to construct learning curves, enabling assessment of the performance gap between pretrained and randomly initialized models as a function of labeled data availability. The data efficiency advantage was evaluated both at the aggregate level across all structures and per individual structure to identify which benefited most from pre-training under limited annotation budgets.

## 4.6 Clinical and Population-Scale Analysis

### 4.6.1 Model-based CT-derived quantification and echocardiography measurement

CT-derived left ventricular myocardial mass was compared against paired echocardiographic LV mass in 953 patients with both measurements available (ExtClin-1, Inselspital Bern and Japan). Mass was computed from the LV-myocardium segmentation at the end-systolic phase using a density factor of 1.05 g/mL. Agreement was assessed by Pearson correlation.

CT-derived LV ejection fraction was compared against paired echocardiographic ejection fraction in 1,091 patients (ExtClin-2, Inselspital Bern). Ejection fraction was computed from the maximum (end-diastolic) and minimum (end-systolic) LV cavity volumes across the 4D cardiac cycle. Patients without both phases, including the Japanese cohort for which only the end-systolic phase was available, were excluded. Agreement was assessed by Pearson correlation.

### 4.6.2 Information gain on clinical and laboratory variables

In a cohort of 1962 patients from Bern and Japan centers (ExtClin-3 dataset), we tested whether the CCT-FM-derived substructure volumes carry information about 17 clinical and laboratory parameters beyond routine demographics. Three outcomes were pre-specified as primary before unblinding (LVEF, BNP, and the mean trans-aortic gradient) and the remaining 14 as exploratory.

Patients with implausible body-surface area (outside 1.2–2.5 $m^2$) were excluded and targets with fewer than 150 complete cases were dropped; creatine kinase and CK-MB were log1p-transformed and implausible albumin values set to missing. For each target we fitted nested CatBoost regressions under a common five-fold cross-validation. We fitted a Baseline model on age, sex, and body-surface area (3 features) and a Full model adding the eight raw substructure voxel volumes (11 features) and summarized incremental value as $\Delta R^2 = R^2_{full} - R^2_{baseline}$ on pooled out-of-fold predictions. For body-mass index, a four-model decomposition (M1, age and sex; M2, +body-surface area; M3, +volumes; M4, volumes without body-surface area) serves as a negative control that separates body size from cardiac contribution. A RidgeCV linear baseline fitted under the same cross-validation tested whether the gains required CatBoost's non-linear modelling rather than a linear fit, as well as showing the linearity among relationships.

$\Delta R^2$ was reported with a paired-patient bias-corrected accelerated bootstrap (1,000 resamples) for 95 % confidence intervals and two-sided p-values, with Benjamini–Hochberg false discovery rate (FDR) control at $q = 0.05$ applied separately within the primary (3) and exploratory (14) families; a per-fold paired t-test across the five folds served as a parametric cross-check, and a primary outcome was deemed confirmed when significant under both the FDR-adjusted bootstrap and the FDR-adjusted t-test. Per-feature contributions were quantified by the SHAP-η ratio ($\eta_j = \sqrt{(Var(\varphi_j)/Var(y))}$) from CatBoost's TreeSHAP on held-out folds, reported pairwise (each predictor alone, unadjusted and adjusted for age and sex) and jointly within the Full model, and calibration of the primary-outcome models was assessed by regressing observed on predicted out-of-fold values. All cross-validation splits, training, and resampling used a fixed seed (42) and a single pre-specified CatBoost configuration (an optional nested hyperparameter search was implemented but not enabled). Full details, hyperparameters, software versions, and the figure mapping are in Supplementary Note S4.

### 4.6.3 Phase-resolved volumetric dynamics across the cardiac cycle

In the longitudinal 4D cohort (1,245 patients, 23,142 phase reconstructions; ExtClin-4 dataset), per-phase volumes were extracted for eight cardiac structures (left ventricle, right ventricle, left atrium, right atrium, left atrial appendage, left ventricular myocardium, epicardial fat, and pericardial fat) across the reconstructed cardiac phases. For each patient, the phase series was temporally aligned to the phase of minimum LV volume, Gaussian-smoothed along the phase axis ($\sigma = 2$), and resampled onto an upsampled phase grid by quadratic interpolation. Per-patient absolute volume–phase curves (mL) and their cohort mean were used to characterize each structure's dynamics descriptively; no scalar metrics were derived from the curves.

### 4.6.4: Population age- and sex-specific cardiac phenotyping

In the cross-sectional 3D cohort (3,307 patients; ExtClin-5 dataset), volumes were derived for the same eight cardiac structures. Sex-specific percentile curves for each structure were modeled as a function of age using Generalized Additive Models for Location, Scale and Shape (GAMLSS)[37] with age and body mass index as covariates, fitted separately for female and male subjects. The independent contribution of age was assessed by ordinary least squares regression adjusted for dataset of origin and body mass index. GAMLSS models were fitted in R; all other analyses were performed in Python.

### *4.7. Computational Environment*

All training and inference used single-GPU configurations: the six trainable models on an NVIDIA A100 PCIE 80GB GPU and self-supervised pre-training on an NVIDIA H200 NVL 140GB GPU. Models were implemented in Python 3.12 with PyTorch (2.4 for SwinUMamba, 2.5.1 for all others).

## *Data Availability*

The manual segmentation of a training dataset of 1,000 patients and the external test set 5 with 20 patients were made publicly available on Hugging Face (https://huggingface.co/datasets/AI-CVM/Cardiac-CT). The remaining external test datasets and the self-supervised pre training corpus contain protected health information and cannot be shared due to patient privacy regulations.

## *Code Availability*

All source code for model training, inference, and evaluation is available at https://github.com/AI-in-Cardiovascular-Medicine/CCT-FM/. The CCT augmentation library (CTAug) is available at https://github.com/AI-in-Cardiovascular-Medicine/CTAug/. The segmentation model implementation is available at https://github.com/ai-in-Cardiovascular-Medicine/nnUZoo. A graphical user interface for CCT (HolOrama) is available at https://github.com/AI-in-Cardiovascular-Medicine/HolOrama. These repositories include configuration files for reproducing all experiments, pretrained model weights, and documentation for applying the models to new CCT datasets.

## *Author Contributions*

Pooya Mohammadi Kazaj: conceptualization, software, methodology, formal analysis, investigation, writing, original draft. Leo Fridolin Weber and Wen Xie: data annotation and curation, writing, review, and editing. Seyed Amir Ahmad Safavi-Naini: software, writing, original draft. Giovanni Baj: software (evaluation), writing, review, and editing. Ali Mokhtari, Anselm Stark and Waldo Valenzuela,: data preprocessing, software, writing, review, and editing. Moritz Hundertmark: supervision of annotation and label quality assurance, writing, review, and editing. Xiaowei Xu: methodology, writing, review, and editing. George CM Siontis, Christoph Ryffel, Toshiya Yoshida, Taishi Okuno, and Yoshihiro Akashi: methodology, writing, review, and editing. Ronny R. Buechel: methodology, writing, review, and editing. Thomas Pilgrim and Stephan Windecker: clinical expertise, writing, review, and editing. Christoph Gräni: provision of the Bern cohort and clinical expertise, writing, review, and editing. Isaac Shiri: conceptualization, supervision, project administration, writing, review, and editing. All authors reviewed and approved the final manuscript.

## *Competing Interests*

**Dr. Okuno** reports research grants from JSPS KAKENHI, the Asahi Intecc–Naohiko Miyata Medical Technology Support Foundation, and the Takeda Science Foundation.
**Dr. Pilgrim** reports research grants from the Swiss National Science Foundation, the Swiss Heart Foundation, the Swiss Polar Institute, the Bangerter-Rhyner Foundation, the Mach-Gaensslen Foundation, and the Monsol Foundation. Research, travel, or educational grants to the institution without personal remuneration from Biotronik, Boston Scientific, Edwards Lifesciences, and ATSens; speaker fees and consultancy fees to the institution from Biotronik, Boston Scientific, Edwards Lifesciences, Abbott, Medtronic, Biosensors, and Highlife.
**Dr. Hundertmark** reports consultancies for Alnylam Pharmaceuticals, AstraZeneca, Bayer Healthcare and Pfizer to the institution. He is supported by a 'Protected Research Time' grant from the Medical Faculty of the University of Bern, Bern, Switzerland.

**Dr. Windecker** reports research, travel and/or educational grants to the institution from Abbott, Abiomed, Alnylam, Amicus Therapeutics, Amgen, Astra Zeneca, Bayer, B.Braun, Bioanalytica, Biotronik, Boehringer Ingelheim, Boston Scientific, Bristol Myers Squibb, Cordis Medical, CorFlow Therapeutics, CSL Behring, Daiichi Sankyo, Edwards Lifesciences, Fumedica, GE Healthcare, Guerbet, IACULIS, Inari Medical, Janssen AI, Johnson & Johnson, Medalliance, Medtronic, MSD Merck Sharp & Dohme, Neovii Pharmaceutica, Neutromedics AG, Novartis, Novo Nordisk, OM Pharma, Optimapharm, Orchestra BioMed, Pfizer, Philips AG, Sanofi-Aventis, Servier, Shockwave Medical, Siemens Healthcare, Sinomed, SMT Sahajanand Medical Technologies, Vascular Medical, V-Wave. Stephan Windecker serves as advisory board member and/or member of the steering/executive group of trials funded by Abbott, Amgen, Abiomed, Edwards Lifesciences, EnCarda Inc., Medtronic, Novartis, Sinomed with payments to the institution but no personal payments. He is also a member of the steering/executive committee group of several investigator-initiated trials that receive funding by industry without impact on his personal remuneration.
**Dr. Gräni** received research funding from the GAMBIT foundation for this work. Dr. Gräni received funding from the Swiss National Science Foundation, InnoSuisse, the Center for Artificial Intelligence in Medicine, the University of Bern, the Novartis Foundation for Medical-Biological Research, the Swiss Heart Foundation, the Schmieder-Bohrisch Foundation, and the Gottfried and Julia Bangerter-Rhyner Foundation, outside of the submitted work. Further, funding to the institution was received from Alnylam Pharmaceuticals, AstraZeneca, Pfizer, and Bayer outside of the submitted work and without impacting his personal remuneration.
**Dr. Shiri** reports travel expenses from Alnylam and Bayer, as well as speaker fees and consultancy fees to the institution from Pfizer.
**All other authors** have no relationships relevant to the contents of this article to disclose.

## Ethics Declarations

This study received institutional review board approval at all in-house participating centers for different datasets (Bern: KEK 2024-01345, KEK 2021-01738, KEK 2021-0058, and KEK 2020-00841; Zurich: KEK 2015-0235 and KEK 2014-0632; Japan: no. 6970).

## Funding

This work was supported under project ID 74 as part of the Swiss AI Initiative, through a small grant from the ETH Domain and computational resources provided by the Swiss National Supercomputing Centre (CSCS) under the Alps infrastructure.

## Acknowledgments

This work was supported under project ID 74 as part of the Swiss AI Initiative, through a small grant from the ETH Domain and computational resources provided by the Swiss National Supercomputing Centre (CSCS) under the Alps infrastructure. We thank Lea Rebecca Zurbriggen from the research study team for her technical and administrative support. We thank Martin Zbinden, Bisrat Kidane, Kevin Ceni, and Juan Paolo de Torres Mirano for their technical and engineering support. We thank the Insel Data Science Center (IDSC) personnel for their support in this project.

# SUPPLEMENTARY FILE

## A Unified Framework for Comprehensive Cardiac CT Segmentation and Phenotyping: Human-in-the-Loop Data Annotation, Vision Foundation Model Development, Multicenter Evaluation and Clinical Validation

Pooya Mohammadi Kazaj[1,2,3], Leo Fridolin Weber[1,2†], Wen Xie[1,2†], Seyed Amir Ahmad Safavi-Naini[†1,2], Anselm Stark[1,2,3], Giovanni Baj[1,2], Ali Mokhtari[1,2], Toshiya Yoshida[4], Christoph Ryffel[1,2], Taishi Okuno[4], Yoshihiro Akashi[4], Ronny R Buechel[5], Thomas Pilgrim[1], Waldo Valenzuela[6], George CM Siontis[1], Xiaowei Xu[7], Moritz Hundertmark[1], Stephan Windecker[1], Christoph Gräni[1,2*], Isaac Shiri[1,2*]

1-Department of Cardiology, Inselspital, Bern University Hospital, University of Bern, Switzerland
2-Department of Digital Medicine, University of Bern, Bern, Switzerland
3-Graduate School for Cellular and Biomedical Sciences, University of Bern, Bern, Switzerland
4-Department of Cardiology, St. Marianna University School of Medicine, 2-16-1, Sugao, Miyamae-ku, Kawasaki, 216-8511, Japan
5-Department of Nuclear Medicine, Cardiac Imaging, University Hospital Zurich, Zurich, Switzerland

6- Institute for Diagnostic and Interventional Neuroradiology, Inselspital Bern, Bern, Switzerland

7-Guangdong Provincial People's Hospital, Guangzhou, China

[†] Leo Fridolin Weber, Wen Xie, Seyed Amir Ahmad Safavi-Naini contributed equally to this work.

[*] Dr. Gräni and Dr. Shiri jointly supervised this work.

*** Corresponding Author:**
Isaac Shiri, PhD
Department of Cardiology
Inselspital, Bern University Hospital
University of Bern,
Freiburgstrasse CH - 3010 Bern, Switzerland
Email: isaac.shirilord@unibe.ch

# Supplementary Note 1. Segmentation Architectures and Training Configurations

## Supplementary Note S1.1. nnU-Net

nnU-Net[1] is a self-configuring segmentation framework that automatically adapts its architecture, preprocessing, and training pipeline to the characteristics of the input dataset, making it a widely adopted baseline for medical image segmentation benchmarks. In this study, nnU-Net was deployed in its three-dimensional full-resolution configuration using a residual encoder backbone (ResEnc), which replaces standard convolutional blocks with residual connections to facilitate gradient flow in deeper networks. The model was trained with a patch size of 128 × 256 × 224 voxels at a voxel spacing of 0.5 × 0.35 × 0.35 mm$^3$, with mirroring augmentation disabled to preserve cardiac orientation consistency. A combined Dice and cross-entropy loss function was optimized using SGD with momentum and polynomial learning rate decay. Two variants were trained: one without CCT augmentation as a clean baseline, and one with CTAug applied during training to assess the contribution of artifact augmentation to robustness of segmentations, with all other training parameters held identical between variants.

## Supplementary Note S1.2. SwinUNETR

SwinUNETR[2] is a transformer-based segmentation architecture that combines a SwinTransformer encoder with a CNN-based U-Net decoder, enabling the model to capture both long-range spatial dependencies and local anatomical features. The original implementation sourced from the MONAI[3] framework was integrated into the nnUZoo pipeline, inheriting the same preprocessing, CTAug augmentation strategy, and training configuration as the other models in this study. The encoder operates on non-overlapping 3D patch tokens of size 2 × 2 × 2 voxels with an initial embedding dimension of 48, organized into four stages with depths of 2, 2, 2, and 2 transformer blocks and multi-head self-attention with 3, 6, 12, and 24 heads respectively. Attention is computed within local 3D windows of size 7 × 7 × 7 voxels using the shifted window mechanism, which enables cross-window information exchange while maintaining computational efficiency. For each window, the multi-head self-attention is computed as:

$$Attention(Q,K,V) = softmax(\frac{QK^T}{\sqrt{d_k}} + B)\,V \tag{S1}$$

where Q, K, and V are the query, key, and value matrices derived from the input tokens, $d_k$ is the per-head dimension, and B is the relative position bias matrix encoding spatial relationships between tokens within the window. The model was trained on input patches of 128 × 256 × 224 voxels at a voxel spacing of 0.5 × 0.35 × 0.35 mm$^3$.

## Supplementary Note S1.3. SwinUMamba

SwinUMamba is a Mamba-based segmentation architecture developed in this study as a three-dimensional extension of the original VMamba[4], integrated into the nnUZoo pipeline with the same preprocessing, CTAug augmentation strategy, and training configuration as the other models. The architecture follows the same hierarchical encoder-decoder design as SwinUNETR, replacing the SwinTransformer encoder with a state space model (SSM) block referred to as SS3D.

The foundation of the Mamba block is a continuous-time state space model[5] that maps an input sequence x(t) to an output y(t) through a hidden state h(t):

$$h'(t) = A\,h(t) + B\,x(t) \tag{S2}$$

$$y(t) = C\,h(t) \tag{S3}$$

where $A \in \mathbb{R}^{d_{state} \times d_{state}}$ is the state transition matrix, $B \in \mathbb{R}^{d_{state} \times 1}$ and $C \in \mathbb{R}^{1 \times d_{state}}$ are the input and output projection matrices, and $d_{state} = 16$. To enable discrete sequence processing, the continuous parameters are discretized using a timescale parameter $\Delta$ via the zero-order hold rule:

$$\bar{A} = exp(\Delta A) \tag{S4}$$

$$\bar{B} = (\Delta A)^{-1}(exp(\Delta A) - I).\Delta B \tag{S5}$$

yielding the discrete recurrence:

$$h_t = \bar{A}\,h_{t-1} + \bar{B}\,x_t \tag{S6}$$

$$y_t = C\,h_t \tag{S7}$$

The selective scan mechanism (S6) makes $B$, $C$, and $\Delta$ input-dependent[5], allowing the model to selectively retain or discard information based on content. The time-step $\Delta$ is projected using a rank-reduced matrix of rank $dt_{rank} = \lceil \frac{d_{model}}{16} \rceil$, and the inner feature dimension is expanded as $d_{inner} = 2 \times d_{model}$ before entering the SSM to increase representational capacity.

SS3D extends the two-dimensional cross-scan strategy of VMamba to full volumetric feature extraction by applying six independent scanning trajectories along the width, height, and depth dimensions in both forward and backward directions (w+, w−, h+, h−, z+, z−), enabling comprehensive spatial context modeling across all three axes. For each scanning direction, the SSM projects the inner features to the state space using:

$$x_{proj} : d_{inner} \rightarrow (dt_{rank} + d_{state} \times 2) \quad \text{(S8)}$$

The selective scan is computed using the optimized CUDA kernel from the mamba-ssm[5] library (selective_scan_fn). The encoder is organized into four hierarchical stages with feature dimensions of 48, 96, 192, and 384, depths of 2, 2, 9, and 2 Mamba blocks, and a hidden size of 768 at the deepest stage, with a patch token size of 2 × 2 × 2 voxels and kernel size of 7. The decoder employs UnetrBasicBlock convolutional blocks for three-dimensional feature extraction, mirroring the SwinUNETR decoder design, with deep supervision applied across decoder stages to improve gradient flow during training. The model was trained on input patches of 128 × 256 × 224 voxels at a voxel spacing of 0.5 × 0.35 × 0.35 $mm^3$.

## Supplementary Note S1.4. $U^2$-Net

$U^2$-Net[6] is a multi-scale segmentation architecture originally proposed for two-dimensional salient object detection, extended to three-dimensional volumetric segmentation in this study within the nnUZoo pipeline. The key architectural innovation of $U^2$-Net is the Residual U-block (RSU), a nested U-Net structure where each encoder and decoder block is itself a full U-Net with dilated convolutions, enabling multi-scale feature extraction at every stage without increasing computational cost through dilation. The three-dimensional extension replaces all convolutional operations within the RSU blocks with their 3D counterparts while preserving the original block design and dilation rates. The architecture consists of six encoder and five decoder stages connected by skip connections, with an additional RSU bridge block, totaling 11 RSU blocks across the network. Two variants were evaluated: $U^2$-Net Small with encoder feature dimensions of 64 and decoder feature dimensions of 128 across all stages, and $U^2$-Net Large with progressively expanding encoder feature dimensions of 64, 128, 256, 512, 512, and 512 and corresponding decoder dimensions of 128, 256, 512, 1024, and 1024 to increase representational capacity at deeper stages. Both variants employed deep supervision across decoder stages and were trained with a combined Dice and cross-entropy loss on input patches of 128 × 256 × 224 voxels at a voxel spacing of 0.5 × 0.35 × 0.35 $mm^3$.

## Supplementary Note S2. Carbon Emission Calculation

Carbon dioxide equivalent ($CO_2e$) emissions were estimated for model training and inference using the Green Algorithms framework (Lannelongue et al., Advanced Science, 2021), as $CO_2e$ (kg) = t × n_GPU × TDP × PUE × CI × $10^{-6}$, where t is wall-clock runtime (hours), n_GPU the number of GPUs, TDP the GPU thermal design power, PUE the facility power usage effectiveness (set to 1.1 to be comparable with other foundation-model studies, e.g. DINOv2[7]), and CI the grid carbon intensity. Self-supervised pre-training was a one-time cost (50 GPU-hours on a single NVIDIA H200 NVL 140GB, TDP = 600 W); each of the six benchmark models was then trained on a single NVIDIA A100 80GB PCIe (TDP = 300 W) for 1000 epochs. Because $CO_2e$ scales linearly with grid carbon intensity, each task is reported under four grid scenarios drawn from a single source (Ember, 2023 values): Switzerland (41 $gCO_2$/kWh, the low-carbon grid on which the models were trained), the EU-27 (242), the United States (369), and the global average (480) (**Supplementary Figure S42**). Inference emissions are reported per patient. The Swiss intensity is production/territorial-based; a consumption-based value (accounting for imports) would be higher. Estimates capture GPU operational (Scope-2) energy only and exclude the CPU/RAM term of the full Green Algorithms model and embodied hardware emissions, so they should be read as approximate rather than as a strict upper bound.

On the low-carbon Swiss grid on which the models were trained (CI = 41 $gCO_2e$/kWh), one-time self-supervised pre-training incurred 1.35 kg $CO_2e$ (50 GPU-hours on an NVIDIA H200 NVL 140GB), training each of the six benchmark models on a single NVIDIA A100 80GB PCIe emitted between 0.34 kg (Nnunet Wo CT-Aug) and 2.44 kg $CO_2e$ (SwinUMamba W CT-Aug), and inference cost 0.18–0.22 g $CO_2e$ per patient (**Supplementary Figure S42**). Emissions were modest across all architectures, with nnU-Net the most efficient and the transformer- and Mamba-based models incurring higher cost; identical computation on more carbon-intensive grids would produce several-fold higher emissions (up to ≈12× at the global average). For context, Selvan et al. reported 11.43 kg $CO_2e$ in total for training nnU-Net across three medical-imaging datasets in Denmark, and Prajwal et al. 0.48–0.64 kg $CO_2e$ for U-Net training on KiTS-19 in California, with absolute differences driven largely by grid carbon intensity and training duration rather than by the models themselves. To make these magnitudes tangible, pre-training plus training all six models on the Swiss grid (≈11 kg $CO_2e$ in total) is equivalent to roughly 63 km driven in an average passenger car, less than the 17.5 kg $CO_2e$ emitted by the single 100 km car trip shown for reference in **Supplementary Figure S42** (car factor ≈175 $gCO_2e$/km; Green Algorithms[8]).

## Supplementary Note S3. Full method of information gain on clinical and laboratory data

Using CCT imaging data from an aortic stenosis cohort, together with clinical parameters (up to 1,962 patients; Inselspital Bern and Japan), we tested whether segmentation-derived substructure volumes convey information beyond routine demographics regarding clinical and laboratory parameters. Three primary outcomes were pre-specified before unblinded analysis, left-ventricular ejection fraction (LVEF), brain natriuretic peptide (BNP), and the mean trans-aortic gradient. The remaining fourteen targets (creatinine, eGFR, hemoglobin, platelets, creatine kinase, CK-MB, albumin, leucocytes, body mass index, EuroSCORE II, logistic EuroSCORE, STS Score, AVA Index, and NYHA class) were declared exploratory.

Patients with implausible body-surface area (outside 1.2 to 2.5 $m^2$) were excluded from all clinical-variable analyses, and targets with fewer than 150 complete cases after this exclusion were dropped. Albumin values exceeding 60 g/L were biologically implausible (2 of 1,813 cases) and were set to missing before analysis. Creatine kinase and CK-MB were log1p-transformed before modeling to handle right skew and the CK-MB assay detection-limit floor of 0.5 µg/L, so all reported $\Delta R^2$ and SHAP-η for these two targets are on the log-transformed scale.

For each target, we fitted two nested CatBoost regression models under identical five-fold cross-validation. The Baseline model received age, sex, and body-surface area (3 features), and the Full model additionally received the eight raw substructure voxel volumes (11 features). Inter-chamber volume ratios were excluded from every feature set, because each ratio is an algebraic transform of two raw volumes already present and is therefore collinear with, and redundant for, tree-based learners. CatBoost was run with depth 6, learning rate 0.05, L2 leaf regularisation 3, up to 600 boosting iterations, and early stopping after 30 rounds without improvement on a 15% inner-validation slice of the training fold; thus, the held-out test fold was never used for model selection. Out-of-fold predictions were pooled across the five folds before computing $R^2$. A linear sensitivity baseline using RidgeCV with the α grid {0.01, 0.1, 1, 10, 100} was fitted under the same cross-validation to test whether the gains arose from non-linear interactions (**Supplementary Figure S40.**a).

For each target, we report $\Delta R^2 = R^2_full - R^2_baseline$ with a paired-patient bias-corrected accelerated (BCa) bootstrap 95% confidence interval (1,000 resamples, leave-one-out jackknife for the acceleration term). Two-sided bootstrap p-values were derived from the bootstrap-$\Delta R^2$ distribution as $2 \cdot \min(\Pr[\Delta R^2_b \leq 0], \Pr[\Delta R^2_b \geq 0])$. Benjamini-Hochberg FDR control at $q = 0.05$ was applied separately within two pre-specified families, the three primary outcomes and the fourteen exploratory outcomes. A parametric per-fold paired t-test on $\Delta R^2$ across the five outer folds (degrees of freedom = 4) was reported as a parametric cross-check, and a primary finding was considered confirmed when both the BH-adjusted bootstrap test and the BH-adjusted per-fold paired t-test were significant at $q = 0.05$..

To establish that the incremental signal was cardiac rather than demographic, we added a third, volume-only model that received the eight raw substructure volumes with no demographic input (8 features), fitted under the same cross-validation, and we report its out-of-fold $R^2$ with a percentile bootstrap 95% confidence interval alongside the Baseline and Full models (**Supplementary Figure S39.**a). For body-mass index, which is recoverable from the body-surface area already in the baseline, we fitted four nested models to separate body size from any cardiac contribution, M1 (age, sex), M2 (age, sex, body-surface area), M3 (M2 plus the eight volumes), and M4 (age, sex, and the eight volumes with body-surface area removed) (**Supplementary Figure S39**.b).

Per-feature contribution was summarised by the SHAP-η ratio, $\eta_j = \sqrt{(\mathrm{Var}(\varphi_j)/\mathrm{Var}(y))}$, computed patient-wise from CatBoost's native TreeSHAP on each held-out fold. This statistic is non-negative and directly comparable across targets with different units (it remained below 1 for every feature and target in this analysis). We report η in three complementary views. In the pairwise views, each predictor was fitted on its own, first without adjustment (**Supplementary Figure S36.a**) and then with age and sex included as covariates (**Supplementary Figure S36.b**), the three demographics and the eight body-surface-area-indexed volumes served in turn as the predictor of interest, which isolates the marginal signal of each structure. In the joint view, η was computed for every feature inside the eleven-feature Full model, using the raw volumes, which shows what each structure contributes once all features compete (**Supplementary Figure S36.c**). Calibration of the Full model on each primary outcome was assessed by ordinary least-squares regression of observed on predicted out-of-fold values, reporting the calibration slope, intercept, and out-of-fold $R^2$ (**Supplementary Figure S40.b**).

All analyses were implemented in Python 3.10.19 with CatBoost 1.2.10, scikit-learn 1.7.2, statsmodels 0.14.6, NumPy 1.26.4, and SciPy 1.15.3; SHAP values were obtained from CatBoost's native TreeSHAP. All cross-validation splits, CatBoost training, and bootstrap resampling used a fixed random seed of 42. An optional nested hyperparameter search for the three primary outcomes (depth 4 or 6; learning rate 0.03, 0.05, or 0.10; L2 leaf regularisation 1, 3, or 5; applied only to the Full model and selected per outer fold by the highest inner three-fold cross-validation $R^2$) is implemented but was not enabled in this analysis. All reported models (the Baseline, Full, and volume-only models across the 17 targets) therefore used the single fixed CatBoost configuration given above, so the main incremental-value figure and **Supplementary Figure S39**, **Supplementary Figure S36**, and **Supplementary Figure S40** all reflect that configuration.

# Supplementary Figures

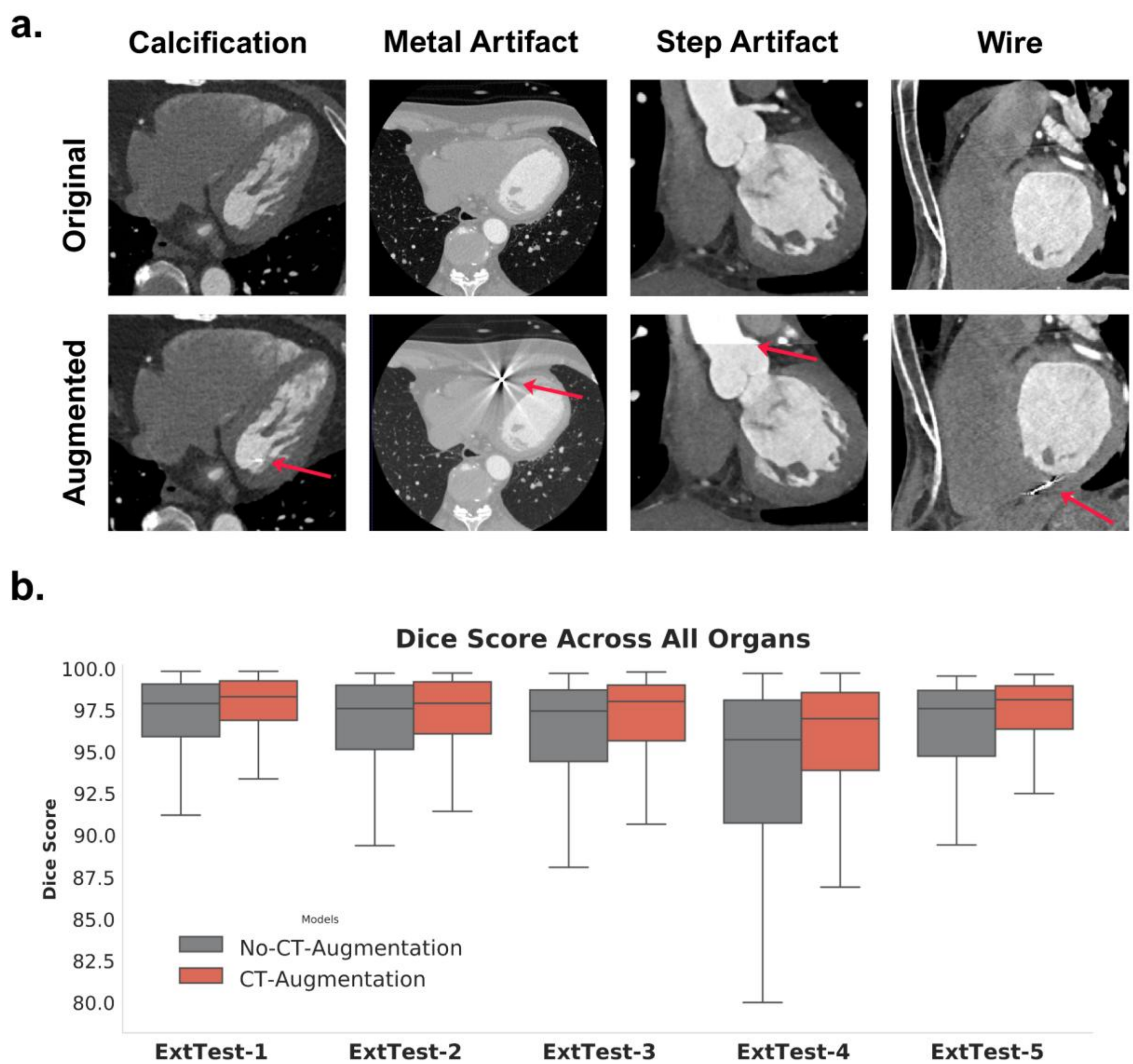

**Supplementary Figure S1.** Cardiac CT augmentation (CTAug) and its effect on segmentation robustness. (a) Representative examples of the four simulated artifact categories (calcification, metal artifact, step artifact, and wire) shown as original (top) and augmented (bottom) image pairs; red arrows indicate the introduced artifacts. (b) Dice score across all structures on the five external test datasets for models trained without (gray) versus with (red) CTAug, showing consistent improvement, most notably on the artifact-rich ExtTest-4 cohort.

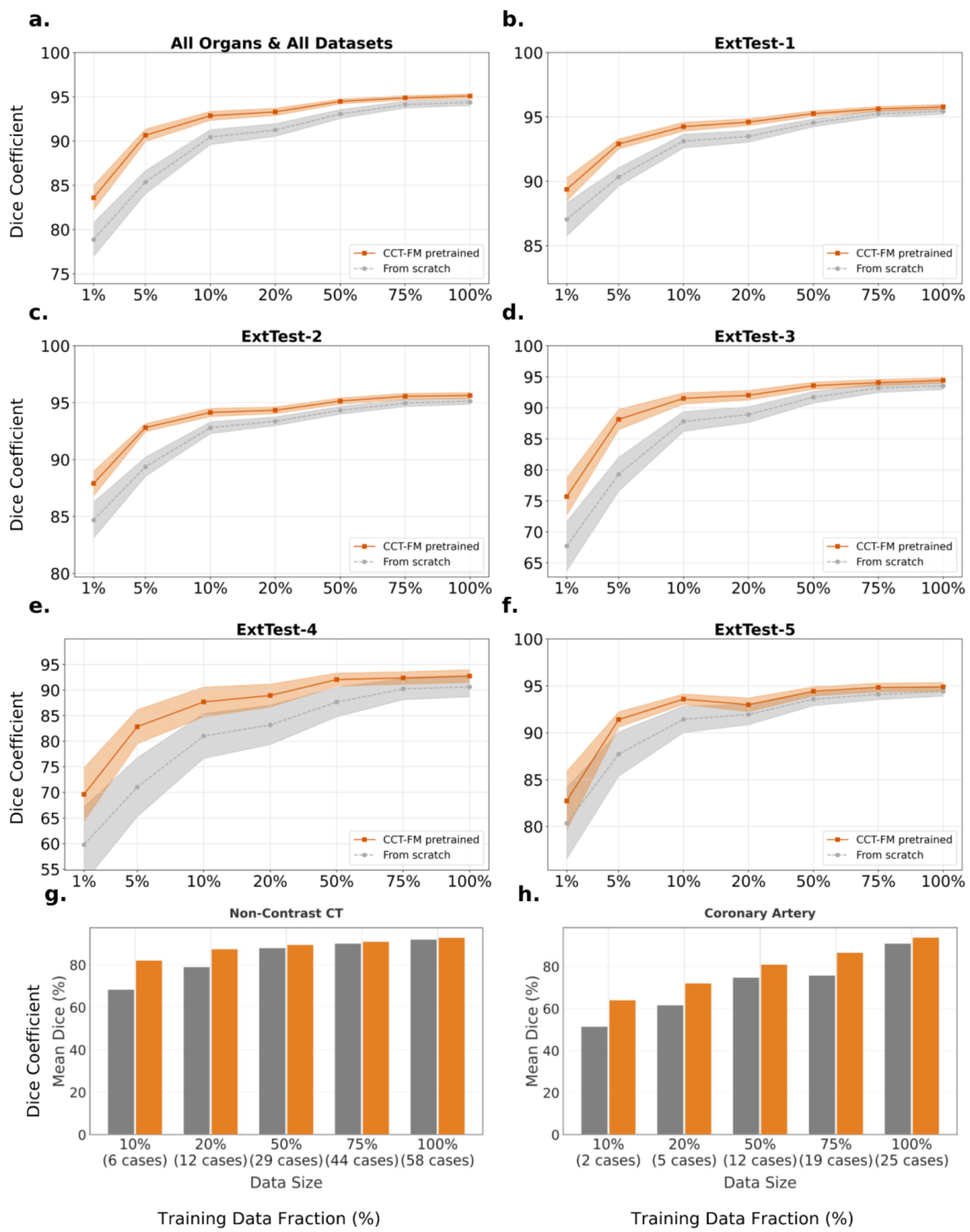

**Supplementary Figure S2. Data efficiency of self-supervised pre-training across datasets and downstream tasks, in female sex.** Mean Dice coefficient as a function of labeled training data fraction (1–100%) for models initialized from CCT-FM pre-trained weights (orange) versus trained from scratch (gray), aggregated across datasets. Shaded bands denote 95% confidence intervals. (a) All organs across all datasets micro-average (N = 235) and (b–f) each external test dataset individually (ExtTest-1: N = 80; ExtTest-2: N = 76; ExtTest-3: N = 52; ExtTest-4: N = 27; ExtTest-5: N = 20). (g–h) Mean Dice on the downstream non-contrast CT and coronary anomaly tasks (Non-Contrast CT: N = 1; Coronary Artery: N = 2). ExtTest-5 lacks sex metadata and is shown in aggregate (all subjects); two ExtTest-4 cases without sex metadata were excluded.

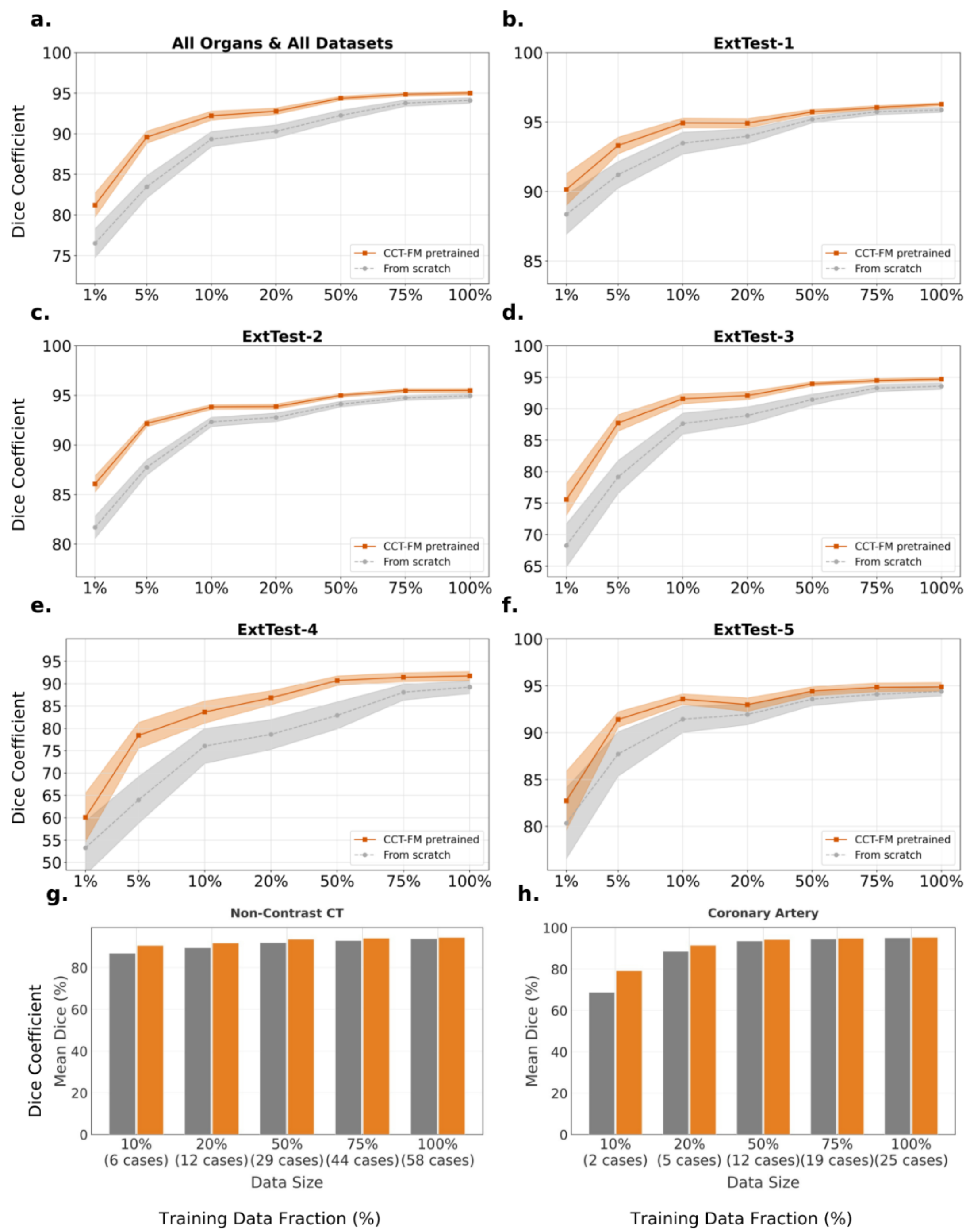

**Supplementary Figure S3. Data efficiency of self-supervised pre-training across datasets and downstream tasks, in male sex.** Mean Dice coefficient as a function of labeled training data fraction (1–100%) for models initialized from CCT-FM pre-trained weights (orange) versus trained from scratch (gray), aggregated across datasets. Shaded bands denote 95% confidence intervals. (a) All organs across all datasets micro-average (N = 341) and (b–f) each external test dataset individually (ExtTest-1: N = 93; ExtTest-2: N = 122; ExtTest-3: N = 80; ExtTest-4: N = 46; ExtTest-5: N = 20). (g–h) Mean Dice on the downstream non-contrast CT and coronary anomaly tasks (Non-Contrast CT: N = 5; Coronary Artery: N = 2). ExtTest-5 lacks sex metadata and is shown in aggregate (all subjects); two ExtTest-4 cases without sex metadata were excluded.

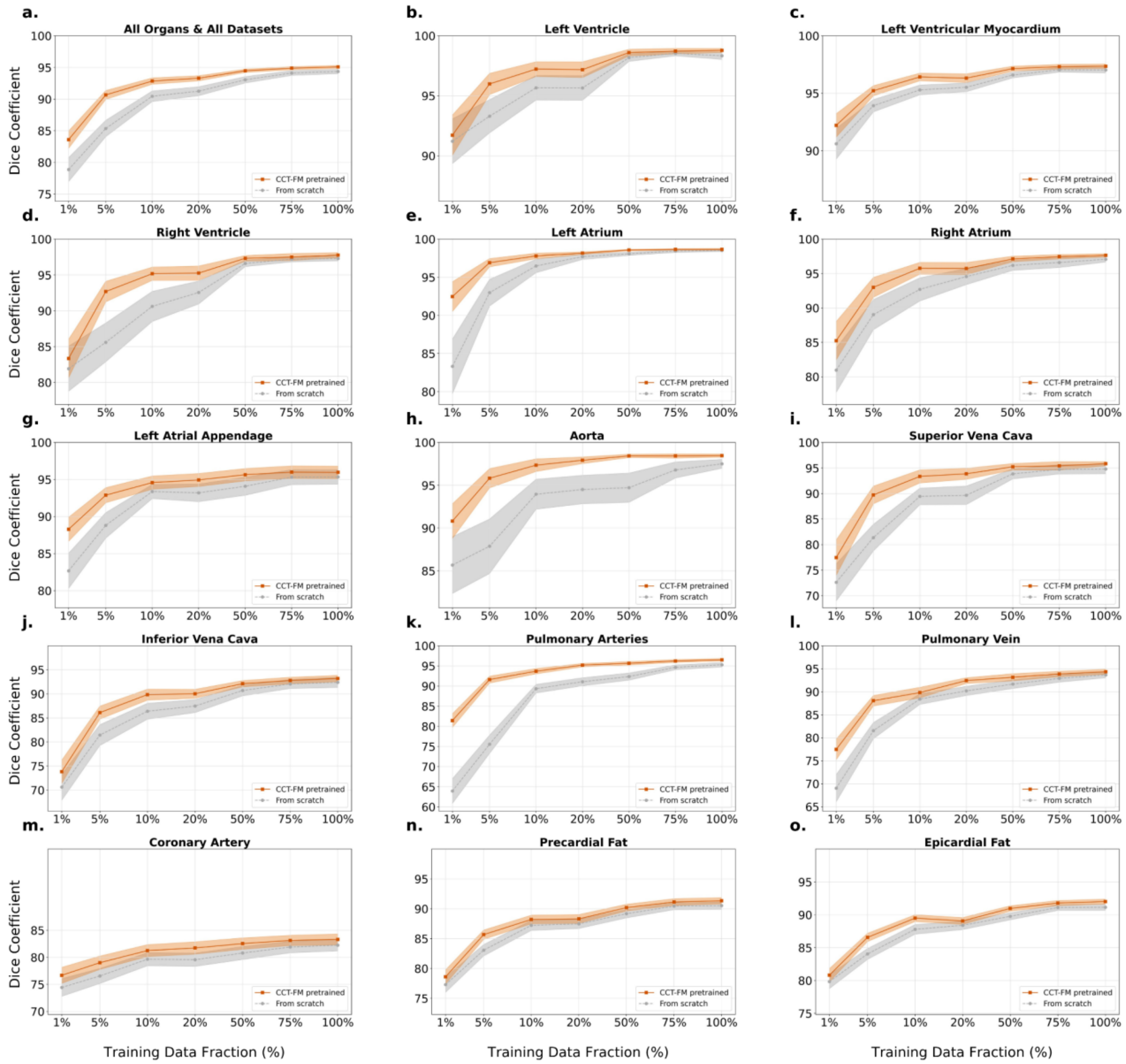

**Supplementary Figure S4. Per-structure data efficiency of self-supervised pre-training in female patients, across all datasets.** Mean Dice coefficient, reported as a micro-average across external datasets (N = 235) and as a function of labeled training data fraction (1–100%), for CCT-FM pre-trained (orange) versus from-scratch (gray) models. Results are shown for **(a)** all organs combined and **(b–o)** each of the 14 cardiac structures individually; shaded bands denote 95% confidence intervals. ExtTest-5 and two cases from ExtTest-4 lack sex metadata and were excluded.

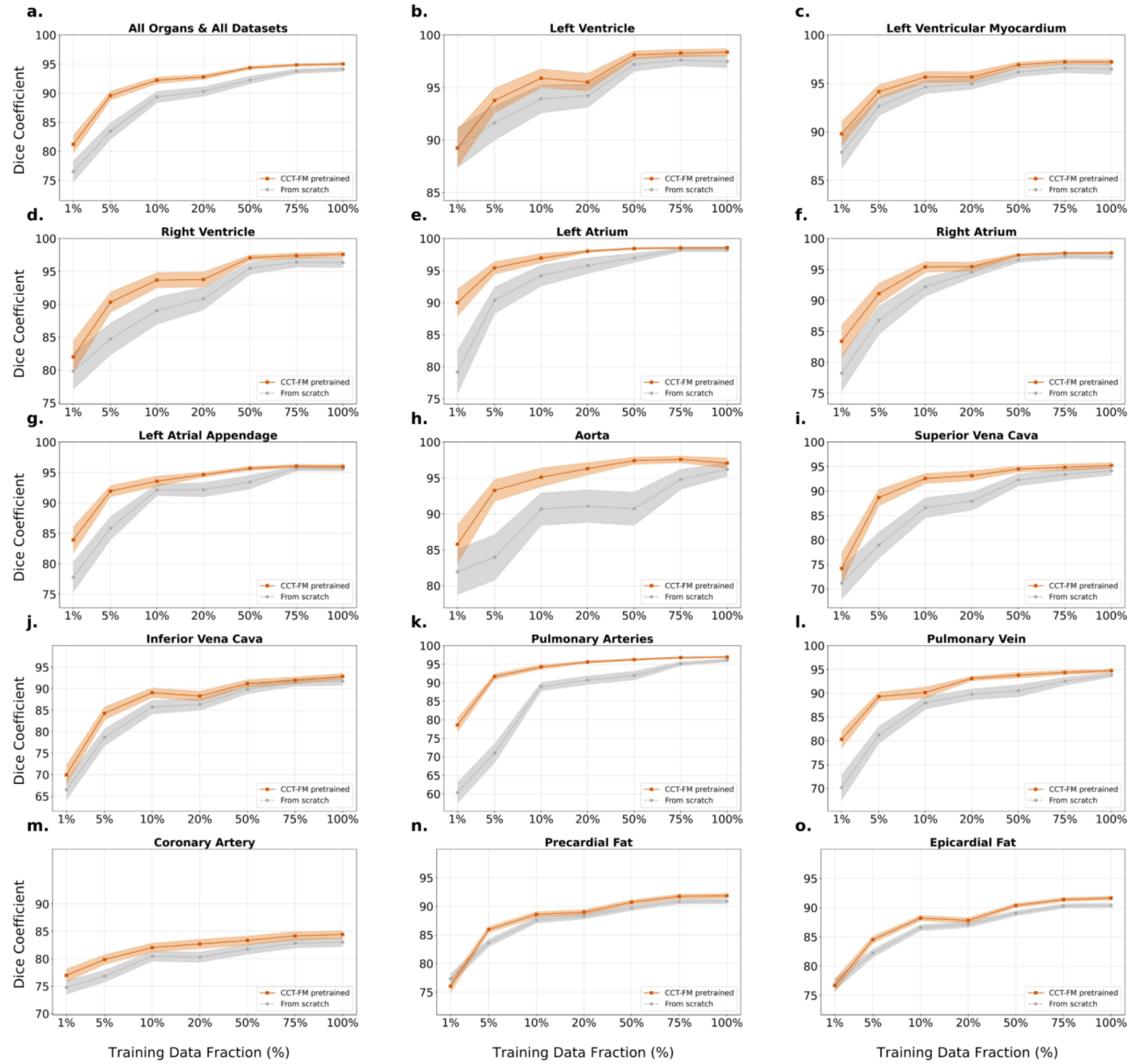

**Supplementary Figure S5. Per-structure data efficiency of self-supervised pre-training in male patients, across all datasets.** Mean Dice coefficient, reported as a micro-average across external datasets (N = 341) and as a function of labeled training data fraction (1–100%), for CCT-FM pre-trained (orange) versus from-scratch (gray) models. Results are shown for (a) all organs combined and (b–o) each of the 14 cardiac structures individually; shaded bands denote 95% confidence intervals. ExtTest-5 and two cases from ExtTest-4 lack sex metadata and were excluded.

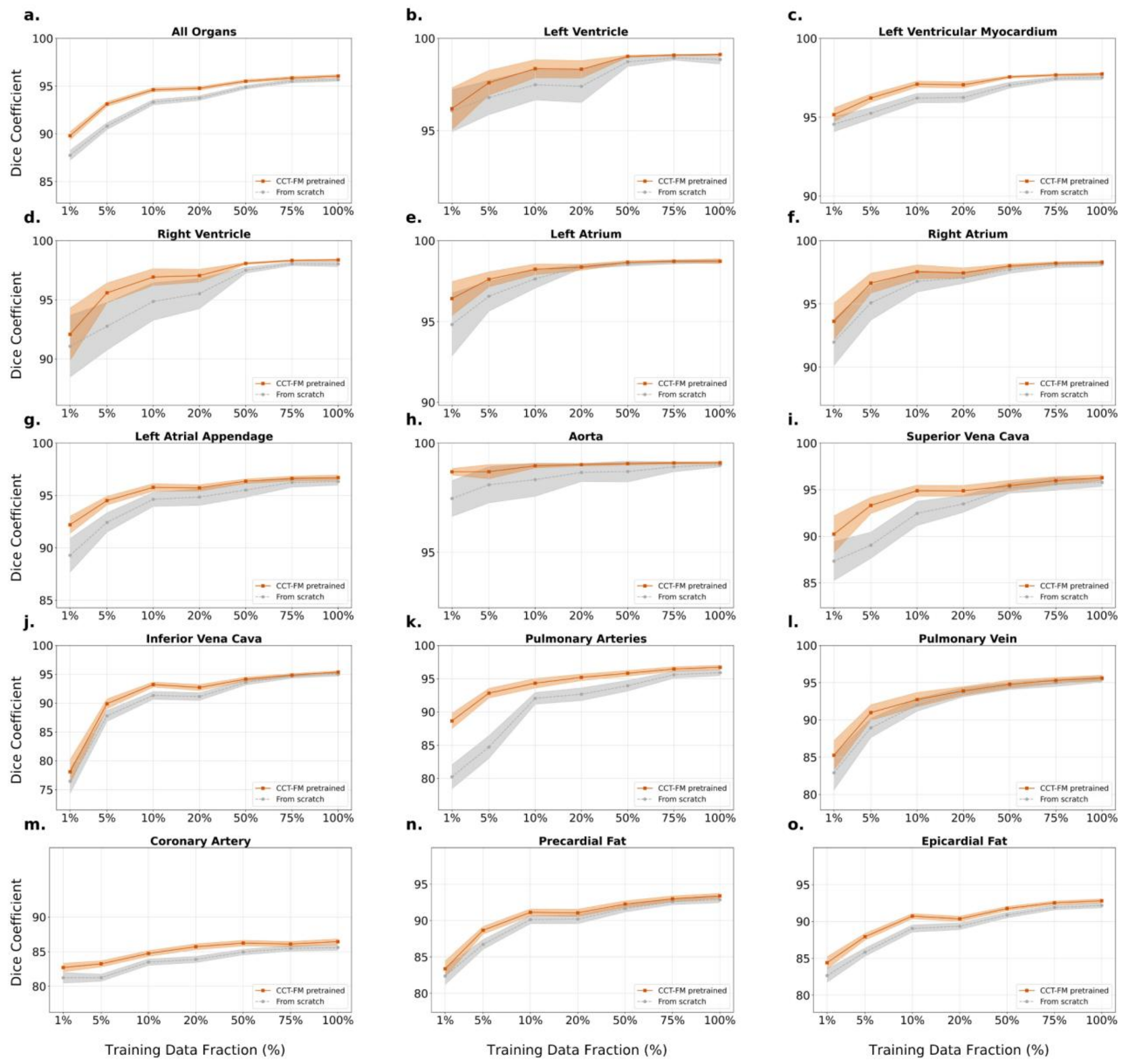

**Supplementary Figure S6.Per-structure data efficiency of self-supervised pre-training in all patients, across the ExtTest-1 dataset.** Mean Dice coefficient, reported as a micro-average across external datasets (N = 173) and as a function of labeled training data fraction (1–100%), for CCT-FM pre-trained (orange) versus from-scratch (gray) models. Results are shown for (a) all organs combined and (b–o) each of the 14 cardiac structures individually; shaded bands denote 95% confidence intervals.

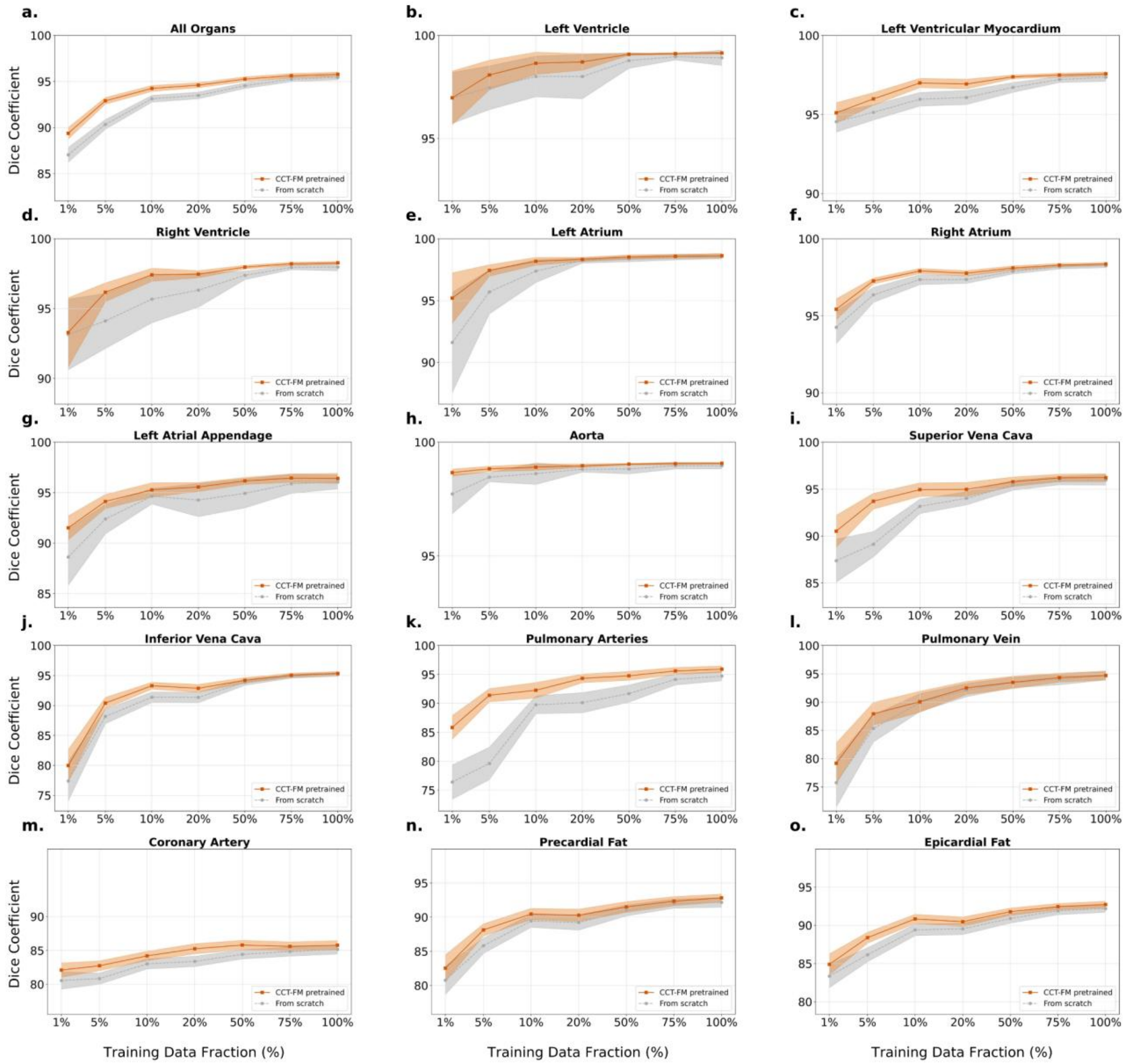

**Supplementary Figure S7. Per-structure data efficiency of self-supervised pre-training in female patients, across the ExtTest-1 dataset.** Mean Dice coefficient, reported as a micro-average across external datasets (N = 80) and as a function of labeled training data fraction (1–100%), for CCT-FM pre-trained (orange) versus from-scratch (gray) models. Results are shown for (a) all organs combined and (b–o) each of the 14 cardiac structures individually; shaded bands denote 95% confidence intervals.

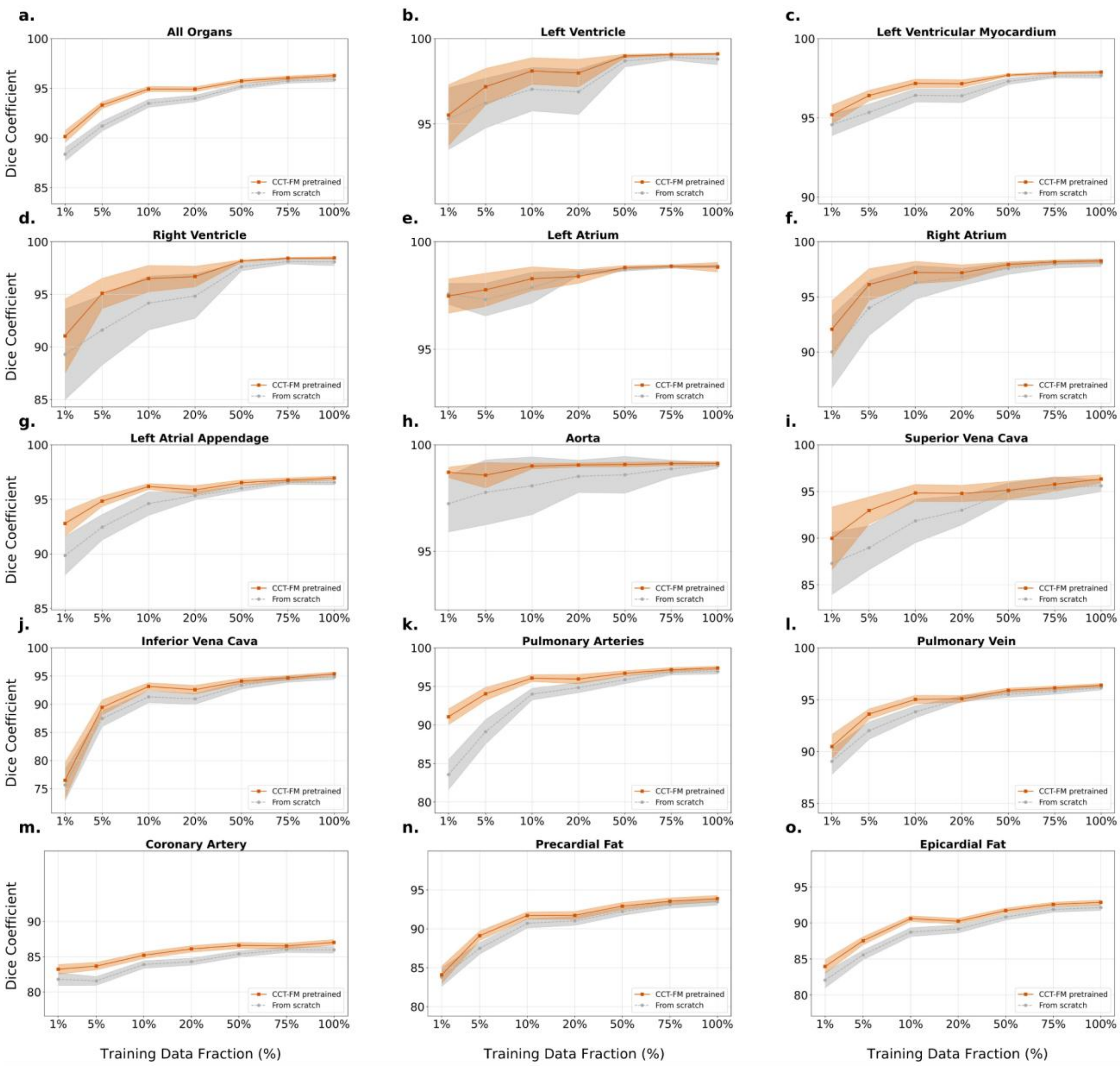

**Supplementary Figure S8. Per-structure data efficiency of self-supervised pre-training in male patients, across the ExtTest-1 dataset.** Mean Dice coefficient, reported as a micro-average across external datasets (N = 93) and as a function of labeled training data fraction (1–100%), for CCT-FM pre-trained (orange) versus from-scratch (gray) models. Results are shown for (a) all organs combined and (b–o) each of the 14 cardiac structures individually; shaded bands denote 95% confidence intervals.

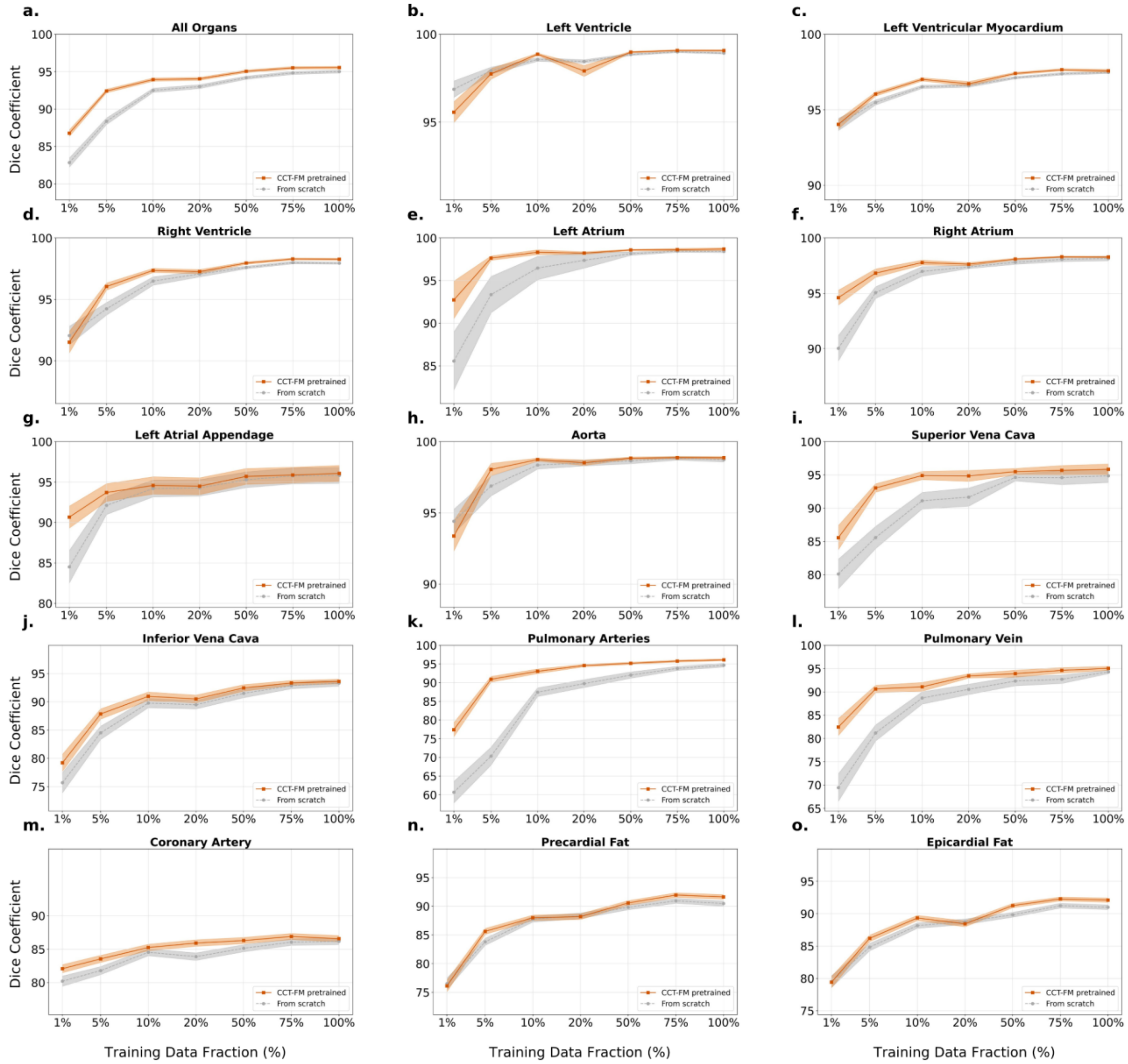

**Supplementary Figure S9. Per-structure data efficiency of self-supervised pre-training in all patients, across the ExtTest-2 dataset.** Mean Dice coefficient, reported as a micro-average across external datasets (N = 198) and as a function of labeled training data fraction (1–100%), for CCT-FM pre-trained (orange) versus from-scratch (gray) models. Results are shown for (a) all organs combined and (b–o) each of the 14 cardiac structures individually; shaded bands denote 95% confidence intervals.

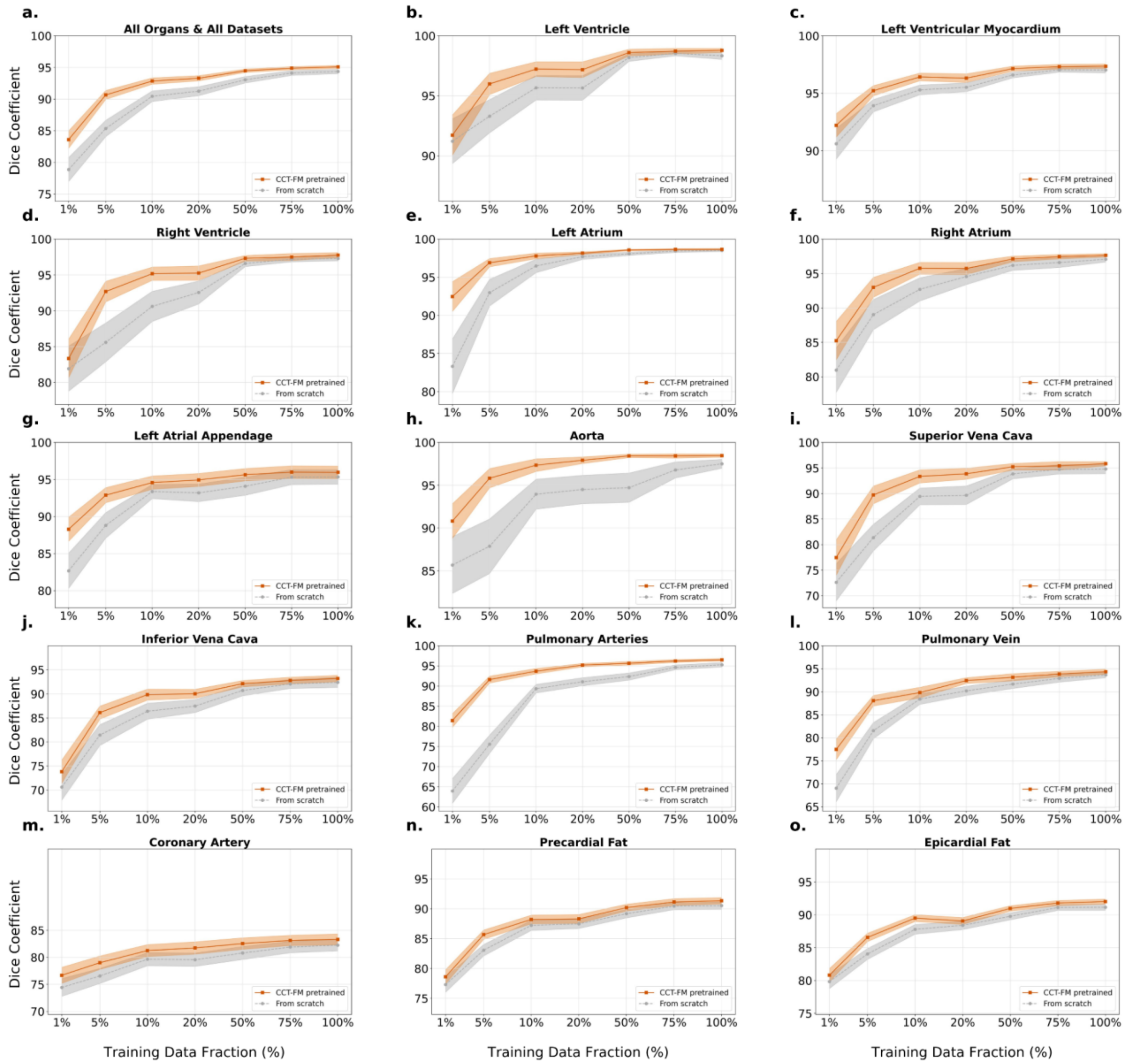

**Supplementary Figure S10. Per-structure data efficiency of self-supervised pre-training in female patients, across the ExtTest-2 dataset.** Mean Dice coefficient, reported as a micro-average across external datasets (N = 76) and as a function of labeled training data fraction (1–100%), for CCT-FM pre-trained (orange) versus from-scratch (gray) models. Results are shown for (a) all organs combined and (b–o) each of the 14 cardiac structures individually; shaded bands denote 95% confidence intervals.

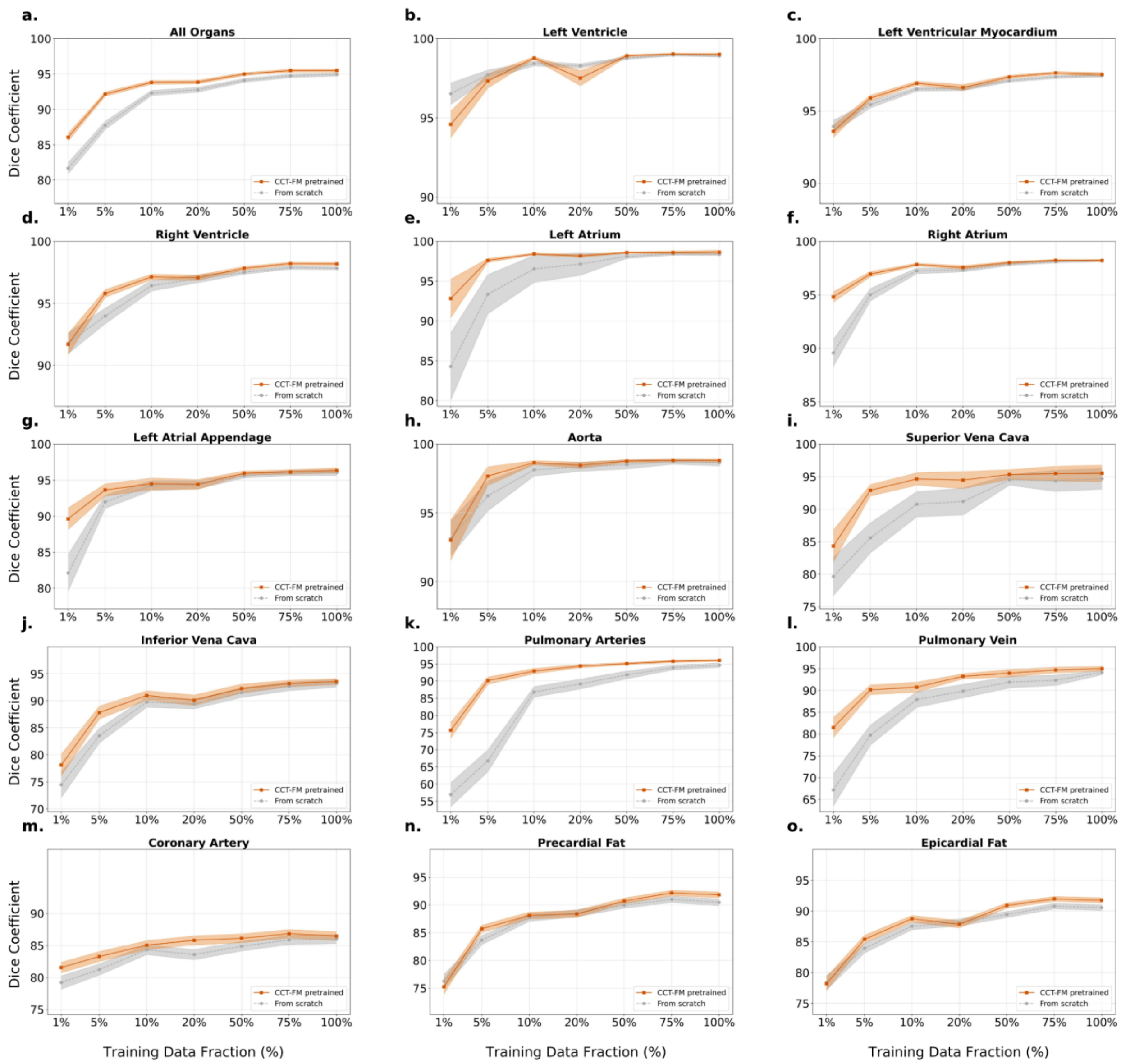

**Supplementary Figure S11.Per-structure data efficiency of self-supervised pre-training in male patients, across the ExtTest-2 dataset.** Mean Dice coefficient, reported as a micro-average across external datasets (N = 122) and as a function of labeled training data fraction (1–100%), for CCT-FM pre-trained (orange) versus from-scratch (gray) models. Results are shown for (a) all organs combined and (b–o) each of the 14 cardiac structures individually; shaded bands denote 95% confidence intervals.

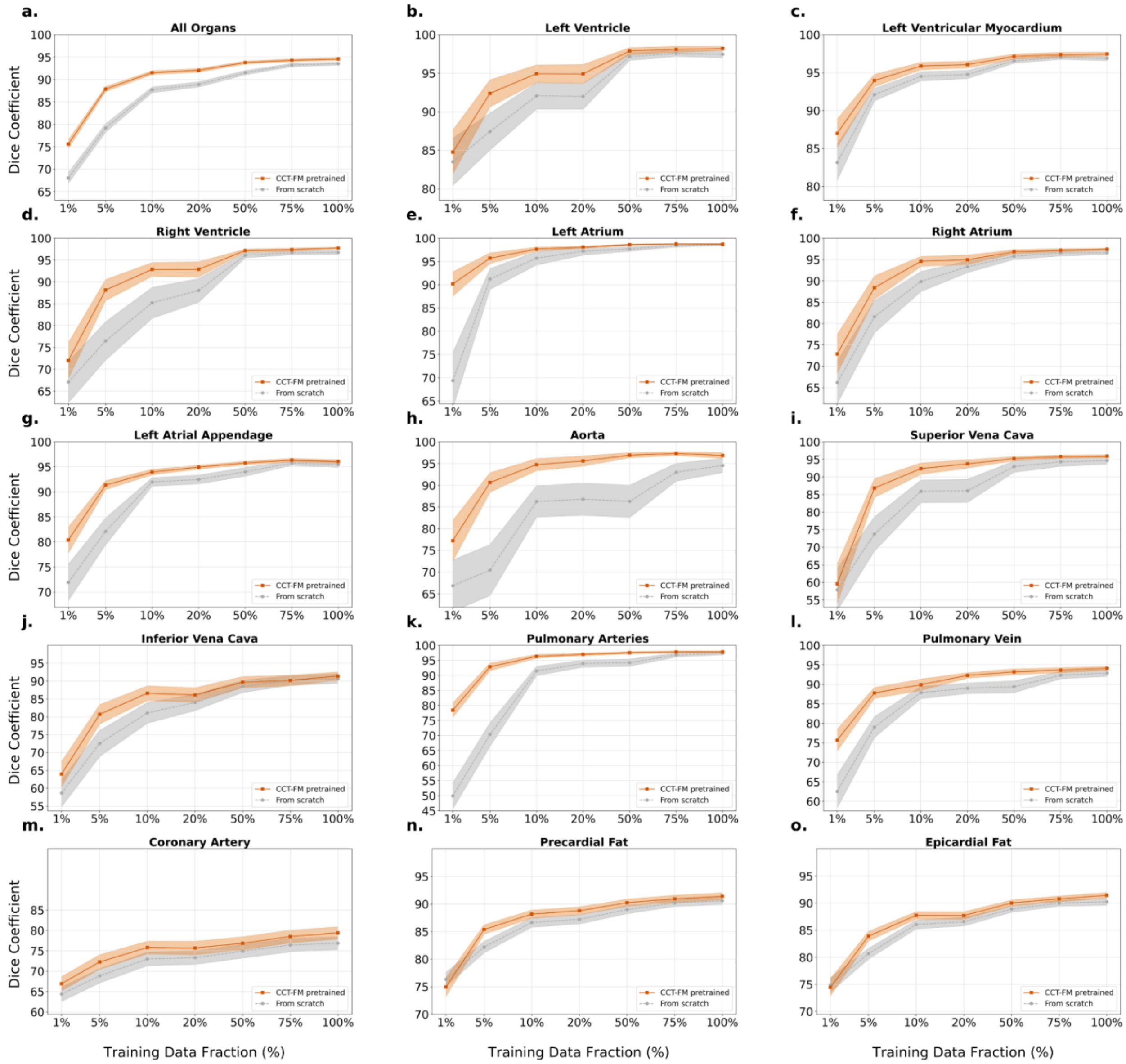

**Supplementary Figure S12. Per-structure data efficiency of self-supervised pre-training in all patients, across the ExtTest-3 dataset.** Mean Dice coefficient, reported as a micro-average across external datasets (N = 132) and as a function of labeled training data fraction (1–100%), for CCT-FM pre-trained (orange) versus from-scratch (gray) models. Results are shown for (a) all organs combined and (b–o) each of the 14 cardiac structures individually; shaded bands denote 95% confidence intervals.

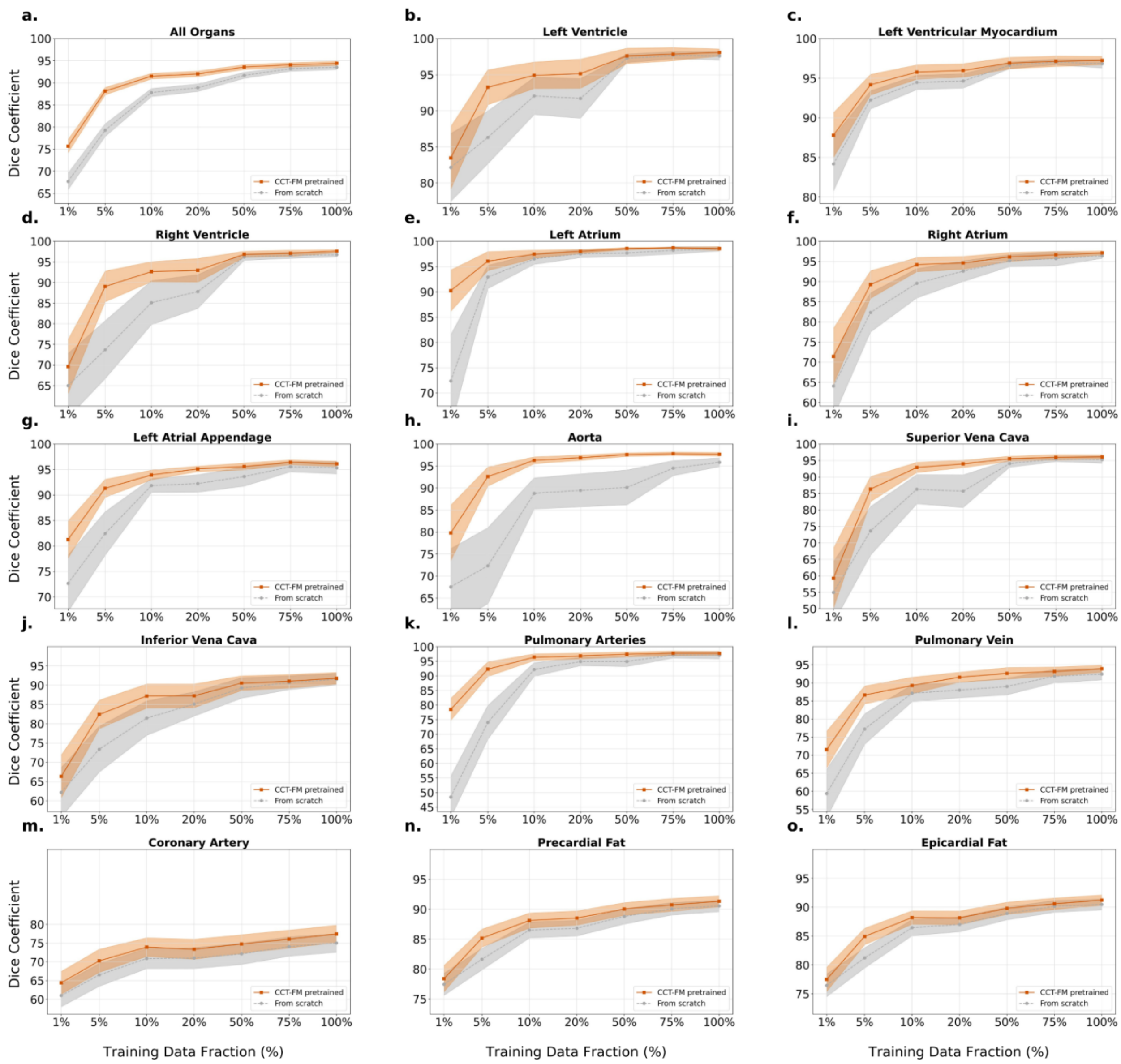

**Supplementary Figure S13. Per-structure data efficiency of self-supervised pre-training in female patients, across the ExtTest-3 dataset.** Mean Dice coefficient, reported as a micro-average across external datasets (N = 52) and as a function of labeled training data fraction (1–100%), for CCT-FM pre-trained (orange) versus from-scratch (gray) models. Results are shown for (a) all organs combined and (b–o) each of the 14 cardiac structures individually; shaded bands denote 95% confidence intervals.

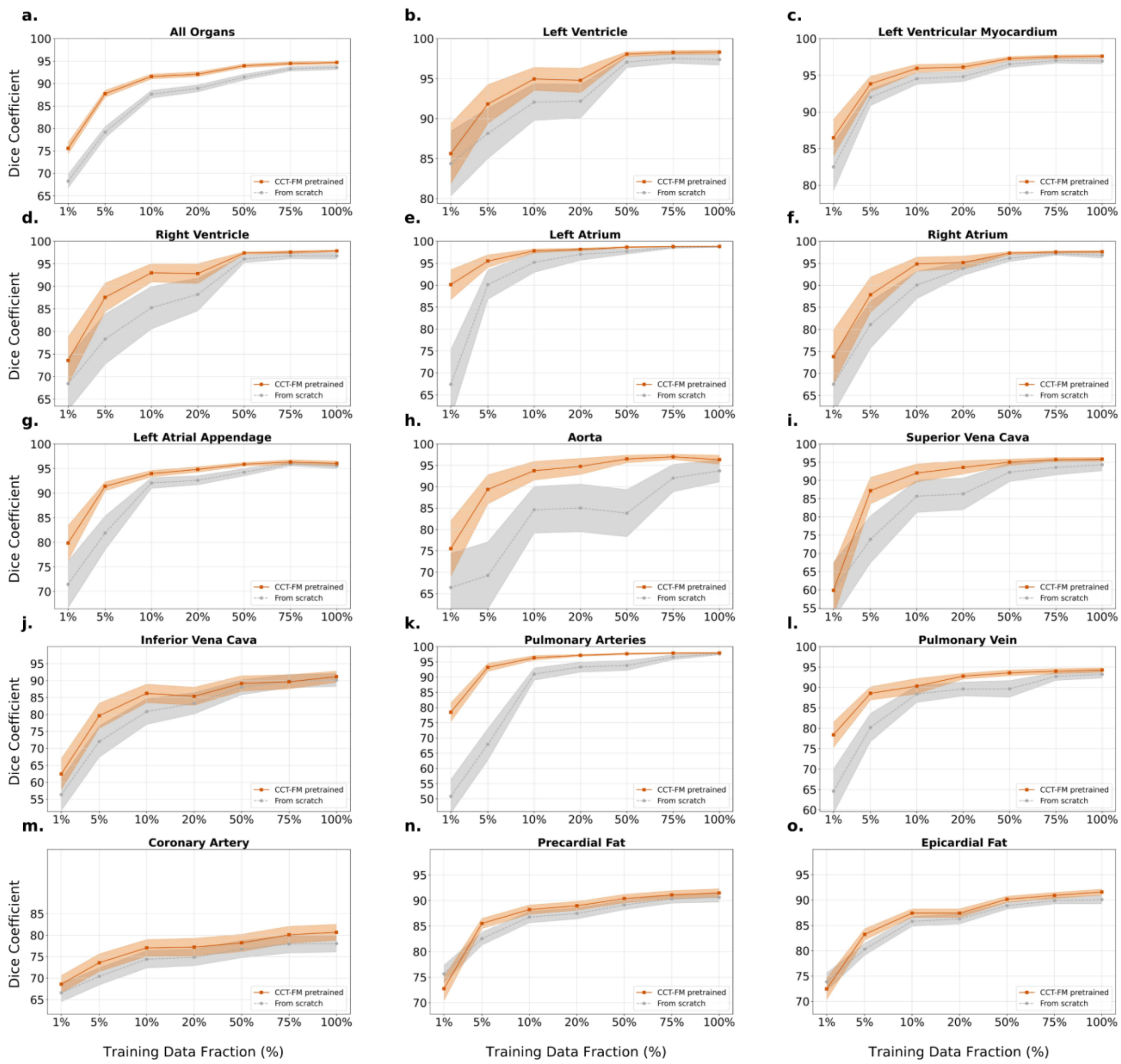

**Supplementary Figure S14. Per-structure data efficiency of self-supervised pre-training in male patients, across the ExtTest-3 dataset.** Mean Dice coefficient, reported as a micro-average across external datasets (N = 80) and as a function of labeled training data fraction (1–100%), for CCT-FM pre-trained (orange) versus from-scratch (gray) models. Results are shown for (a) all organs combined and (b–o) each of the 14 cardiac structures individually; shaded bands denote 95% confidence intervals.

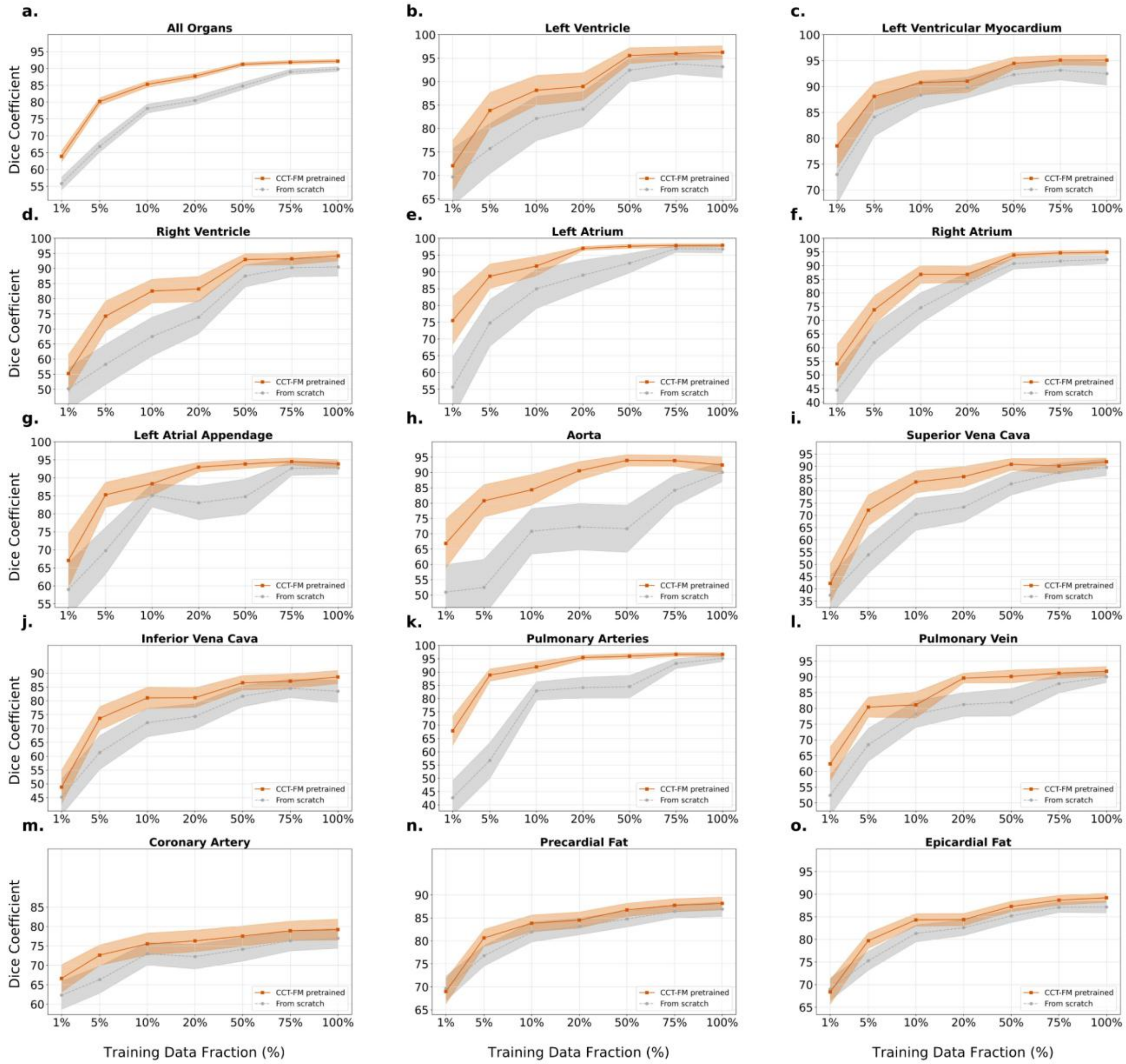

**Supplementary Figure S15. Per-structure data efficiency of self-supervised pre-training in all patients, across the ExtTest-4 dataset.** Mean Dice coefficient, reported as a micro-average across external datasets (N = 75) and as a function of labeled training data fraction (1–100%), for CCT-FM pre-trained (orange) versus from-scratch (gray) models. Results are shown for (a) all organs combined and (b–o) each of the 14 cardiac structures individually; shaded bands denote 95% confidence intervals.

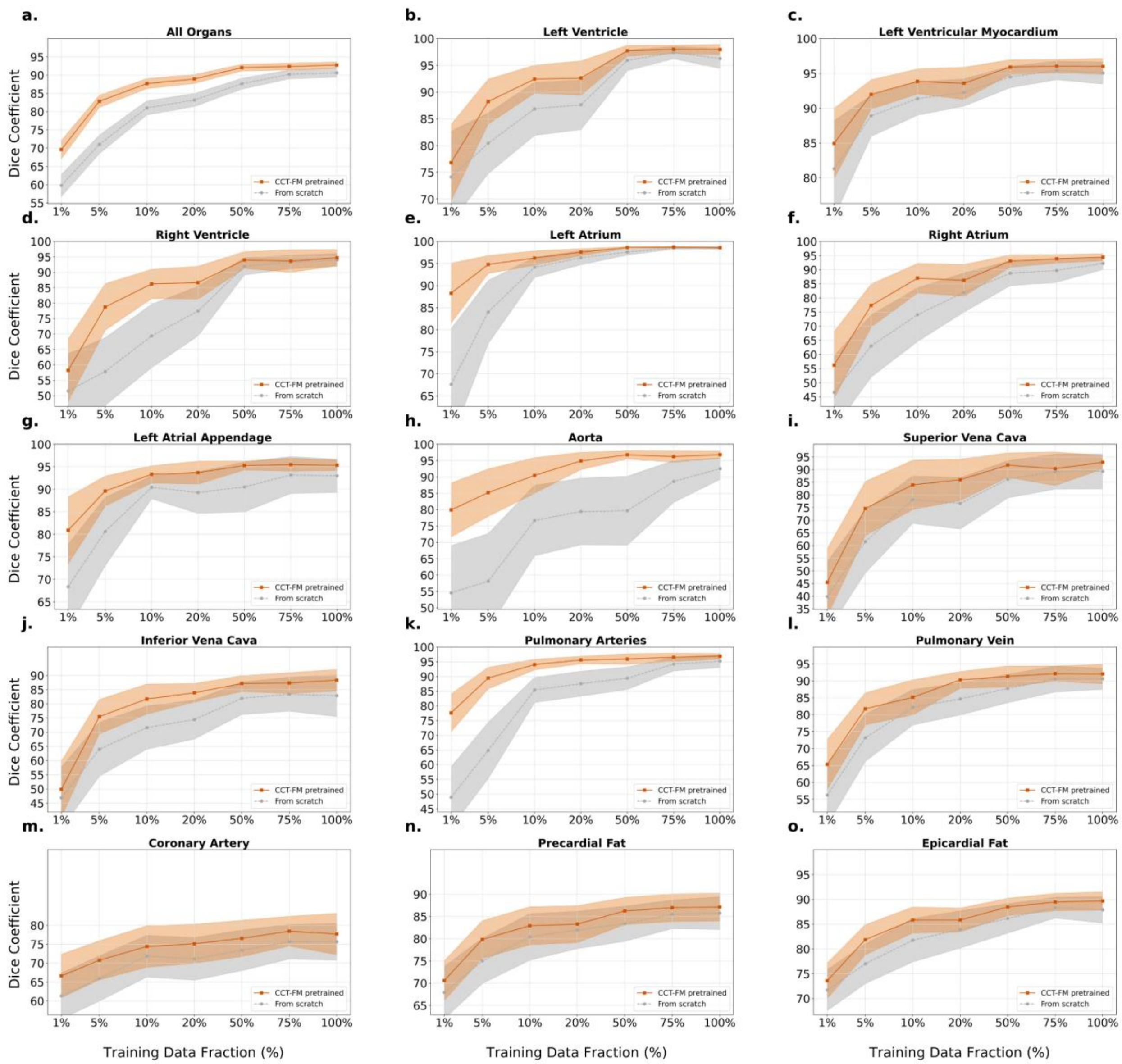

**Supplementary Figure S16. Per-structure data efficiency of self-supervised pre-training in female patients, across the ExtTest-4 dataset.** Mean Dice coefficient, reported as a micro-average across external datasets (N = 27) and as a function of labeled training data fraction (1–100%), for CCT-FM pre-trained (orange) versus from-scratch (gray) models. Results are shown for (a) all organs combined and (b–o) each of the 14 cardiac structures individually; shaded bands denote 95% confidence intervals. Two cases lack sex metadata and were excluded.

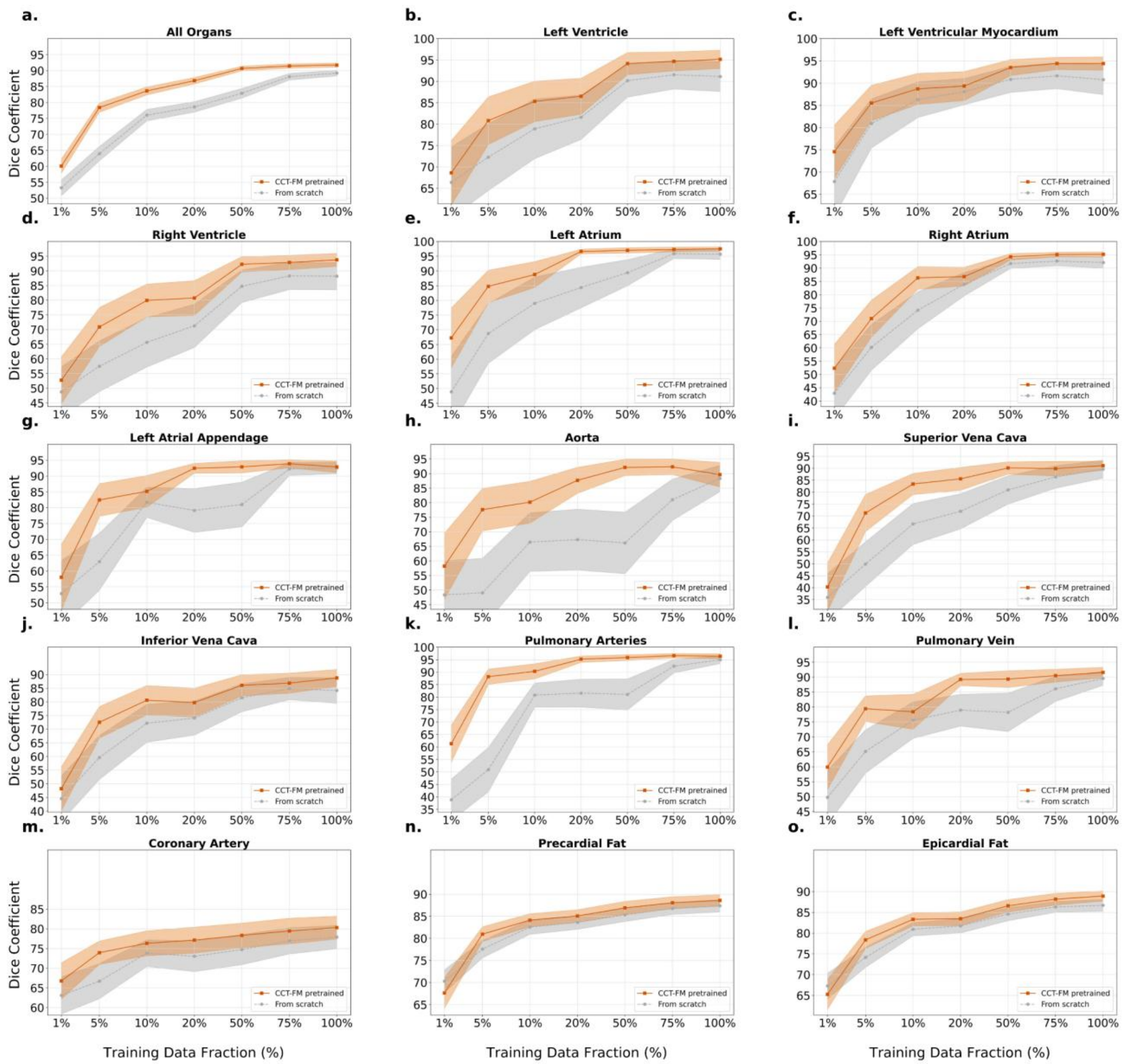

**Supplementary Figure S17. Per-structure data efficiency of self-supervised pre-training in male patients, across the ExtTest-4 dataset.** Mean Dice coefficient, reported as a micro-average across external datasets (N = 46) and as a function of labeled training data fraction (1–100%), for CCT-FM pre-trained (orange) versus from-scratch (gray) models. Results are shown for (a) all organs combined and (b–o) each of the 14 cardiac structures individually; shaded bands denote 95% confidence intervals. Two cases lack sex metadata and were excluded.

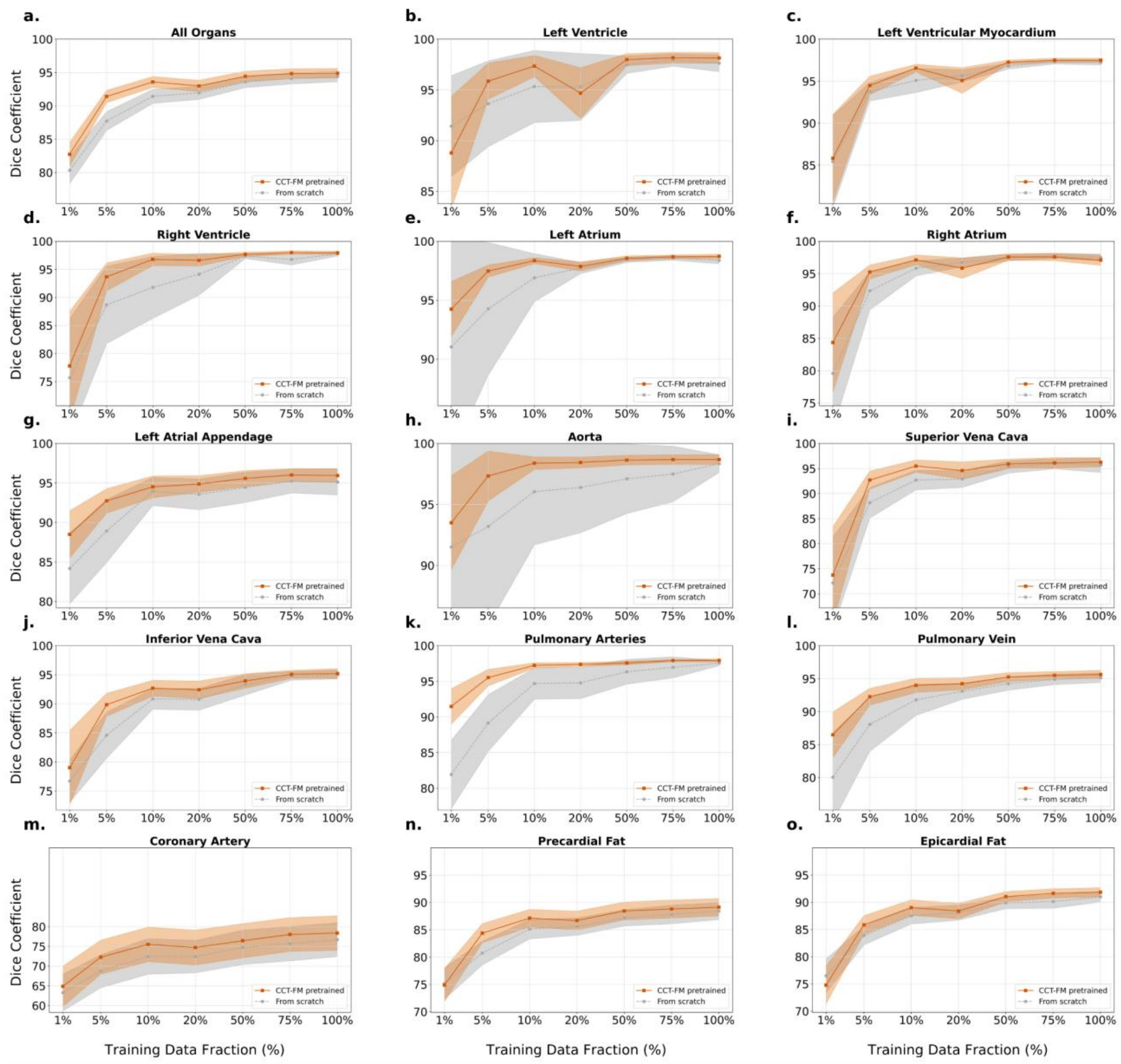

**Supplementary Figure S18. Per-structure data efficiency of self-supervised pre-training in all patients, across the ExtTest-5 dataset.** Mean Dice coefficient, reported as a micro-average across external datasets (N = 20) and as a function of labeled training data fraction (1–100%), for CCT-FM pre-trained (orange) versus from-scratch (gray) models. Results are shown for (a) all organs combined and (b–o) each of the 14 cardiac structures individually; shaded bands denote 95% confidence intervals.

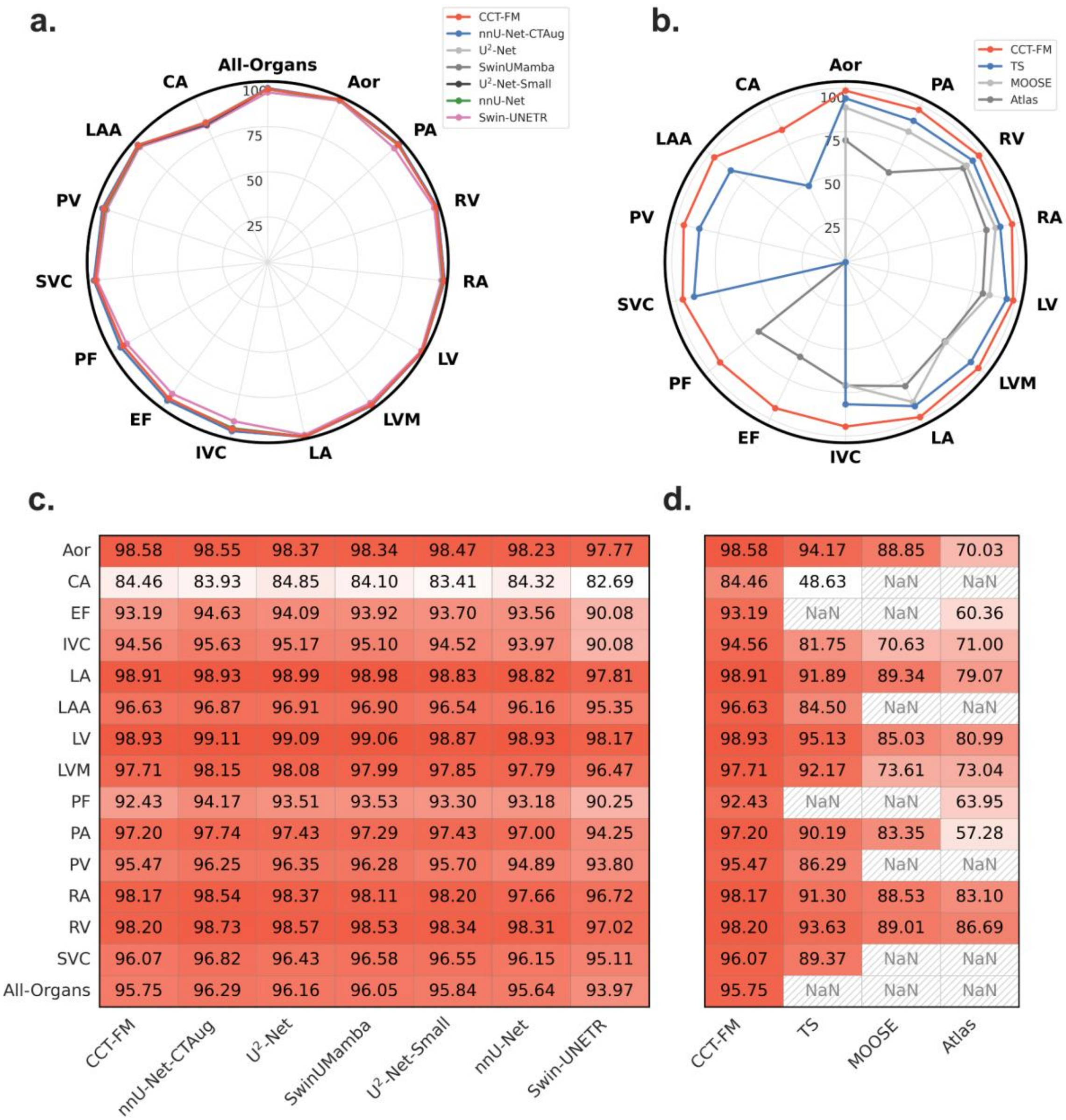

| | CCT-FM | nnU-Net-CTAug | $U^2$-Net | SwinUMamba | $U^2$-Net-Small | nnU-Net | Swin-UNETR |
|---|---|---|---|---|---|---|---|
| Aor | 98.58 | 98.55 | 98.37 | 98.34 | 98.47 | 98.23 | 97.77 |
| CA | 84.46 | 83.93 | 84.85 | 84.10 | 83.41 | 84.32 | 82.69 |
| EF | 93.19 | 94.63 | 94.09 | 93.92 | 93.70 | 93.56 | 90.08 |
| IVC | 94.56 | 95.63 | 95.17 | 95.10 | 94.52 | 93.97 | 90.08 |
| LA | 98.91 | 98.93 | 98.99 | 98.98 | 98.83 | 98.82 | 97.81 |
| LAA | 96.63 | 96.87 | 96.91 | 96.90 | 96.54 | 96.16 | 95.35 |
| LV | 98.93 | 99.11 | 99.09 | 99.06 | 98.87 | 98.93 | 98.17 |
| LVM | 97.71 | 98.15 | 98.08 | 97.99 | 97.85 | 97.79 | 96.47 |
| PF | 92.43 | 94.17 | 93.51 | 93.53 | 93.30 | 93.18 | 90.25 |
| PA | 97.20 | 97.74 | 97.43 | 97.29 | 97.43 | 97.00 | 94.25 |
| PV | 95.47 | 96.25 | 96.35 | 96.28 | 95.70 | 94.89 | 93.80 |
| RA | 98.17 | 98.54 | 98.37 | 98.11 | 98.20 | 97.66 | 96.72 |
| RV | 98.20 | 98.73 | 98.57 | 98.53 | 98.34 | 98.31 | 97.02 |
| SVC | 96.07 | 96.82 | 96.43 | 96.58 | 96.55 | 96.15 | 95.11 |
| All-Organs | 95.75 | 96.29 | 96.16 | 96.05 | 95.84 | 95.64 | 93.97 |

| | CCT-FM | TS | MOOSE | Atlas |
|---|---|---|---|---|
| Aor | 98.58 | 94.17 | 88.85 | 70.03 |
| CA | 84.46 | 48.63 | NaN | NaN |
| EF | 93.19 | NaN | NaN | 60.36 |
| IVC | 94.56 | 81.75 | 70.63 | 71.00 |
| LA | 98.91 | 91.89 | 89.34 | 79.07 |
| LAA | 96.63 | 84.50 | NaN | NaN |
| LV | 98.93 | 95.13 | 85.03 | 80.99 |
| LVM | 97.71 | 92.17 | 73.61 | 73.04 |
| PF | 92.43 | NaN | NaN | 63.95 |
| PA | 97.20 | 90.19 | 83.35 | 57.28 |
| PV | 95.47 | 86.29 | NaN | NaN |
| RA | 98.17 | 91.30 | 88.53 | 83.10 |
| RV | 98.20 | 93.63 | 89.01 | 86.69 |
| SVC | 96.07 | 89.37 | NaN | NaN |
| All-Organs | 95.75 | NaN | NaN | NaN |

**Supplementary Figure S19. Whole-heart segmentation benchmark across in-house models and open-source tools in female patients, across all datasets.** (a) Per-structure Dice coefficients for the seven in-house models (CCT-FM, nnU-Net, nnU-Net-CTAug, SwinUMamba, Swin-UNETR, $U^2$-Net, and $U^2$-Net-Small) across all fourteen cardiac structures, where All-Organs denotes the mean across structures. (b) Per-structure Dice coefficients comparing CCT-FM with the three open-source tools, restricted to the structures each tool produces. (c) Heatmap of mean Dice for every in-house model and structure, with the All-Organs row giving the across-structure mean. (d) Heatmap of mean Dice for CCT-FM against the three open-source tools, where hatched cells mark structures a tool does not produce and that are therefore excluded from its aggregate. All values are mean Dice computed across the external test datasets at the full labeled training set. Structure abbreviations: Aor, aorta; CA, coronary arteries; EF, epicardial fat; IVC, inferior vena cava; LA, left atrium; LAA, left atrial appendage; LV, left ventricle; LVM, left ventricular myocardium; PA, pulmonary arteries; PF, pericardial fat; PV, pulmonary vein; RA, right atrium; RV, right ventricle; SVC, superior vena cava.

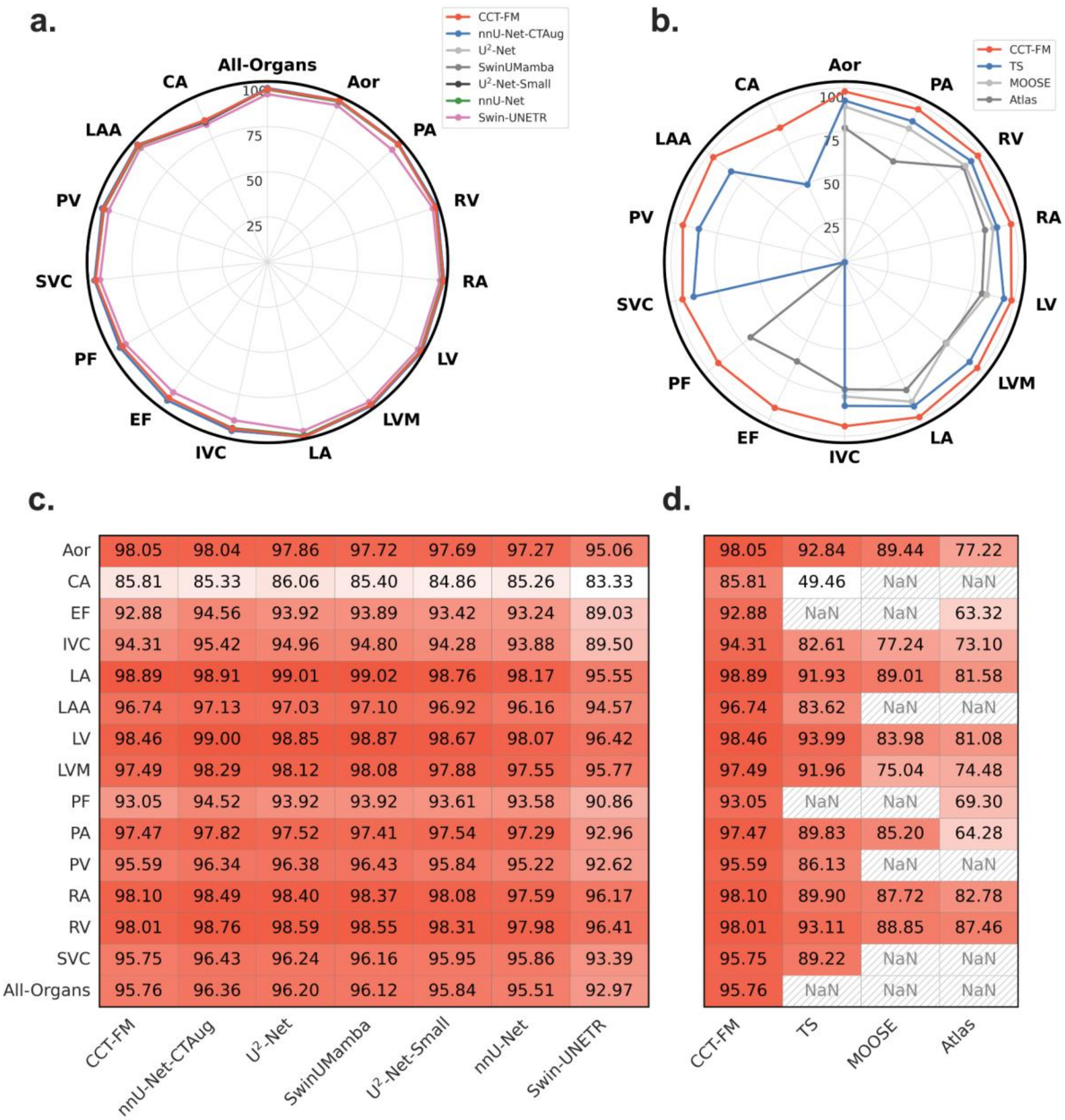

**Supplementary Figure S20. Whole-heart segmentation benchmark across in-house models and open-source tools in male patients, across all datasets.** (a) Per-structure Dice coefficients for the seven in-house models (CCT-FM, nnU-Net, nnU-Net-CTAug, SwinUMamba, Swin-UNETR, U²-Net, and U²-Net-Small) across all fourteen cardiac structures, where All-Organs denotes the mean across structures. (b) Per-structure Dice coefficients comparing CCT-FM with the three open-source tools, restricted to the structures each tool produces. (c) Heatmap of mean Dice for every in-house model and structure, with the All-Organs row giving the across-structure mean. (d) Heatmap of mean Dice for CCT-FM against the three open-source tools, where hatched cells mark structures a tool does not produce and that are therefore excluded from its aggregate. All values are mean Dice computed across the external test datasets at the full labeled training set. Structure abbreviations: Aor, aorta; CA, coronary arteries; EF, epicardial fat; IVC, inferior vena cava; LA, left atrium; LAA, left atrial appendage; LV, left ventricle; LVM, left ventricular myocardium; PA, pulmonary arteries; PF, pericardial fat; PV, pulmonary vein; RA, right atrium; RV, right ventricle; SVC, superior vena cava.

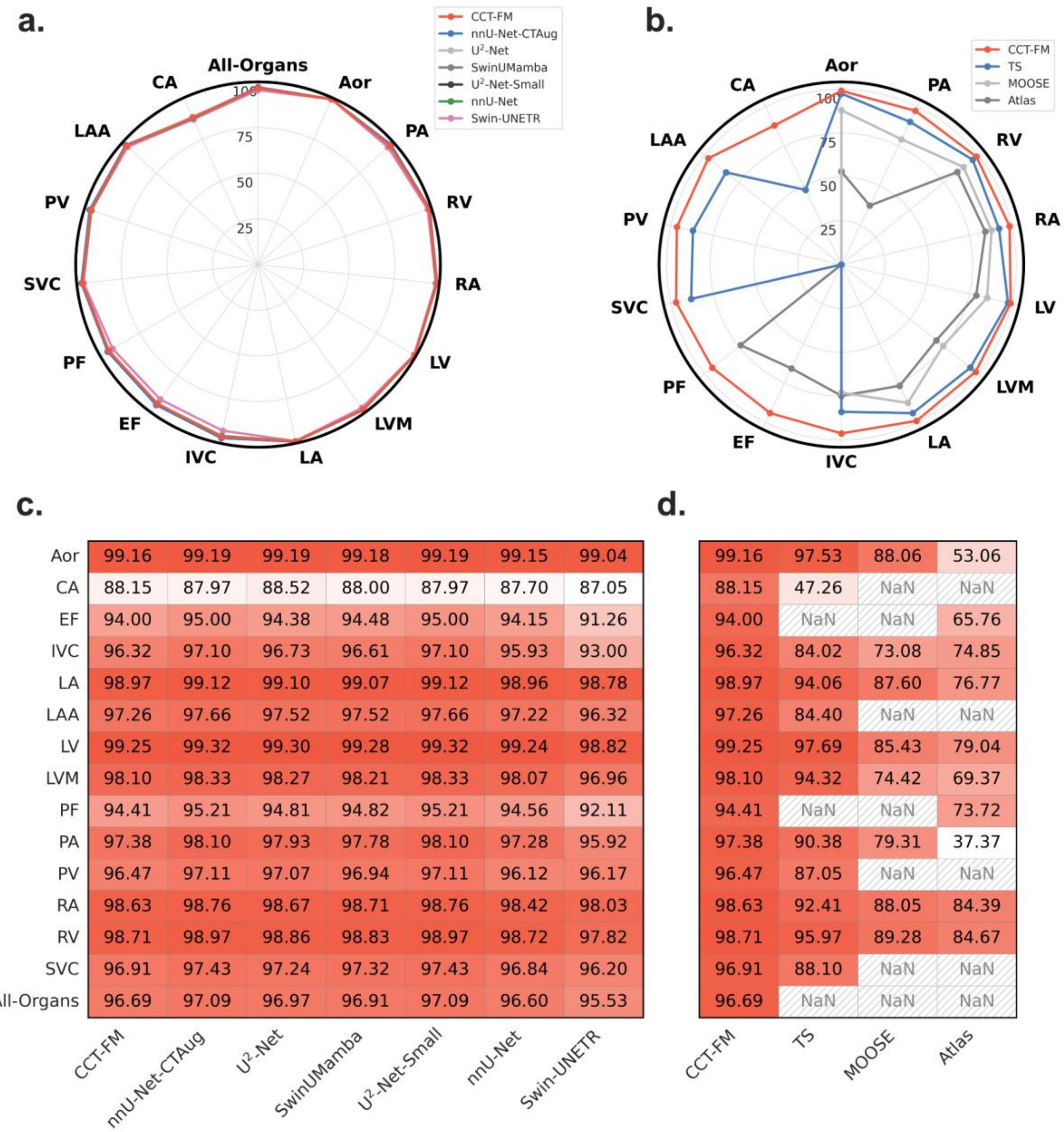

**Supplementary Figure S21. Whole-heart segmentation benchmark across in-house models and open-source tools across all patients in the ExtTest-1 dataset.** (a) Per-structure Dice coefficients for the seven in-house models (CCT-FM, nnU-Net, nnU-Net-CTAug, SwinUMamba, Swin-UNETR, U²-Net, and U²-Net-Small) across all fourteen cardiac structures, where All-Organs denotes the mean across structures. (b) Per-structure Dice coefficients comparing CCT-FM with the three open-source tools, restricted to the structures each tool produces. (c) Heatmap of mean Dice for every in-house model and structure, with the All-Organs row giving the across-structure mean. (d) Heatmap of mean Dice for CCT-FM against the three open-source tools, where hatched cells mark structures a tool does not produce and that are therefore excluded from its aggregate. All values are mean Dice computed across the external test datasets at the full labeled training set. Structure abbreviations: Aor, aorta; CA, coronary arteries; EF, epicardial fat; IVC, inferior vena cava; LA, left atrium; LAA, left atrial appendage; LV, left ventricle; LVM, left ventricular myocardium; PA, pulmonary arteries; PF, pericardial fat; PV, pulmonary vein; RA, right atrium; RV, right ventricle; SVC, superior vena cava.

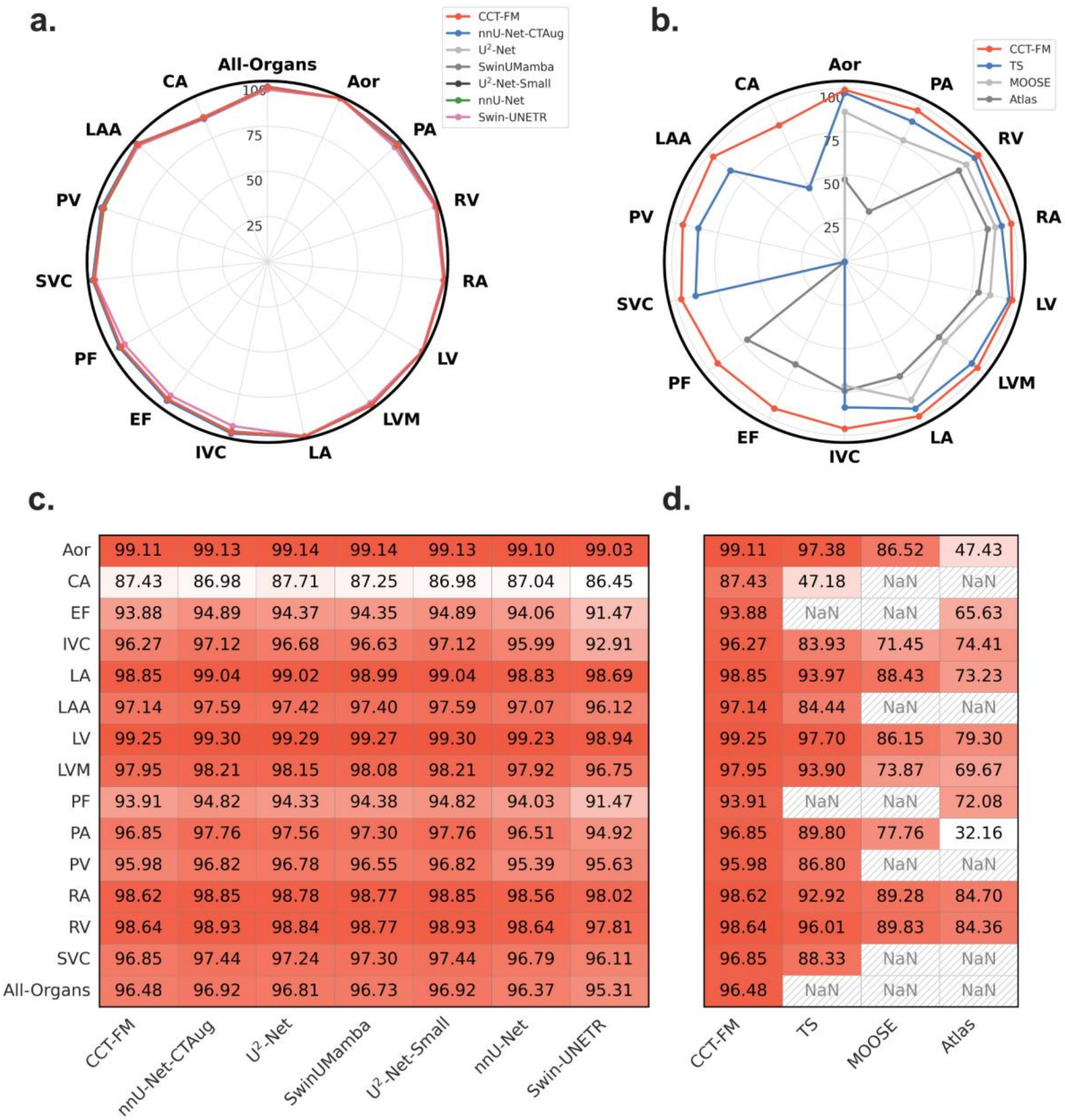

**Supplementary Figure S22. Whole-heart segmentation benchmark across in-house models and open-source tools across female patients in the ExtTest-1 dataset.** (a) Per-structure Dice coefficients for the seven in-house models (CCT-FM, nnU-Net, nnU-Net-CTAug, SwinUMamba, Swin-UNETR, U²-Net, and U²-Net-Small) across all fourteen cardiac structures, where All-Organs denotes the mean across structures. (b) Per-structure Dice coefficients comparing CCT-FM with the three open-source tools, restricted to the structures each tool produces. (c) Heatmap of mean Dice for every in-house model and structure, with the All-Organs row giving the across-structure mean. (d) Heatmap of mean Dice for CCT-FM against the three open-source tools, where hatched cells mark structures a tool does not produce and that are therefore excluded from its aggregate. All values are mean Dice computed across the external test datasets at the full labeled training set. Structure abbreviations: Aor, aorta; CA, coronary arteries; EF, epicardial fat; IVC, inferior vena cava; LA, left atrium; LAA, left atrial appendage; LV, left ventricle; LVM, left ventricular myocardium; PA, pulmonary arteries; PF, pericardial fat; PV, pulmonary vein; RA, right atrium; RV, right ventricle; SVC, superior vena cava.

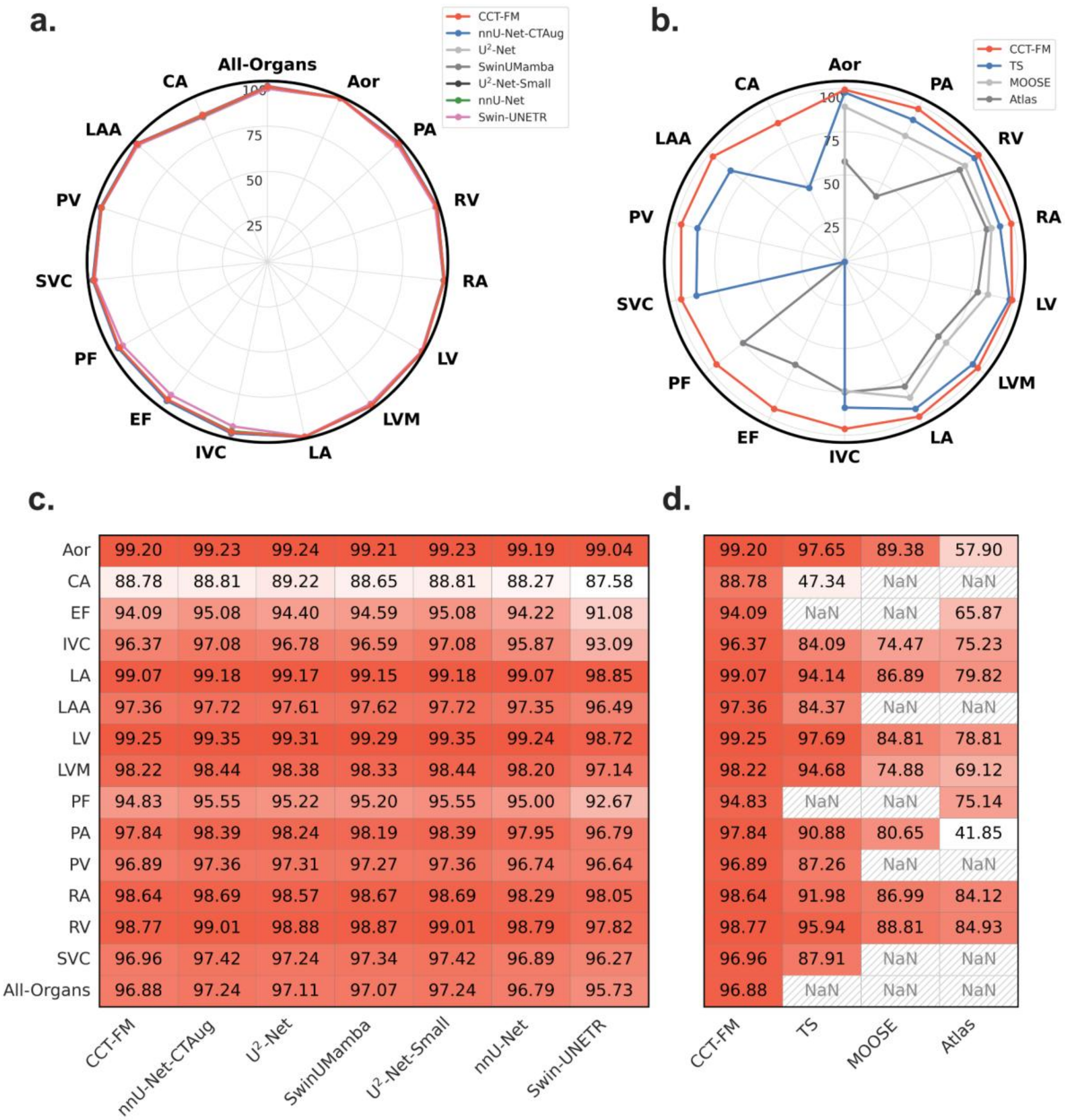

**Supplementary Figure S23. Whole-heart segmentation benchmark across in-house models and open-source tools across male patients in the ExtTest-1 dataset.** (a) Per-structure Dice coefficients for the seven in-house models (CCT-FM, nnU-Net, nnU-Net-CTAug, SwinUMamba, Swin-UNETR, U²-Net, and U²-Net-Small) across all fourteen cardiac structures, where All-Organs denotes the mean across structures. (b) Per-structure Dice coefficients comparing CCT-FM with the three open-source tools, restricted to the structures each tool produces. (c) Heatmap of mean Dice for every in-house model and structure, with the All-Organs row giving the across-structure mean. (d) Heatmap of mean Dice for CCT-FM against the three open-source tools, where hatched cells mark structures a tool does not produce and that are therefore excluded from its aggregate. All values are mean Dice computed across the external test datasets at the full labeled training set. Structure abbreviations: Aor, aorta; CA, coronary arteries; EF, epicardial fat; IVC, inferior vena cava; LA, left atrium; LAA, left atrial appendage; LV, left ventricle; LVM, left ventricular myocardium; PA, pulmonary arteries; PF, pericardial fat; PV, pulmonary vein; RA, right atrium; RV, right ventricle; SVC, superior vena cava.

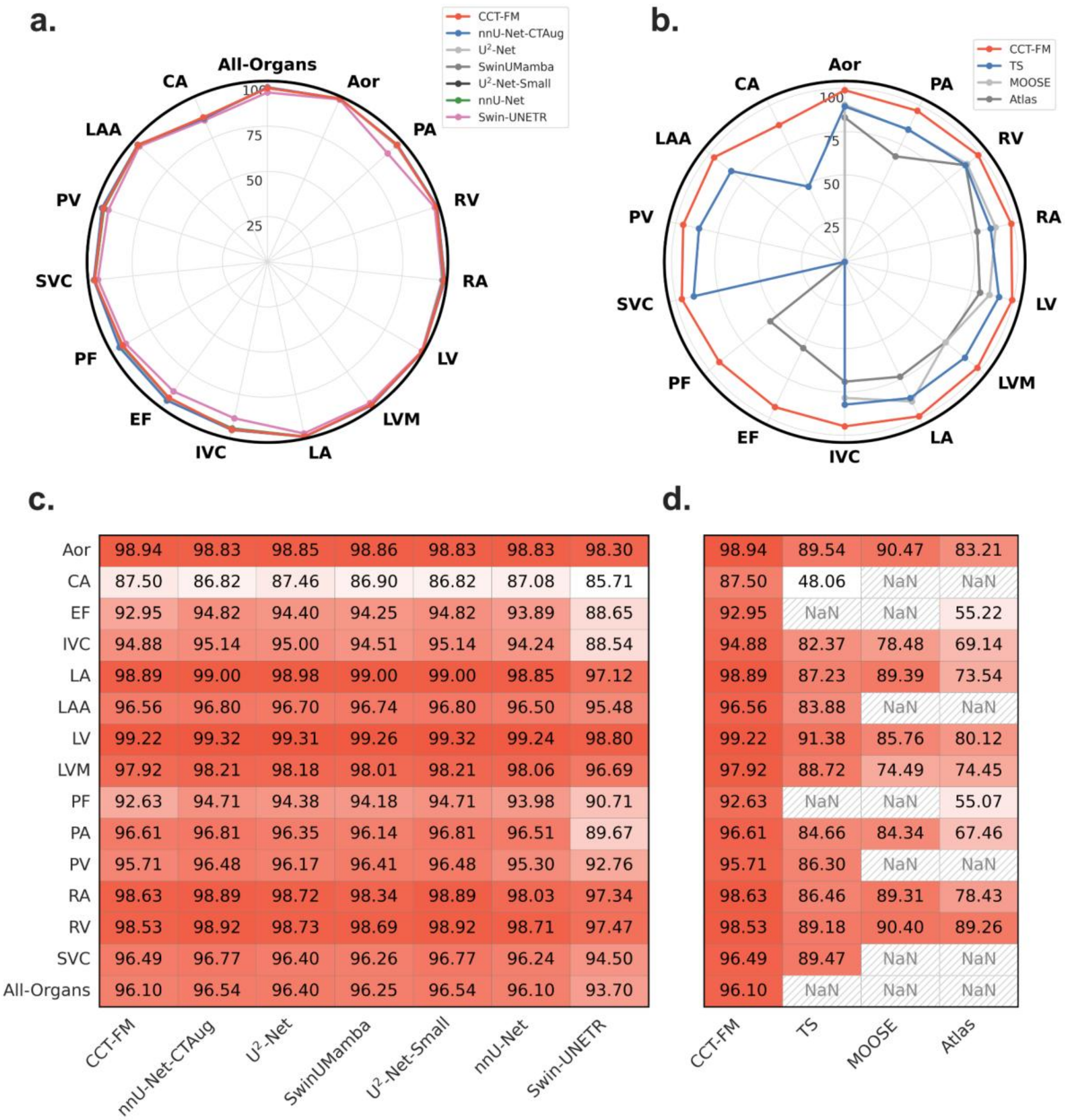

**Supplementary Figure S24. Whole-heart segmentation benchmark across in-house models and open-source tools across all patients in the ExtTest-2 dataset.** (a) Per-structure Dice coefficients for the seven in-house models (CCT-FM, nnU-Net, nnU-Net-CTAug, SwinUMamba, Swin-UNETR, U²-Net, and U²-Net-Small) across all fourteen cardiac structures, where All-Organs denotes the mean across structures. (b) Per-structure Dice coefficients comparing CCT-FM with the three open-source tools, restricted to the structures each tool produces. (c) Heatmap of mean Dice for every in-house model and structure, with the All-Organs row giving the across-structure mean. (d) Heatmap of mean Dice for CCT-FM against the three open-source tools, where hatched cells mark structures a tool does not produce and that are therefore excluded from its aggregate. All values are mean Dice computed across the external test datasets at the full labeled training set. Structure abbreviations: Aor, aorta; CA, coronary arteries; EF, epicardial fat; IVC, inferior vena cava; LA, left atrium; LAA, left atrial appendage; LV, left ventricle; LVM, left ventricular myocardium; PA, pulmonary arteries; PF, pericardial fat; PV, pulmonary vein; RA, right atrium; RV, right ventricle; SVC, superior vena cava.

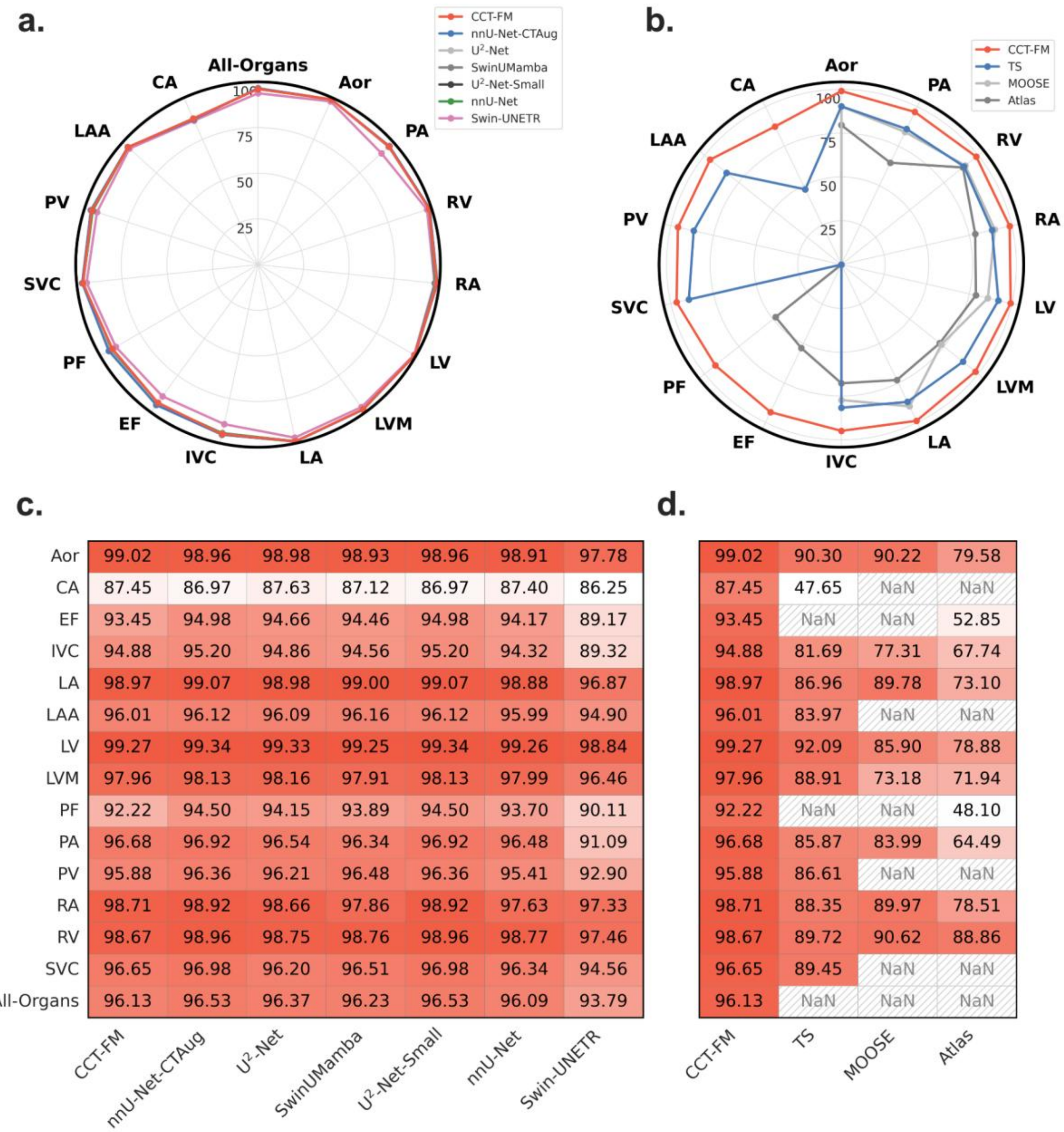

| | CCT-FM | nnU-Net-CTAug | U²-Net | SwinUMamba | U²-Net-Small | nnU-Net | Swin-UNETR |
|---|---|---|---|---|---|---|---|
| Aor | 99.02 | 98.96 | 98.98 | 98.93 | 98.96 | 98.91 | 97.78 |
| CA | 87.45 | 86.97 | 87.63 | 87.12 | 86.97 | 87.40 | 86.25 |
| EF | 93.45 | 94.98 | 94.66 | 94.46 | 94.98 | 94.17 | 89.17 |
| IVC | 94.88 | 95.20 | 94.86 | 94.56 | 95.20 | 94.32 | 89.32 |
| LA | 98.97 | 99.07 | 98.98 | 99.00 | 99.07 | 98.88 | 96.87 |
| LAA | 96.01 | 96.12 | 96.09 | 96.16 | 96.12 | 95.99 | 94.90 |
| LV | 99.27 | 99.34 | 99.33 | 99.25 | 99.34 | 99.26 | 98.84 |
| LVM | 97.96 | 98.13 | 98.16 | 97.91 | 98.13 | 97.99 | 96.46 |
| PF | 92.22 | 94.50 | 94.15 | 93.89 | 94.50 | 93.70 | 90.11 |
| PA | 96.68 | 96.92 | 96.54 | 96.34 | 96.92 | 96.48 | 91.09 |
| PV | 95.88 | 96.36 | 96.21 | 96.48 | 96.36 | 95.41 | 92.90 |
| RA | 98.71 | 98.92 | 98.66 | 97.86 | 98.92 | 97.63 | 97.33 |
| RV | 98.67 | 98.96 | 98.75 | 98.76 | 98.96 | 98.77 | 97.46 |
| SVC | 96.65 | 96.98 | 96.20 | 96.51 | 96.98 | 96.34 | 94.56 |
| All-Organs | 96.13 | 96.53 | 96.37 | 96.23 | 96.53 | 96.09 | 93.79 |

| | CCT-FM | TS | MOOSE | Atlas |
|---|---|---|---|---|
| Aor | 99.02 | 90.30 | 90.22 | 79.58 |
| CA | 87.45 | 47.65 | NaN | NaN |
| EF | 93.45 | NaN | NaN | 52.85 |
| IVC | 94.88 | 81.69 | 77.31 | 67.74 |
| LA | 98.97 | 86.96 | 89.78 | 73.10 |
| LAA | 96.01 | 83.97 | NaN | NaN |
| LV | 99.27 | 92.09 | 85.90 | 78.88 |
| LVM | 97.96 | 88.91 | 73.18 | 71.94 |
| PF | 92.22 | NaN | NaN | 48.10 |
| PA | 96.68 | 85.87 | 83.99 | 64.49 |
| PV | 95.88 | 86.61 | NaN | NaN |
| RA | 98.71 | 88.35 | 89.97 | 78.51 |
| RV | 98.67 | 89.72 | 90.62 | 88.86 |
| SVC | 96.65 | 89.45 | NaN | NaN |
| All-Organs | 96.13 | NaN | NaN | NaN |

**Supplementary Figure S25. Whole-heart segmentation benchmark across in-house models and open-source tools across female patients in the ExtTest-2 dataset.** (a) Per-structure Dice coefficients for the seven in-house models (CCT-FM, nnU-Net, nnU-Net-CTAug, SwinUMamba, Swin-UNETR, U²-Net, and U²-Net-Small) across all fourteen cardiac structures, where All-Organs denotes the mean across structures. (b) Per-structure Dice coefficients comparing CCT-FM with the three open-source tools, restricted to the structures each tool produces. (c) Heatmap of mean Dice for every in-house model and structure, with the All-Organs row giving the across-structure mean. (d) Heatmap of mean Dice for CCT-FM against the three open-source tools, where hatched cells mark structures a tool does not produce and that are therefore excluded from its aggregate. All values are mean Dice computed across the external test datasets at the full labeled training set. Structure abbreviations: Aor, aorta; CA, coronary arteries; EF, epicardial fat; IVC, inferior vena cava; LA, left atrium; LAA, left atrial appendage; LV, left ventricle; LVM, left ventricular myocardium; PA, pulmonary arteries; PF, pericardial fat; PV, pulmonary vein; RA, right atrium; RV, right ventricle; SVC, superior vena cava.

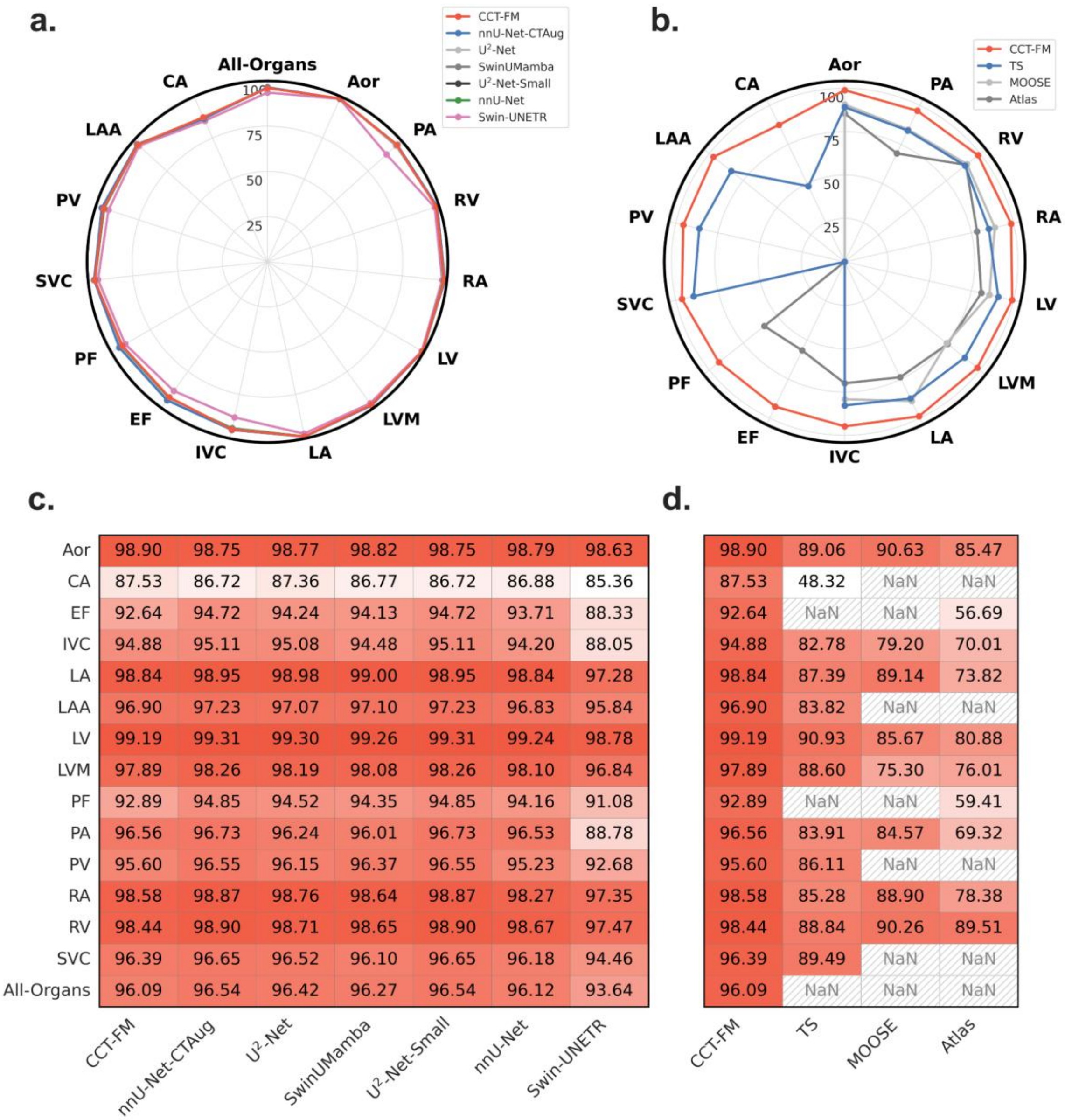

**Supplementary Figure S26. Whole-heart segmentation benchmark across in-house models and open-source tools across male patients in the ExtTest-2 dataset.** (a) Per-structure Dice coefficients for the seven in-house models (CCT-FM, nnU-Net, nnU-Net-CTAug, SwinUMamba, Swin-UNETR, U²-Net, and U²-Net-Small) across all fourteen cardiac structures, where All-Organs denotes the mean across structures. (b) Per-structure Dice coefficients comparing CCT-FM with the three open-source tools, restricted to the structures each tool produces. (c) Heatmap of mean Dice for every in-house model and structure, with the All-Organs row giving the across-structure mean. (d) Heatmap of mean Dice for CCT-FM against the three open-source tools, where hatched cells mark structures a tool does not produce and that are therefore excluded from its aggregate. All values are mean Dice computed across the external test datasets at the full labeled training set. Structure abbreviations: Aor, aorta; CA, coronary arteries; EF, epicardial fat; IVC, inferior vena cava; LA, left atrium; LAA, left atrial appendage; LV, left ventricle; LVM, left ventricular myocardium; PA, pulmonary arteries; PF, pericardial fat; PV, pulmonary vein; RA, right atrium; RV, right ventricle; SVC, superior vena cava.

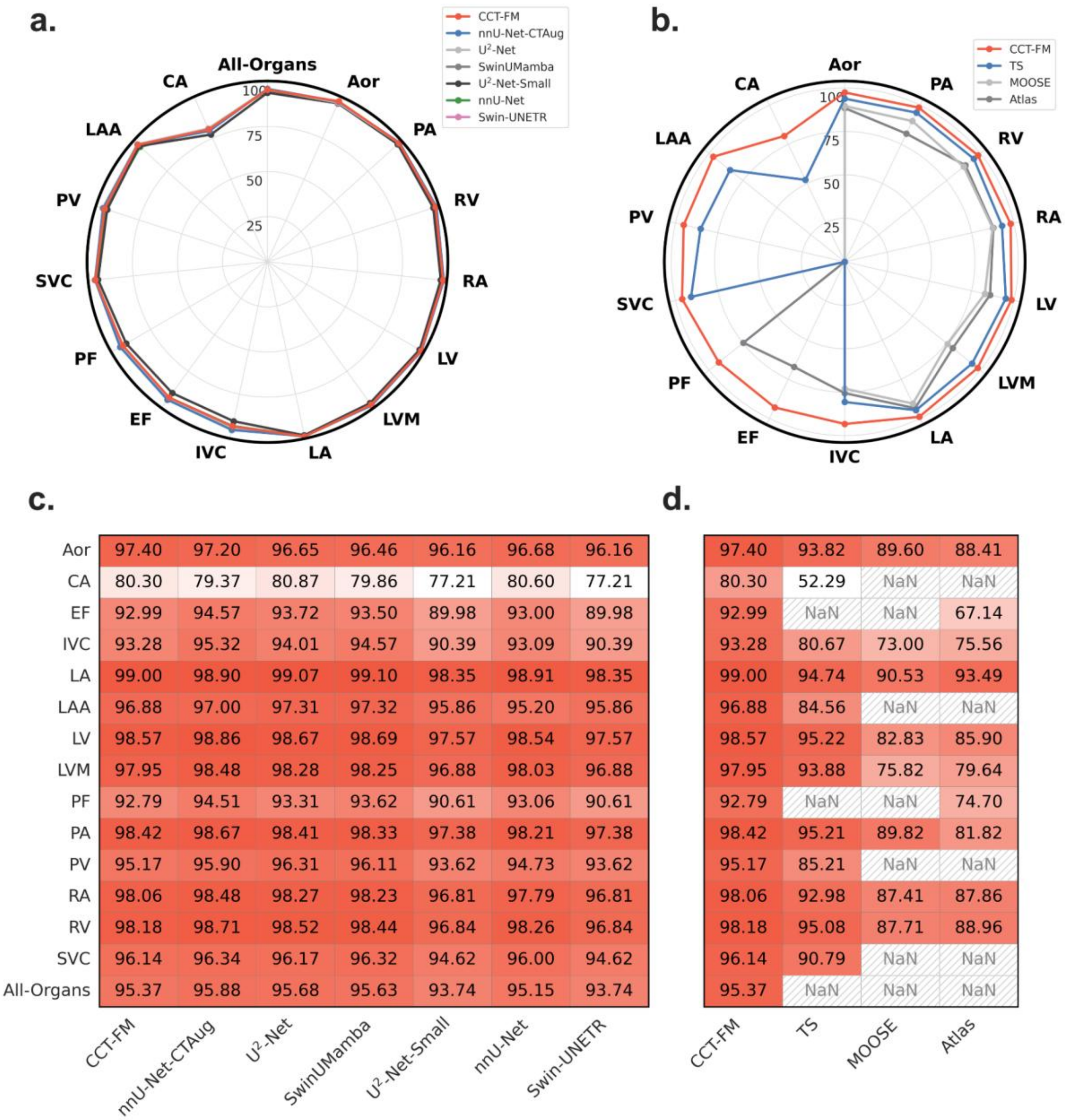

**Supplementary Figure S27. Whole-heart segmentation benchmark across in-house models and open-source tools across all patients in the ExtTest-3 dataset.** (a) Per-structure Dice coefficients for the seven in-house models (CCT-FM, nnU-Net, nnU-Net-CTAug, SwinUMamba, Swin-UNETR, U²-Net, and U²-Net-Small) across all fourteen cardiac structures, where All-Organs denotes the mean across structures. (b) Per-structure Dice coefficients comparing CCT-FM with the three open-source tools, restricted to the structures each tool produces. (c) Heatmap of mean Dice for every in-house model and structure, with the All-Organs row giving the across-structure mean. (d) Heatmap of mean Dice for CCT-FM against the three open-source tools, where hatched cells mark structures a tool does not produce and that are therefore excluded from its aggregate. All values are mean Dice computed across the external test datasets at the full labeled training set. Structure abbreviations: Aor, aorta; CA, coronary arteries; EF, epicardial fat; IVC, inferior vena cava; LA, left atrium; LAA, left atrial appendage; LV, left ventricle; LVM, left ventricular myocardium; PA, pulmonary arteries; PF, pericardial fat; PV, pulmonary vein; RA, right atrium; RV, right ventricle; SVC, superior vena cava.

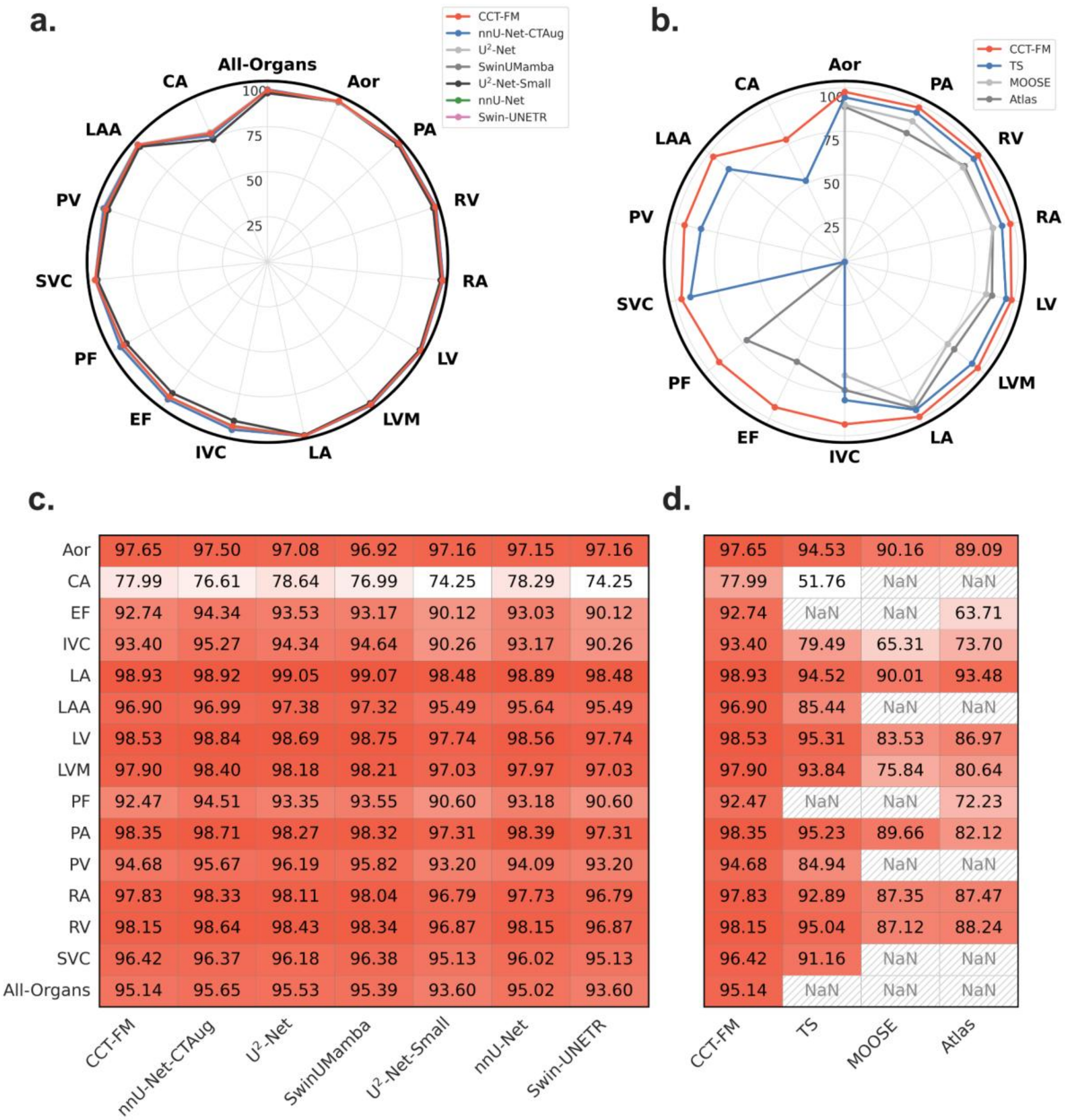

**Supplementary Figure S28. Whole-heart segmentation benchmark across in-house models and open-source tools across female patients in the ExtTest-3 dataset.** (a) Per-structure Dice coefficients for the seven in-house models (CCT-FM, nnU-Net, nnU-Net-CTAug, SwinUMamba, Swin-UNETR, U²-Net, and U²-Net-Small) across all fourteen cardiac structures, where All-Organs denotes the mean across structures. (b) Per-structure Dice coefficients comparing CCT-FM with the three open-source tools, restricted to the structures each tool produces. (c) Heatmap of mean Dice for every in-house model and structure, with the All-Organs row giving the across-structure mean. (d) Heatmap of mean Dice for CCT-FM against the three open-source tools, where hatched cells mark structures a tool does not produce and that are therefore excluded from its aggregate. All values are mean Dice computed across the external test datasets at the full labeled training set. Structure abbreviations: Aor, aorta; CA, coronary arteries; EF, epicardial fat; IVC, inferior vena cava; LA, left atrium; LAA, left atrial appendage; LV, left ventricle; LVM, left ventricular myocardium; PA, pulmonary arteries; PF, pericardial fat; PV, pulmonary vein; RA, right atrium; RV, right ventricle; SVC, superior vena cava.

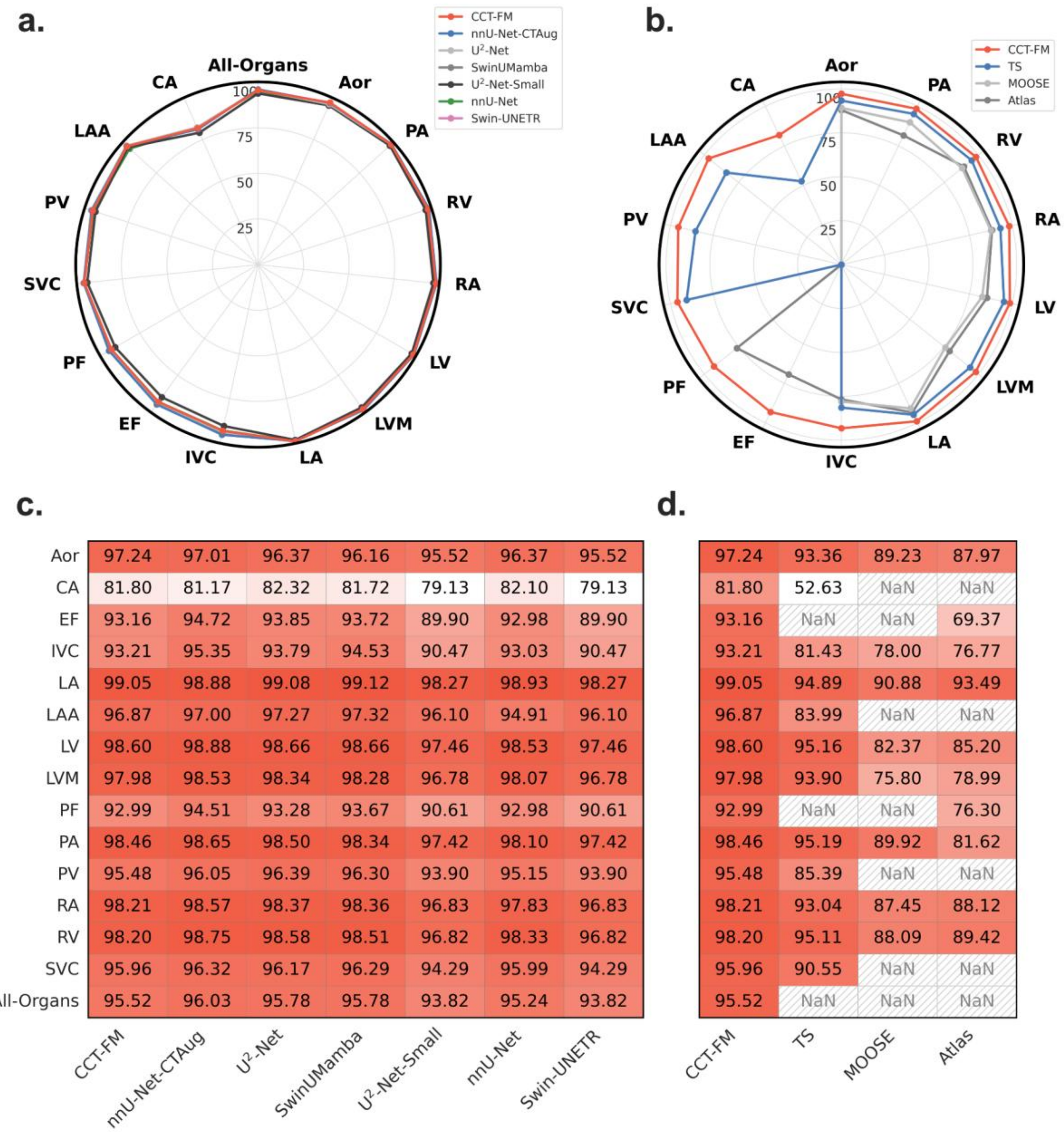

**Supplementary Figure S29. Whole-heart segmentation benchmark across in-house models and open-source tools across male patients in the ExtTest-3 dataset.** (a) Per-structure Dice coefficients for the seven in-house models (CCT-FM, nnU-Net, nnU-Net-CTAug, SwinUMamba, Swin-UNETR, U²-Net, and U²-Net-Small) across all fourteen cardiac structures, where All-Organs denotes the mean across structures. (b) Per-structure Dice coefficients comparing CCT-FM with the three open-source tools, restricted to the structures each tool produces. (c) Heatmap of mean Dice for every in-house model and structure, with the All-Organs row giving the across-structure mean. (d) Heatmap of mean Dice for CCT-FM against the three open-source tools, where hatched cells mark structures a tool does not produce and that are therefore excluded from its aggregate. All values are mean Dice computed across the external test datasets at the full labeled training set. Structure abbreviations: Aor, aorta; CA, coronary arteries; EF, epicardial fat; IVC, inferior vena cava; LA, left atrium; LAA, left atrial appendage; LV, left ventricle; LVM, left ventricular myocardium; PA, pulmonary arteries; PF, pericardial fat; PV, pulmonary vein; RA, right atrium; RV, right ventricle; SVC, superior vena cava.

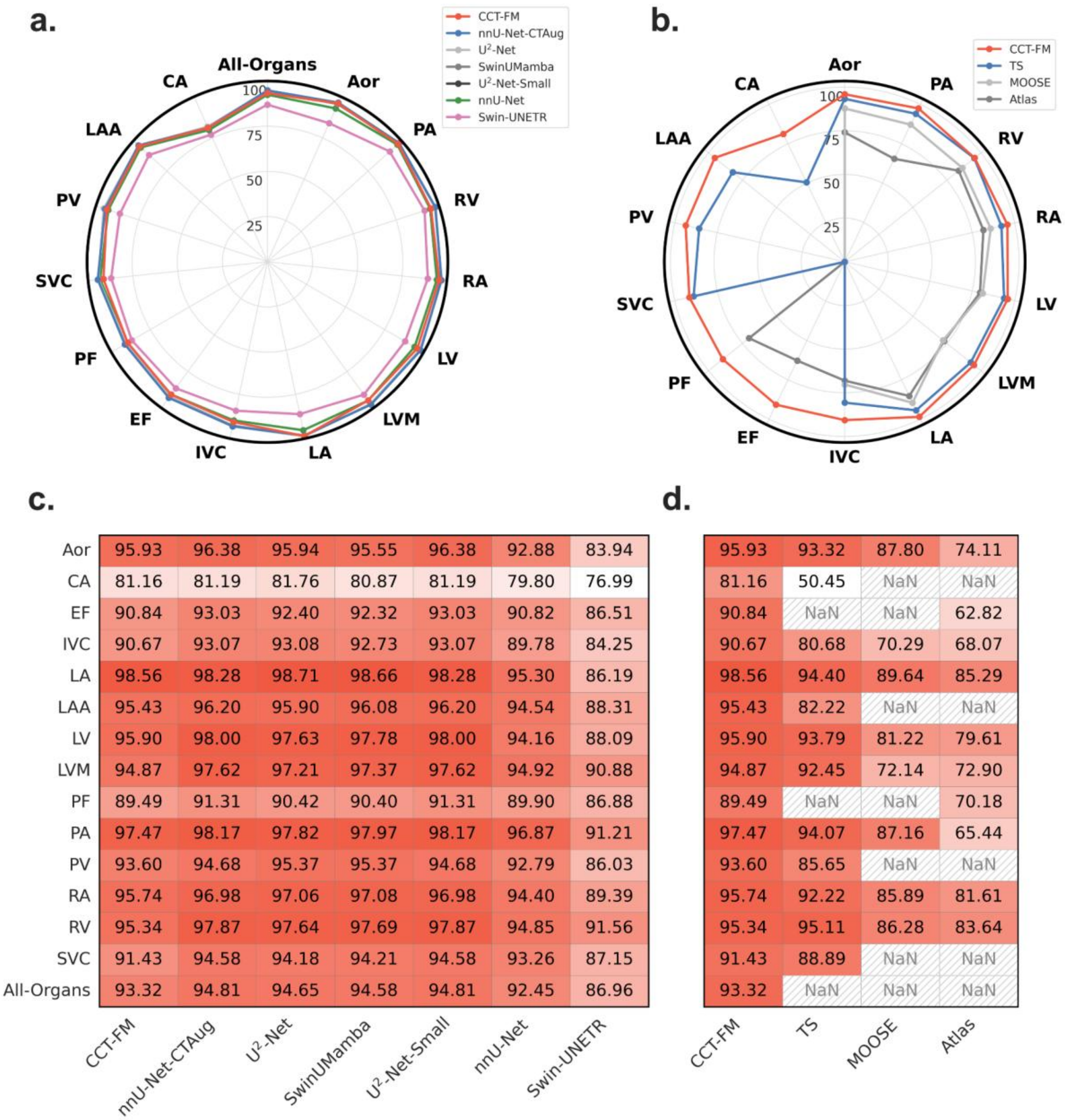

**Supplementary Figure S30. Whole-heart segmentation benchmark across in-house models and open-source tools across all patients in the ExtTest-4 dataset.** (a) Per-structure Dice coefficients for the seven in-house models (CCT-FM, nnU-Net, nnU-Net-CTAug, SwinUMamba, Swin-UNETR, U²-Net, and U²-Net-Small) across all fourteen cardiac structures, where All-Organs denotes the mean across structures. (b) Per-structure Dice coefficients comparing CCT-FM with the three open-source tools, restricted to the structures each tool produces. (c) Heatmap of mean Dice for every in-house model and structure, with the All-Organs row giving the across-structure mean. (d) Heatmap of mean Dice for CCT-FM against the three open-source tools, where hatched cells mark structures a tool does not produce and that are therefore excluded from its aggregate. All values are mean Dice computed across the external test datasets at the full labeled training set. Structure abbreviations: Aor, aorta; CA, coronary arteries; EF, epicardial fat; IVC, inferior vena cava; LA, left atrium; LAA, left atrial appendage; LV, left ventricle; LVM, left ventricular myocardium; PA, pulmonary arteries; PF, pericardial fat; PV, pulmonary vein; RA, right atrium; RV, right ventricle; SVC, superior vena cava.

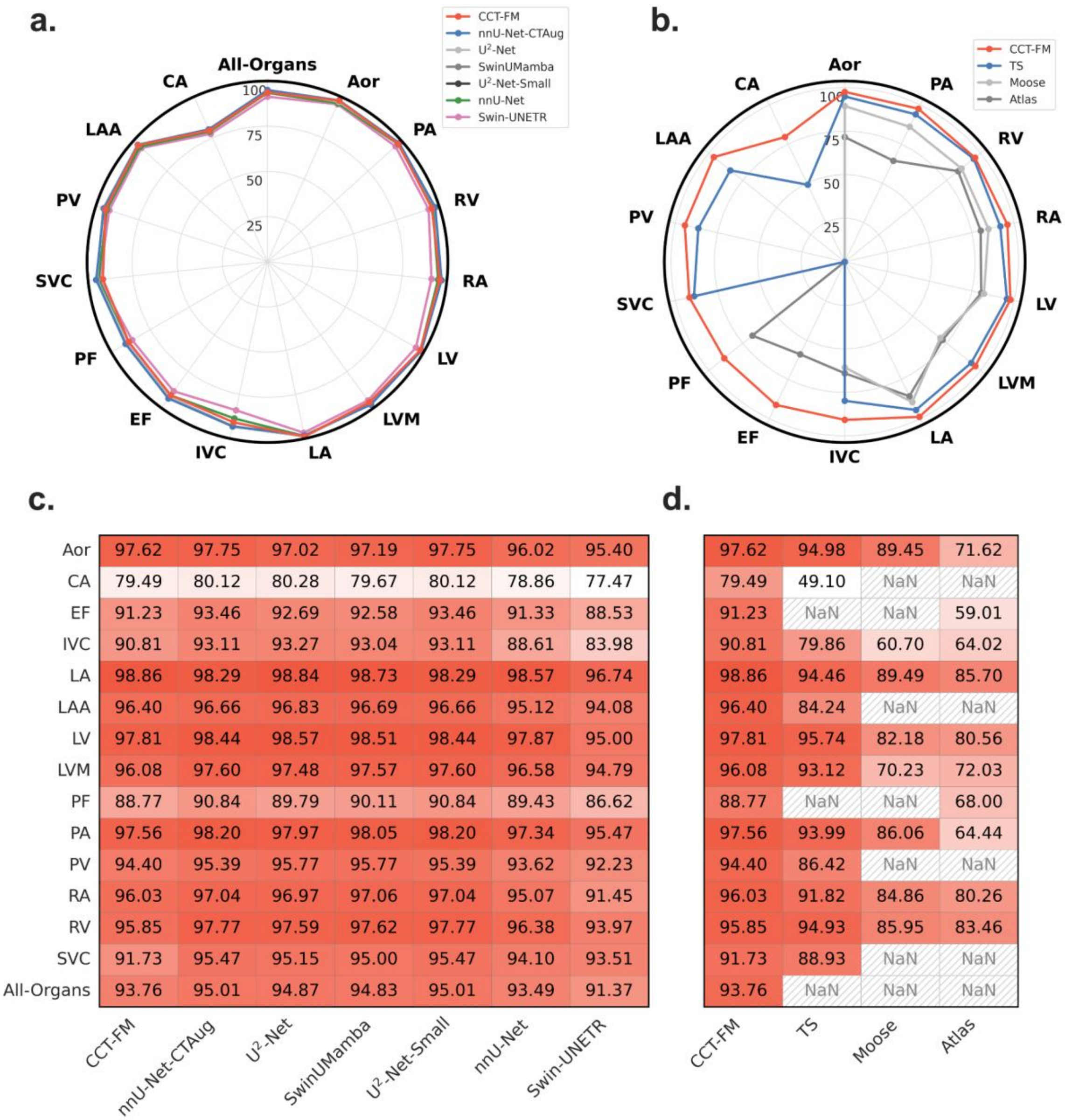

**Supplementary Figure S31. Whole-heart segmentation benchmark across in-house models and open-source tools across female patients in the ExtTest-4 dataset.** (a) Per-structure Dice coefficients for the seven in-house models (CCT-FM, nnU-Net, nnU-Net-CTAug, SwinUMamba, Swin-UNETR, U²-Net, and U²-Net-Small) across all fourteen cardiac structures, where All-Organs denotes the mean across structures. (b) Per-structure Dice coefficients comparing CCT-FM with the three open-source tools, restricted to the structures each tool produces. (c) Heatmap of mean Dice for every in-house model and structure, with the All-Organs row giving the across-structure mean. (d) Heatmap of mean Dice for CCT-FM against the three open-source tools, where hatched cells mark structures a tool does not produce and that are therefore excluded from its aggregate. All values are mean Dice computed across the external test datasets at the full labeled training set. Structure abbreviations: Aor, aorta; CA, coronary arteries; EF, epicardial fat; IVC, inferior vena cava; LA, left atrium; LAA, left atrial appendage; LV, left ventricle; LVM, left ventricular myocardium; PA, pulmonary arteries; PF, pericardial fat; PV, pulmonary vein; RA, right atrium; RV, right ventricle; SVC, superior vena cava.

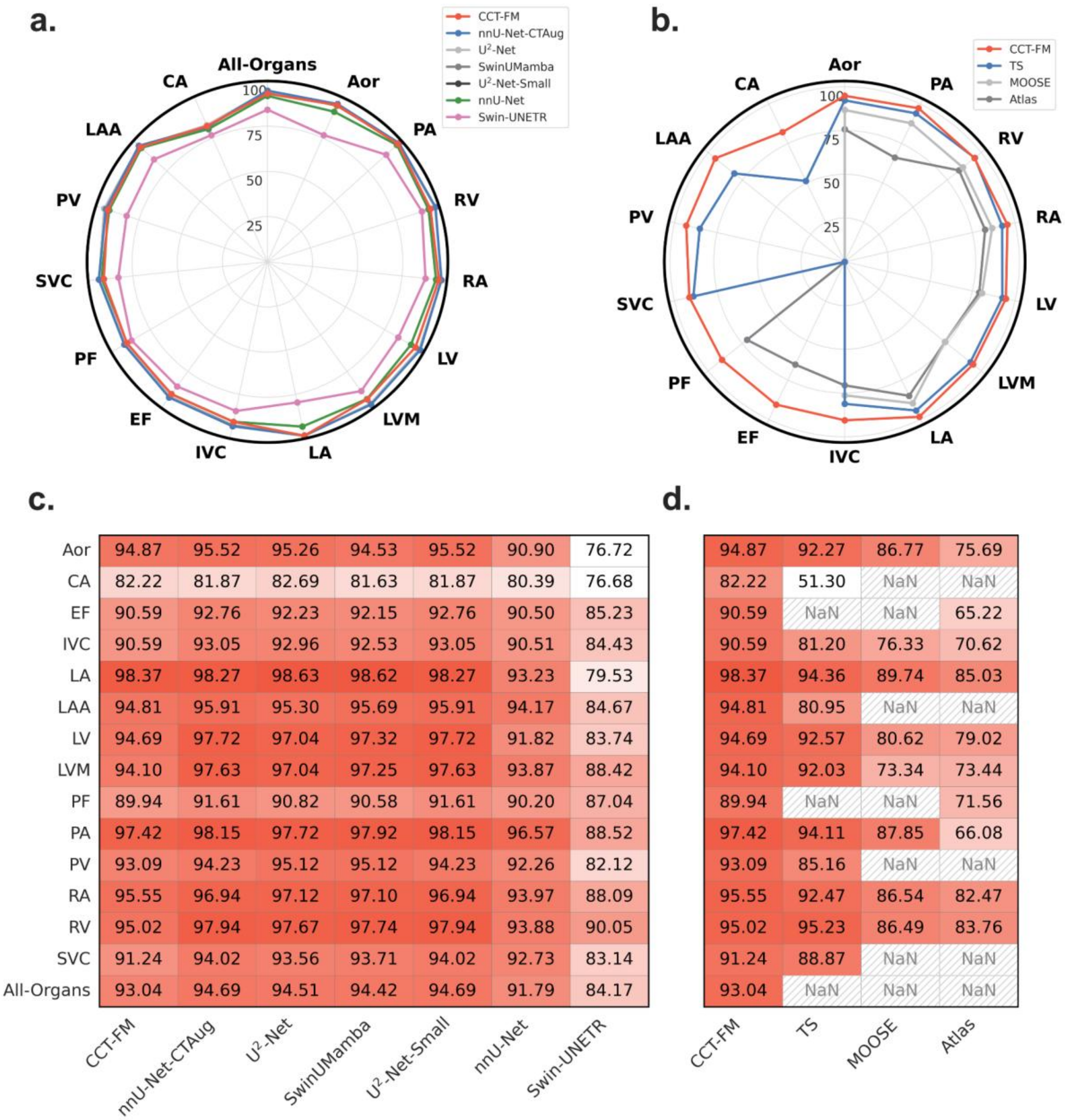

| | CCT-FM | nnU-Net-CTAug | U²-Net | SwinUMamba | U²-Net-Small | nnU-Net | Swin-UNETR |
|---|---|---|---|---|---|---|---|
| Aor | 94.87 | 95.52 | 95.26 | 94.53 | 95.52 | 90.90 | 76.72 |
| CA | 82.22 | 81.87 | 82.69 | 81.63 | 81.87 | 80.39 | 76.68 |
| EF | 90.59 | 92.76 | 92.23 | 92.15 | 92.76 | 90.50 | 85.23 |
| IVC | 90.59 | 93.05 | 92.96 | 92.53 | 93.05 | 90.51 | 84.43 |
| LA | 98.37 | 98.27 | 98.63 | 98.62 | 98.27 | 93.23 | 79.53 |
| LAA | 94.81 | 95.91 | 95.30 | 95.69 | 95.91 | 94.17 | 84.67 |
| LV | 94.69 | 97.72 | 97.04 | 97.32 | 97.72 | 91.82 | 83.74 |
| LVM | 94.10 | 97.63 | 97.04 | 97.25 | 97.63 | 93.87 | 88.42 |
| PF | 89.94 | 91.61 | 90.82 | 90.58 | 91.61 | 90.20 | 87.04 |
| PA | 97.42 | 98.15 | 97.72 | 97.92 | 98.15 | 96.57 | 88.52 |
| PV | 93.09 | 94.23 | 95.12 | 95.12 | 94.23 | 92.26 | 82.12 |
| RA | 95.55 | 96.94 | 97.12 | 97.10 | 96.94 | 93.97 | 88.09 |
| RV | 95.02 | 97.94 | 97.67 | 97.74 | 97.94 | 93.88 | 90.05 |
| SVC | 91.24 | 94.02 | 93.56 | 93.71 | 94.02 | 92.73 | 83.14 |
| All-Organs | 93.04 | 94.69 | 94.51 | 94.42 | 94.69 | 91.79 | 84.17 |

| | CCT-FM | TS | MOOSE | Atlas |
|---|---|---|---|---|
| Aor | 94.87 | 92.27 | 86.77 | 75.69 |
| CA | 82.22 | 51.30 | NaN | NaN |
| EF | 90.59 | NaN | NaN | 65.22 |
| IVC | 90.59 | 81.20 | 76.33 | 70.62 |
| LA | 98.37 | 94.36 | 89.74 | 85.03 |
| LAA | 94.81 | 80.95 | NaN | NaN |
| LV | 94.69 | 92.57 | 80.62 | 79.02 |
| LVM | 94.10 | 92.03 | 73.34 | 73.44 |
| PF | 89.94 | NaN | NaN | 71.56 |
| PA | 97.42 | 94.11 | 87.85 | 66.08 |
| PV | 93.09 | 85.16 | NaN | NaN |
| RA | 95.55 | 92.47 | 86.54 | 82.47 |
| RV | 95.02 | 95.23 | 86.49 | 83.76 |
| SVC | 91.24 | 88.87 | NaN | NaN |
| All-Organs | 93.04 | NaN | NaN | NaN |

**Supplementary Figure S32. Whole-heart segmentation benchmark across in-house models and open-source tools across male patients in the ExtTest-4 dataset.** (a) Per-structure Dice coefficients for the seven in-house models (CCT-FM, nnU-Net, nnU-Net-CTAug, SwinUMamba, Swin-UNETR, U²-Net, and U²-Net-Small) across all fourteen cardiac structures, where All-Organs denotes the mean across structures. (b) Per-structure Dice coefficients comparing CCT-FM with the three open-source tools, restricted to the structures each tool produces. (c) Heatmap of mean Dice for every in-house model and structure, with the All-Organs row giving the across-structure mean. (d) Heatmap of mean Dice for CCT-FM against the three open-source tools, where hatched cells mark structures a tool does not produce and that are therefore excluded from its aggregate. All values are mean Dice computed across the external test datasets at the full labeled training set. Structure abbreviations: Aor, aorta; CA, coronary arteries; EF, epicardial fat; IVC, inferior vena cava; LA, left atrium; LAA, left atrial appendage; LV, left ventricle; LVM, left ventricular myocardium; PA, pulmonary arteries; PF, pericardial fat; PV, pulmonary vein; RA, right atrium; RV, right ventricle; SVC, superior vena cava.

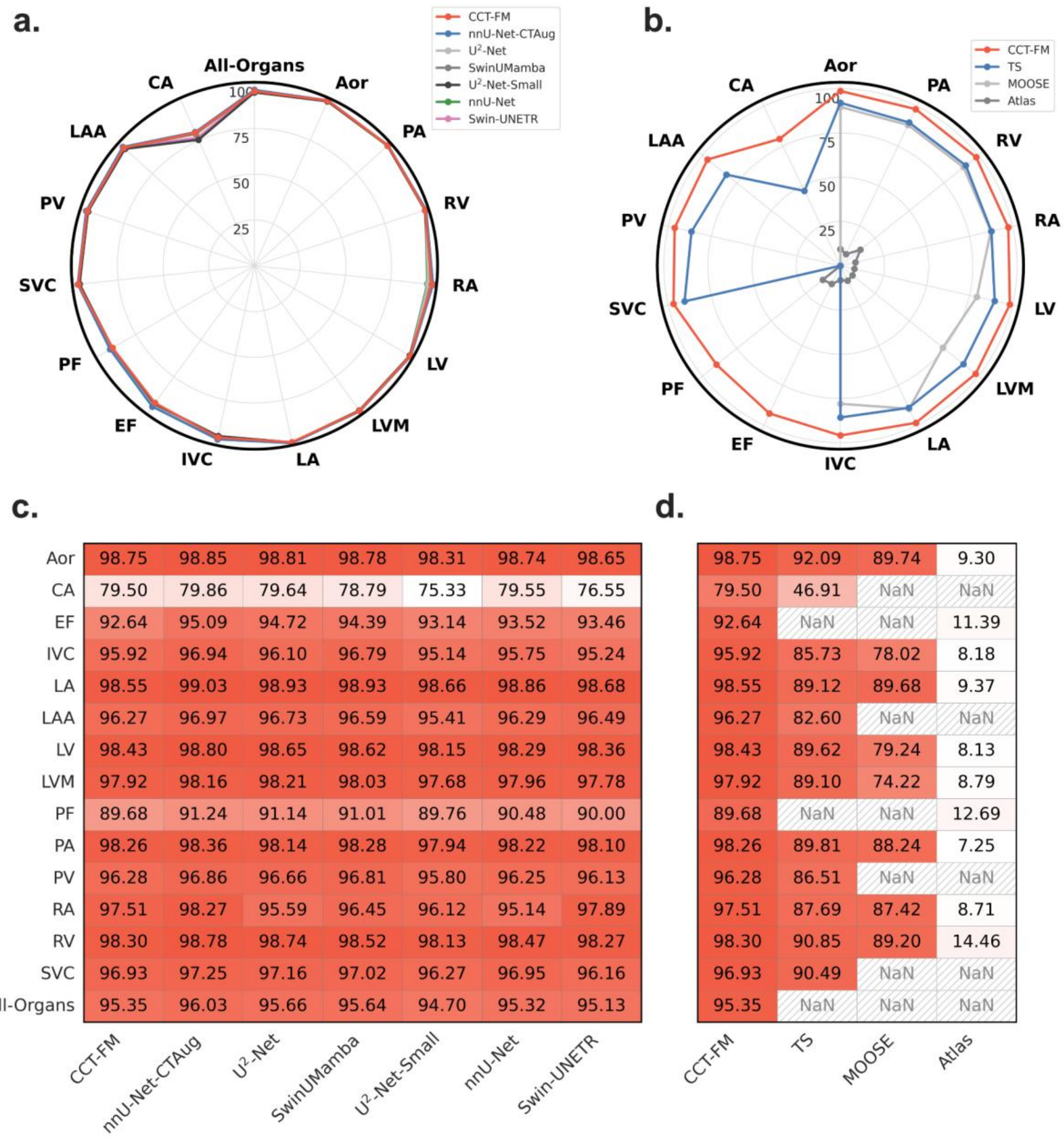

| | CCT-FM | nnU-Net-CTAug | U²-Net | SwinUMamba | U²-Net-Small | nnU-Net | Swin-UNETR |
|---|---|---|---|---|---|---|---|
| Aor | 98.75 | 98.85 | 98.81 | 98.78 | 98.31 | 98.74 | 98.65 |
| CA | 79.50 | 79.86 | 79.64 | 78.79 | 75.33 | 79.55 | 76.55 |
| EF | 92.64 | 95.09 | 94.72 | 94.39 | 93.14 | 93.52 | 93.46 |
| IVC | 95.92 | 96.94 | 96.10 | 96.79 | 95.14 | 95.75 | 95.24 |
| LA | 98.55 | 99.03 | 98.93 | 98.93 | 98.66 | 98.86 | 98.68 |
| LAA | 96.27 | 96.97 | 96.73 | 96.59 | 95.41 | 96.29 | 96.49 |
| LV | 98.43 | 98.80 | 98.65 | 98.62 | 98.15 | 98.29 | 98.36 |
| LVM | 97.92 | 98.16 | 98.21 | 98.03 | 97.68 | 97.96 | 97.78 |
| PF | 89.68 | 91.24 | 91.14 | 91.01 | 89.76 | 90.48 | 90.00 |
| PA | 98.26 | 98.36 | 98.14 | 98.28 | 97.94 | 98.22 | 98.10 |
| PV | 96.28 | 96.86 | 96.66 | 96.81 | 95.80 | 96.25 | 96.13 |
| RA | 97.51 | 98.27 | 95.59 | 96.45 | 96.12 | 95.14 | 97.89 |
| RV | 98.30 | 98.78 | 98.74 | 98.52 | 98.13 | 98.47 | 98.27 |
| SVC | 96.93 | 97.25 | 97.16 | 97.02 | 96.27 | 96.95 | 96.16 |
| All-Organs | 95.35 | 96.03 | 95.66 | 95.64 | 94.70 | 95.32 | 95.13 |

| | CCT-FM | TS | MOOSE | Atlas |
|---|---|---|---|---|
| Aor | 98.75 | 92.09 | 89.74 | 9.30 |
| CA | 79.50 | 46.91 | NaN | NaN |
| EF | 92.64 | NaN | NaN | 11.39 |
| IVC | 95.92 | 85.73 | 78.02 | 8.18 |
| LA | 98.55 | 89.12 | 89.68 | 9.37 |
| LAA | 96.27 | 82.60 | NaN | NaN |
| LV | 98.43 | 89.62 | 79.24 | 8.13 |
| LVM | 97.92 | 89.10 | 74.22 | 8.79 |
| PF | 89.68 | NaN | NaN | 12.69 |
| PA | 98.26 | 89.81 | 88.24 | 7.25 |
| PV | 96.28 | 86.51 | NaN | NaN |
| RA | 97.51 | 87.69 | 87.42 | 8.71 |
| RV | 98.30 | 90.85 | 89.20 | 14.46 |
| SVC | 96.93 | 90.49 | NaN | NaN |
| All-Organs | 95.35 | NaN | NaN | NaN |

**Supplementary Figure S33. Whole-heart segmentation benchmark across in-house models and open-source tools across all patients in the ExtTest-5dataset.** (a) Per-structure Dice coefficients for the seven in-house models (CCT-FM, nnU-Net, nnU-Net-CTAug, SwinUMamba, Swin-UNETR, U²-Net, and U²-Net-Small) across all fourteen cardiac structures, where All-Organs denotes the mean across structures. (b) Per-structure Dice coefficients comparing CCT-FM with the three open-source tools, restricted to the structures each tool produces. (c) Heatmap of mean Dice for every in-house model and structure, with the All-Organs row giving the across-structure mean. (d) Heatmap of mean Dice for CCT-FM against the three open-source tools, where hatched cells mark structures a tool does not produce and that are therefore excluded from its aggregate. All values are mean Dice computed across the external test datasets at the full labeled training set. Structure abbreviations: Aor, aorta; CA, coronary arteries; EF, epicardial fat; IVC, inferior vena cava; LA, left atrium; LAA, left atrial appendage; LV, left ventricle; LVM, left ventricular myocardium; PA, pulmonary arteries; PF, pericardial fat; PV, pulmonary vein; RA, right atrium; RV, right ventricle; SVC, superior vena cava.

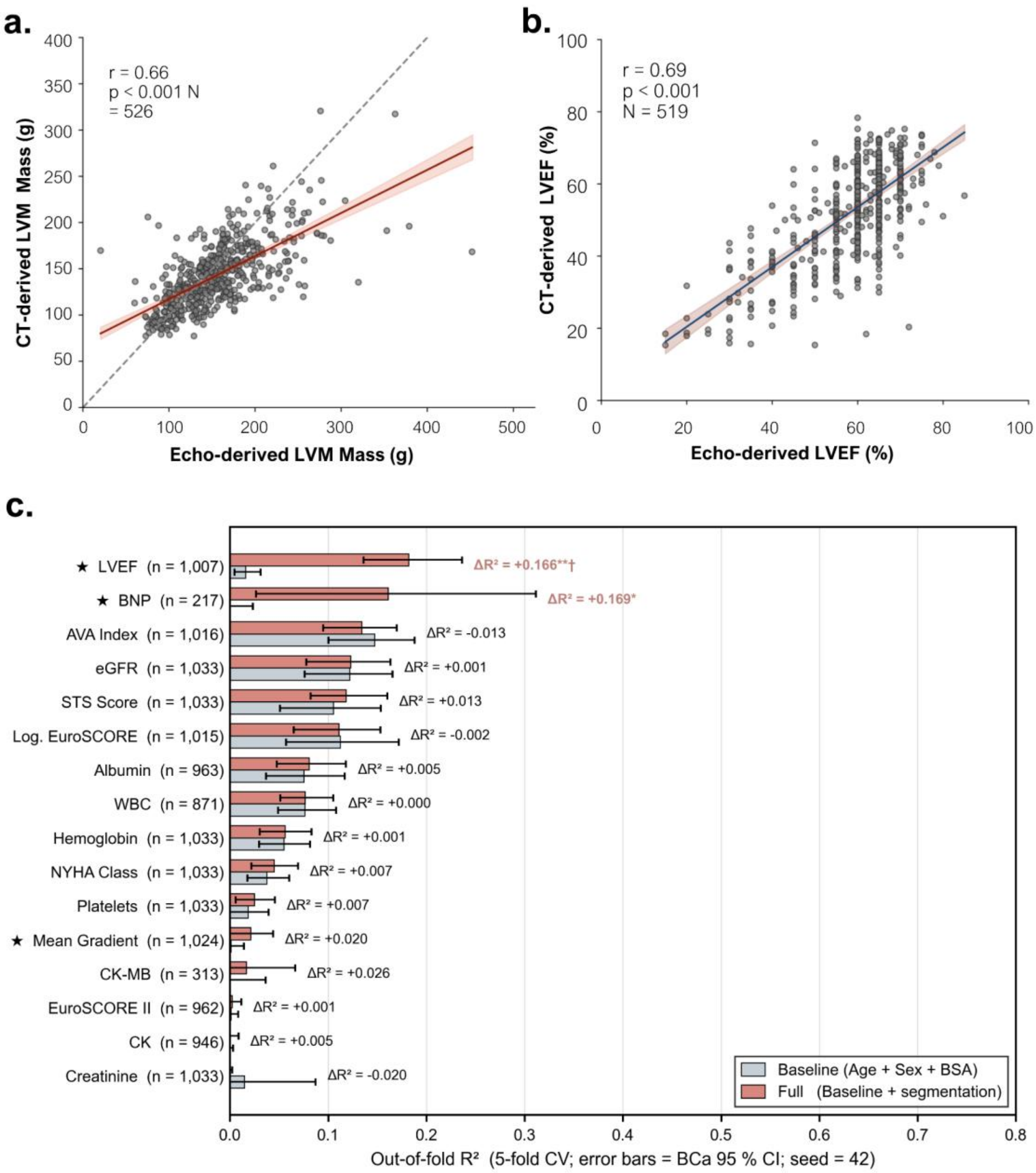

**Supplementary Figure S34. Clinical validation of CT-derived cardiac measurements by concordance with echocardiography (a, b) and association with clinical parameters in female patients.** (a, b) CT-derived left-ventricular myocardial mass (a, N = 526) and LVEF (b, N = 519) plotted against the matched echocardiographic measurement, with OLS fit and shaded 95 % band (Pearson r = 0.70 and 0.72, both $p < 0.001$). (c) Out-of-fold $R^2$ (5-fold cross-validation) of a Baseline CatBoost model (age, sex, body-surface area; grey) and a Full model adding the eight CT-derived substructure volumes (red), ordered by full-model $R^2$. $\Delta R^2$ (Full − Baseline) is annotated beside each pair. Error bars are bias-corrected accelerated (BCa) 95 % confidence intervals for each model's $R^2$ (patient-level bootstrap); the significance of $\Delta R^2$ is assessed from the paired bootstrap of the difference. The false discovery rate (FDR) was controlled at q = 0.05 with the Benjamini–Hochberg procedure, applied separately within the pre-specified primary outcomes (★; LVEF, BNP, mean trans-aortic gradient) and the exploratory family. Asterisks mark targets whose $\Delta R^2$ remains significant after FDR control (* adjusted $p \leq 0.05$, ** $\leq 0.01$); a dagger (†) marks those additionally corroborated by the FDR-adjusted per-fold paired t-test.

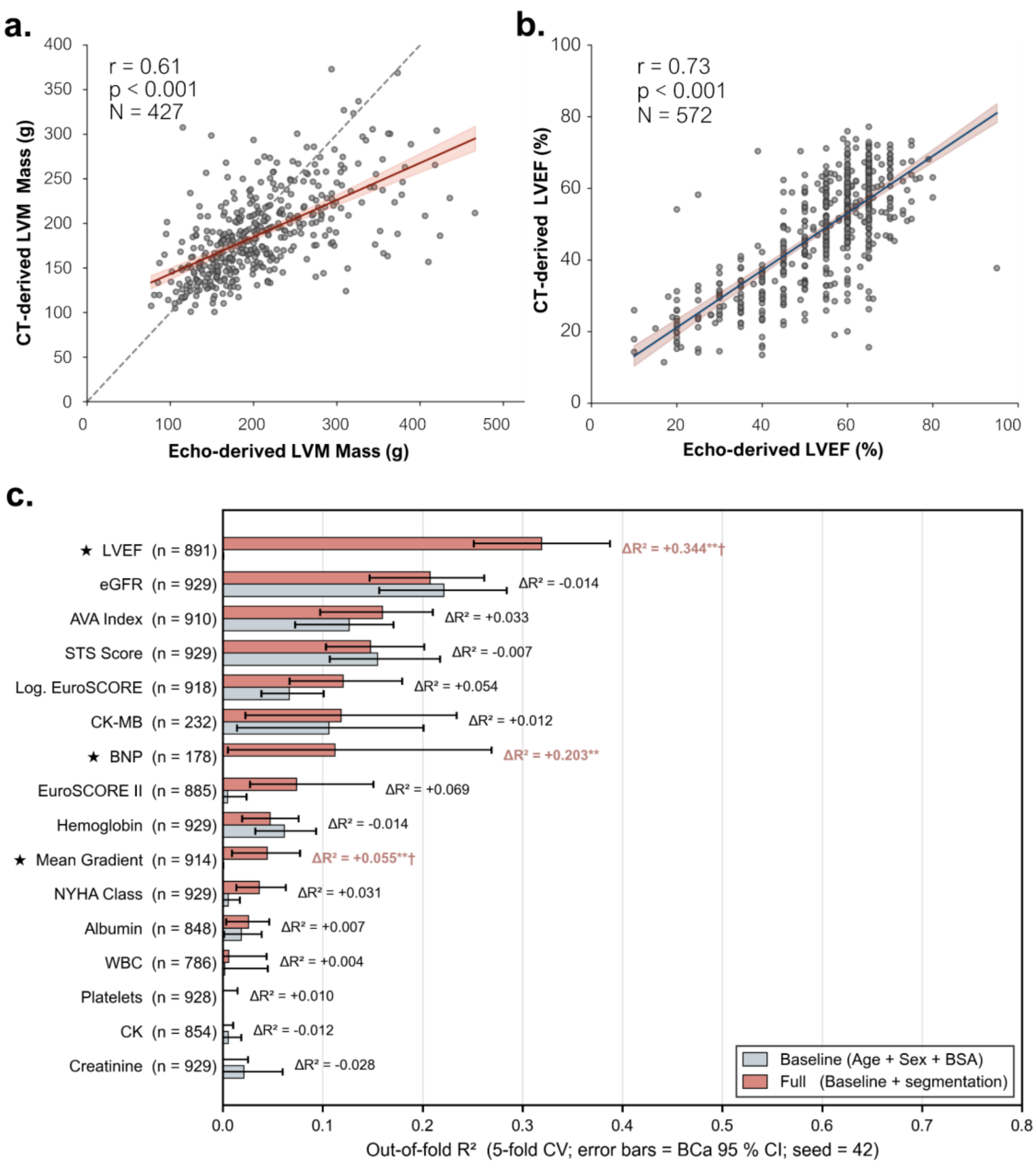

**Supplementary Figure S35. Clinical validation of CT-derived cardiac measurements by concordance with echocardiography (a, b) and association with clinical parameters in male patients.** (a, b) CT-derived left-ventricular myocardial mass (a, N = 427) and LVEF (b, N = 572) plotted against the matched echocardiographic measurement, with OLS fit and shaded 95 % band (Pearson r = 0.70 and 0.72, both $p < 0.001$). (c) Out-of-fold $R^2$ (5-fold cross-validation) of a Baseline CatBoost model (age, sex, body-surface area; grey) and a Full model adding the eight CT-derived substructure volumes (red), ordered by full-model $R^2$. $\Delta R^2$ (Full − Baseline) is annotated beside each pair. Error bars are bias-corrected accelerated (BCa) 95 % confidence intervals for each model's $R^2$ (patient-level bootstrap); the significance of $\Delta R^2$ is assessed from the paired bootstrap of the difference. The false discovery rate (FDR) was controlled at q = 0.05 with the Benjamini–Hochberg procedure, applied separately within the pre-specified primary outcomes (★; LVEF, BNP, mean trans-aortic gradient) and the exploratory family. Asterisks mark targets whose $\Delta R^2$ remains significant after FDR control (* adjusted $p \leq 0.05$, ** $\leq 0.01$); a dagger (†) marks those additionally corroborated by the FDR-adjusted per-fold paired t-test.

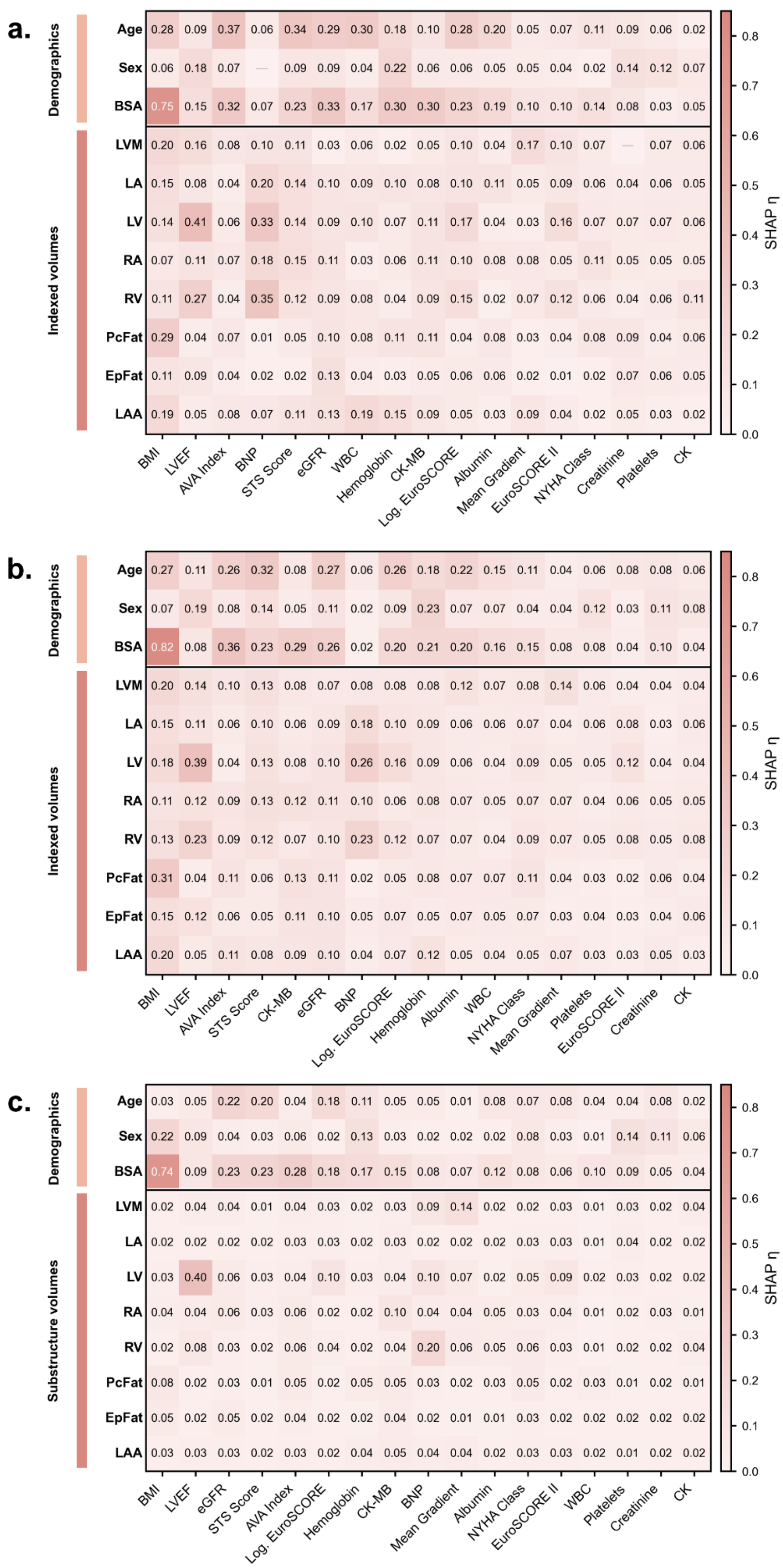

**Supplementary Figure S36 Anatomical attribution of the CT-derived volume signal by SHAP-η, from single predictors without and with demographic adjustment and within the joint model.** (a, b) Pairwise SHAP-η heatmaps. For every predictor-by-target combination a separate five-fold cross-validated CatBoost model was fitted with that predictor as the sole substantive feature, either alone (a, unadjusted) or with age and sex added as covariates (b, adjusted), and the SHAP-η of the predictor is shown. Rows are the three demographic predictors and the eight body-surface-area-indexed substructure volumes, and columns are the 17 targets ordered by their maximum η. (c) Full-model SHAP-η heatmap. Per-feature SHAP-η for the 11-feature Full model (age, sex, body-surface area, and the eight raw substructure volumes), computed patient-wise from CatBoost TreeSHAP on held-out folds, with columns ordered by full-model $R^2$ as in **Supplementary Figure S39** panel a. SHAP-η = $\sqrt{\mathrm{Var}(\varphi)/\mathrm{Var}(y)}$ is a unit-free correlation-ratio analogue bounded in [0, 1] that measures each feature's contribution to the variance of the prediction. The three panels share a unified colour scale running from 0 to the maximum observed η across all three matrices, so cells are directly comparable between panels. Body-surface area on body-mass index dominates the scale at $\eta \approx 0.82$, reflecting the algebraic body-surface-area to body-mass-index relationship rather than a cardiac signal (see **Supplementary Figure S39** panel b). Cells with $\eta < 0.005$ are shown as a dash. The orange row accent denotes demographic predictors and the red accent denotes substructure volumes.

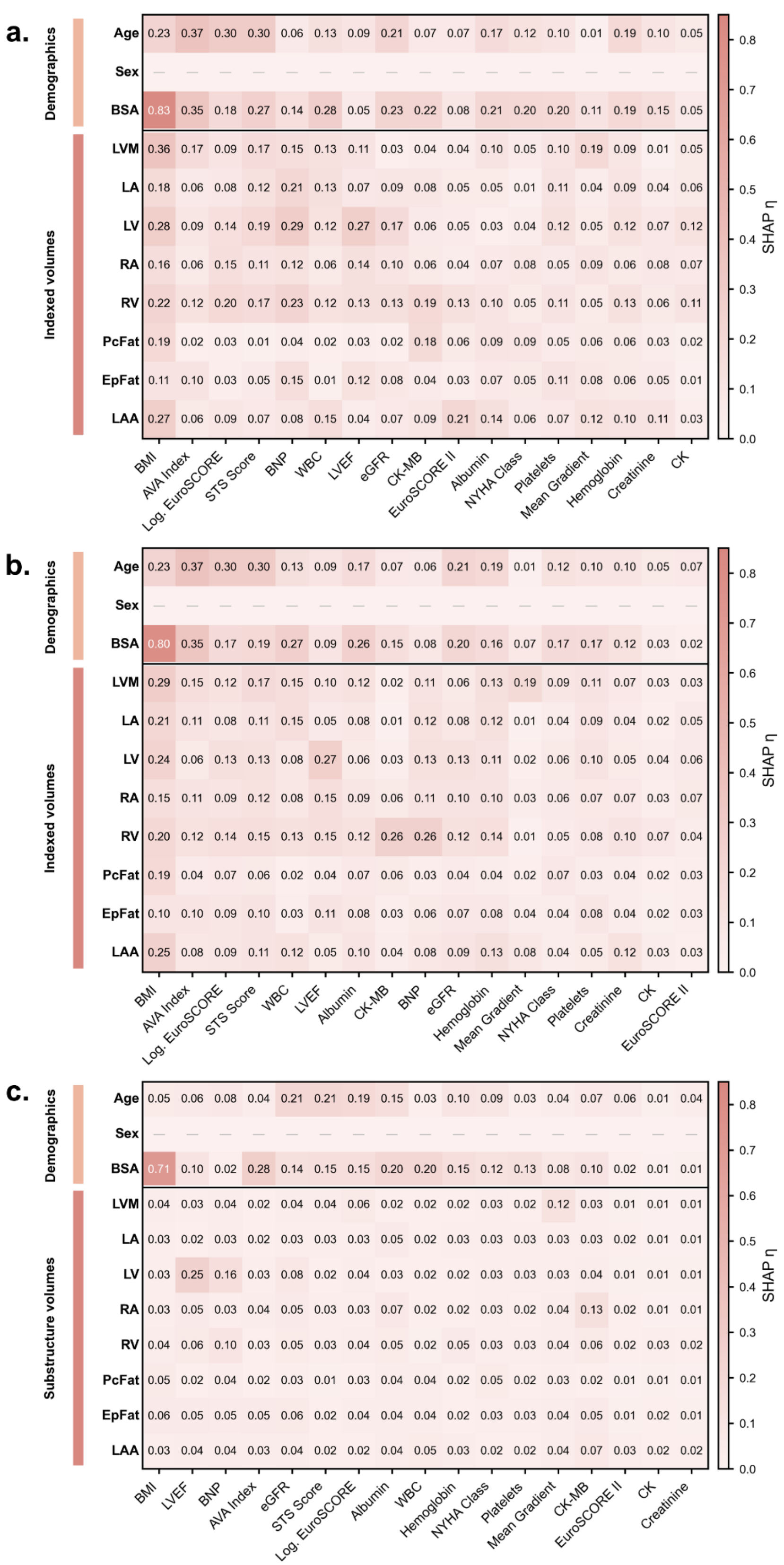

**Supplementary Figure S37. Anatomical attribution of the CT-derived volume signal by SHAP-η, from single predictors without and with demographic adjustment and within the joint model in females.** (a, b) Pairwise SHAP-η heatmaps. For every predictor-by-target combination a separate five-fold cross-validated CatBoost model was fitted with that predictor as the sole substantive feature, either alone (a, unadjusted) or with age and sex added as covariates (b, adjusted), and the SHAP-η of the predictor is shown. Rows are the three demographic predictors and the eight body-surface-area-indexed substructure volumes, and columns are the 17 targets ordered by their maximum η. (c) Full-model SHAP-η heatmap. Per-feature SHAP-η for the 11-feature Full model (age, sex, body-surface area, and the eight raw substructure volumes), computed patient-wise from CatBoost TreeSHAP on held-out folds, with columns ordered by full-model $R^2$ as in **Supplementary Figure S39** panel a. SHAP-η = $\sqrt{(\mathrm{Var}(\varphi)/\mathrm{Var}(y))}$ is a unit-free correlation-ratio analogue bounded in [0, 1] that measures each feature's contribution to the variance of the prediction. The three panels share a unified colour scale running from 0 to the maximum observed η across all three matrices, so cells are directly comparable between panels. Body-surface area on body-mass index dominates the scale at $\eta \approx 0.82$, reflecting the algebraic body-surface-area to body-mass-index relationship rather than a cardiac signal (see **Supplementary Figure S39** panel b). Cells with $\eta < 0.005$ are shown as a dash. The orange row accent denotes demographic predictors and the red accent denotes substructure volumes.

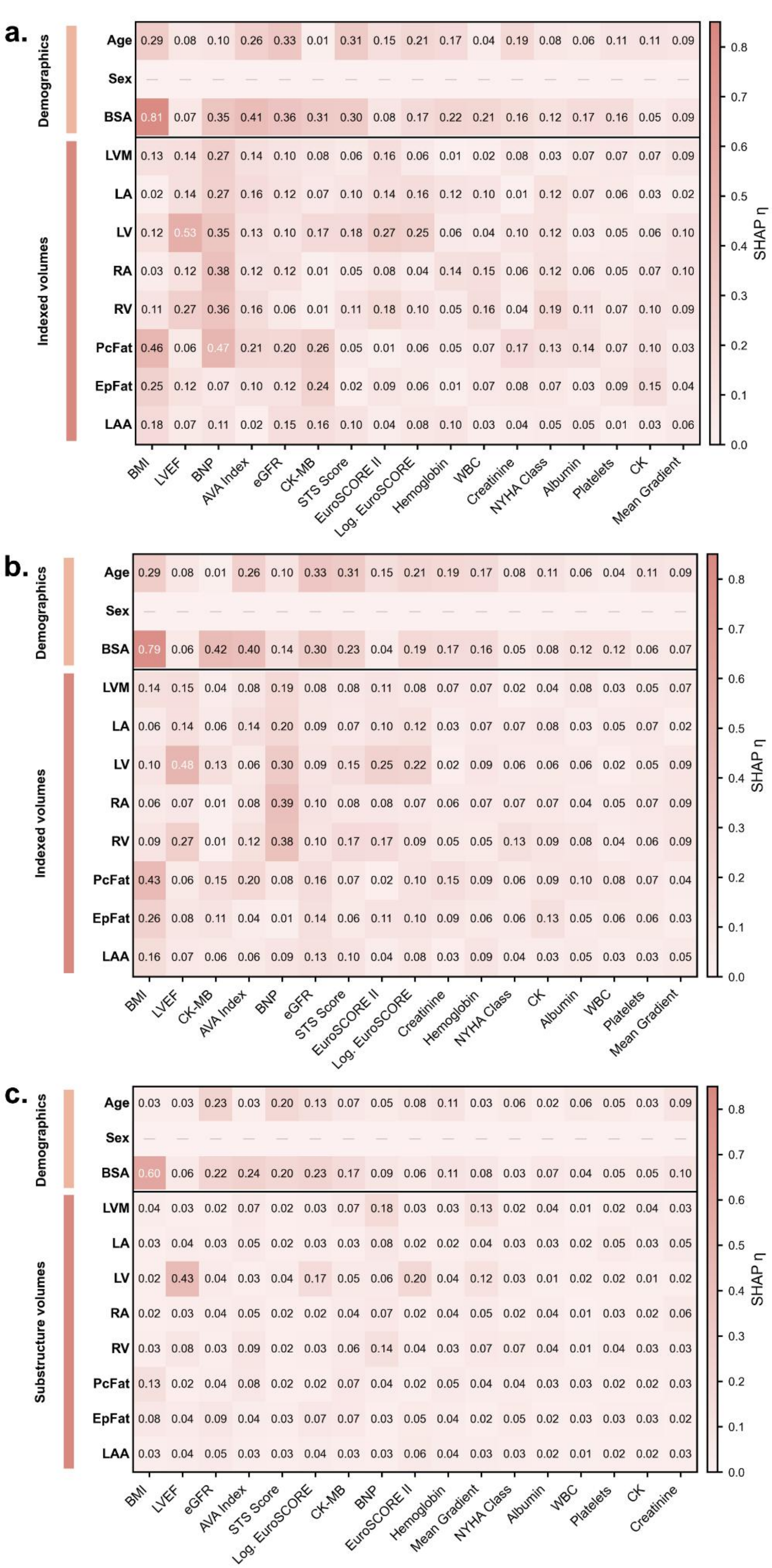

**Supplementary Figure S38 Anatomical attribution of the CT-derived volume signal by SHAP-η, from single predictors without and with demographic adjustment and within the joint model in males.** (a, b) Pairwise SHAP-η heatmaps. For every predictor-by-target combination a separate five-fold cross-validated CatBoost model was fitted with that predictor as the sole substantive feature, either alone (a, unadjusted) or with age and sex added as covariates (b, adjusted), and the SHAP-η of the predictor is shown. Rows are the three demographic predictors and the eight body-surface-area-indexed substructure volumes, and columns are the 17 targets ordered by their maximum η. (c) Full-model SHAP-η heatmap. Per-feature SHAP-η for the 11-feature Full model (age, sex, body-surface area, and the eight raw substructure volumes), computed patient-wise from CatBoost TreeSHAP on held-out folds, with columns ordered by full-model $R^2$ as in **Supplementary Figure S39** panel a. SHAP-η = $\sqrt{(\mathrm{Var}(\varphi)/\mathrm{Var}(y))}$ is a unit-free correlation-ratio analogue bounded in [0, 1] that measures each feature's contribution to the variance of the prediction. The three panels share a unified colour scale running from 0 to the maximum observed η across all three matrices, so cells are directly comparable between panels. Body-surface area on body-mass index dominates the scale at $\eta \approx 0.82$, reflecting the algebraic body-surface-area to body-mass-index relationship rather than a cardiac signal (see **Supplementary Figure S39** panel b). Cells with $\eta < 0.005$ are shown as a dash. The orange row accent denotes demographic predictors and the red accent denotes substructure volumes.

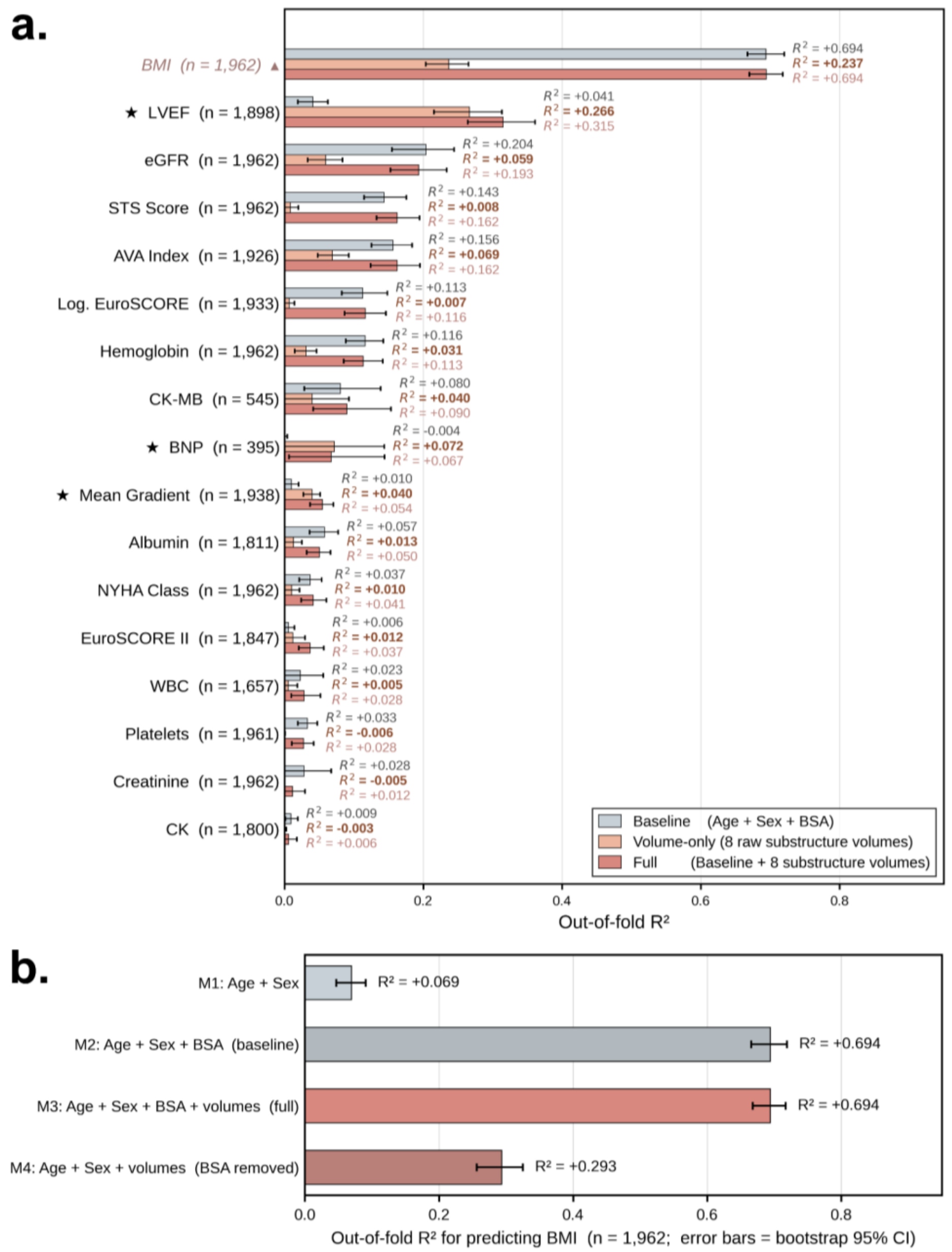

**Supplementary Figure S39. Specificity of the CT-derived volume signal: a three-model decomposition across all targets and a body-mass-index negative control.** (a) Out-of-fold $R^2$ for three nested CatBoost regression models fitted under identical five-fold cross-validation for each of the 17 clinical and laboratory targets, a Baseline model on age, sex, and body-surface area (grey), a Volume-only model on the eight raw CT-derived substructure volumes with no demographic input (orange), and a Full model combining both feature sets (red, 11 features). Targets are ordered by full-model $R^2$. Error bars are 95% bootstrap confidence intervals (bias-corrected and accelerated for the Baseline and Full models, percentile for the Volume-only model, 1,000 resamples each). Stars mark the three pre-specified primary outcomes, and the triangle flags body-mass index, whose baseline $R^2$ reflects the algebraic relationship between body-surface area and body-mass index. Creatine kinase and CK-MB are shown on the log-transformed scale, and inter-chamber ratios are excluded as algebraic transforms of the raw volumes. (b) Out-of-fold $R^2$ for four nested CatBoost models predicting body-mass index, M1 (age and sex), M2 (age, sex, and body-surface area, the Baseline model), M3 (M2 plus the eight volumes, the Full model), and M4 (age, sex, and the eight volumes with body-surface area removed). Sample size is shown in the axis label, and error bars are percentile bootstrap 95% confidence intervals.

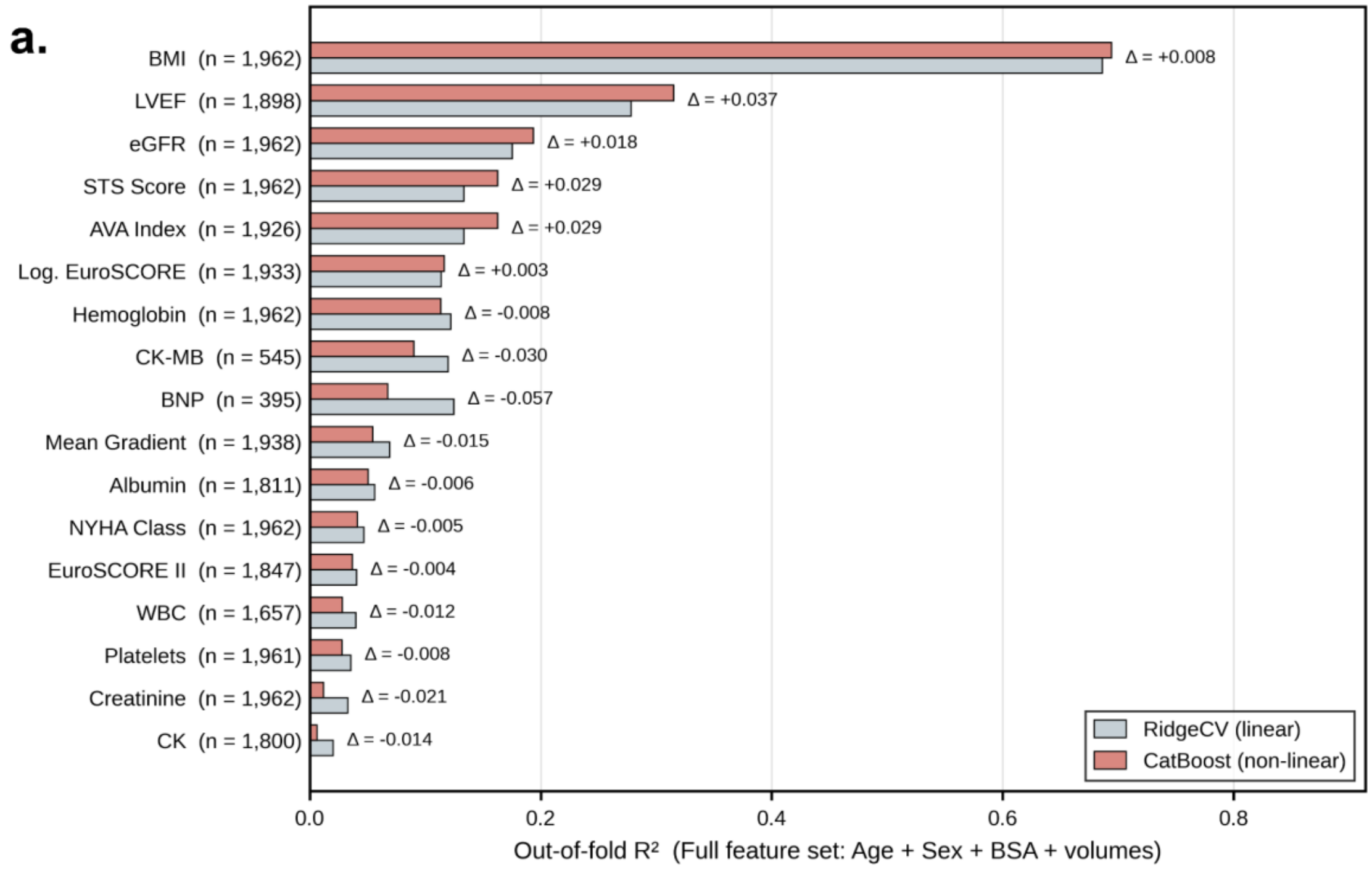

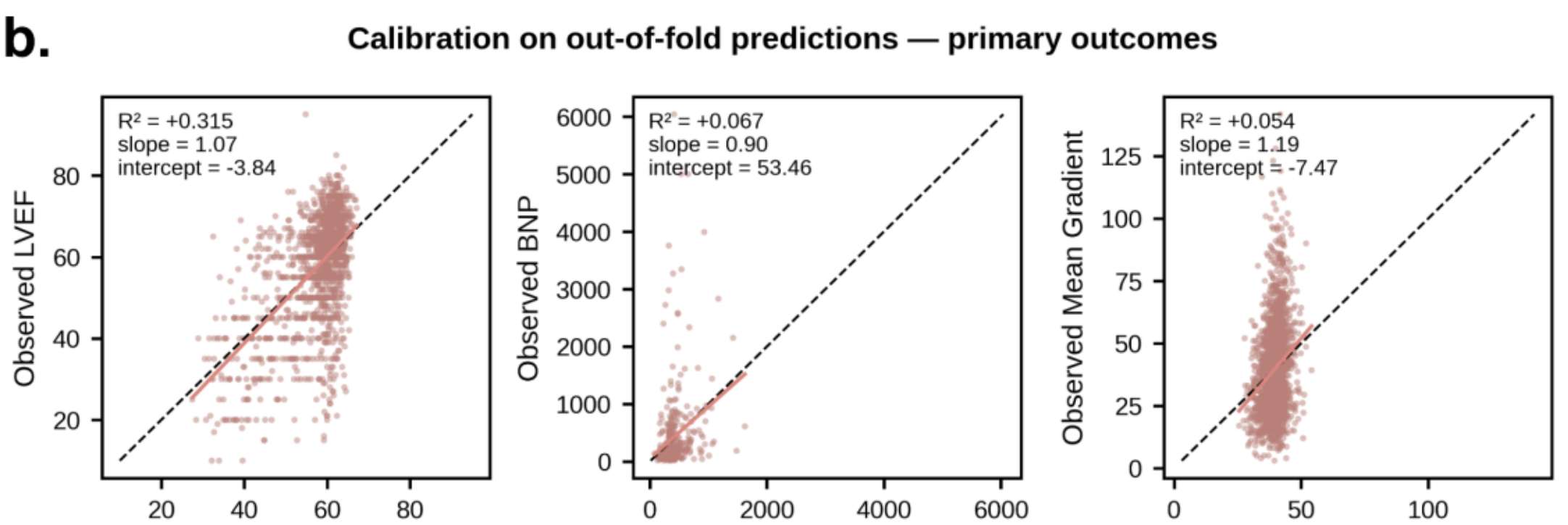

**Supplementary Figure S40. Linearity and calibration checks for the CT-derived volume models, a comparison of CatBoost with a linear model across all targets and the calibration of the primary outcomes.** (a) Out-of-fold $R^2$ for the Full feature set (age, sex, body surface area, and the eight raw substructure volumes, 11 features) fitted under identical five-fold cross-validation with CatBoost (red) and with a linear RidgeCV model (grey), for each of the 17 clinical and laboratory targets ordered by CatBoost $R^2$. The annotation $\Delta = R^2(\text{CatBoost}) - R^2(\text{RidgeCV})$ quantifies the additional non-linear gain, which is modest for every target and largest for left-ventricular ejection fraction (+0.04); the asterisk flags any target whose gain exceeds 0.05, a threshold that no target reaches in this cohort. (b) Calibration of the Full model on the three primary outcomes, left-ventricular ejection fraction, brain natriuretic peptide, and the mean trans-aortic gradient. Observed values are plotted against pooled out-of-fold predictions, the dashed line is the identity, and the red line is the ordinary-least-squares regression of observed on predicted, with the out-of-fold $R^2$, calibration slope, and intercept inset.

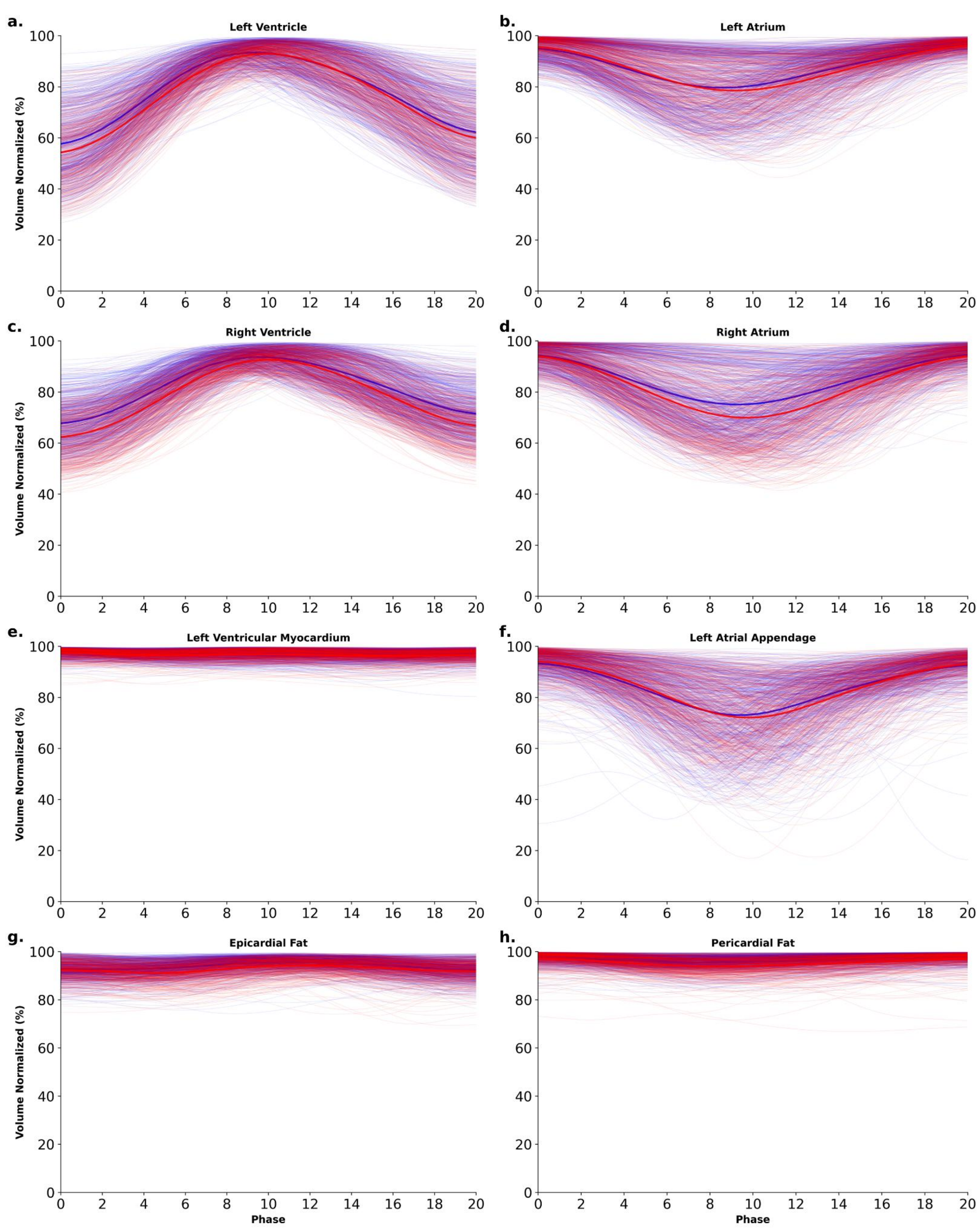

**Supplementary Figure S41. Sex-stratified analysis of phase-resolved volumetric dynamics across the cardiac cycle.** Volume (mL) across cardiac phases for eight cardiac structures (a–h). Faint lines represent individual patients and bold lines the cohort mean, illustrating the characteristic filling and emptying dynamics of each structure recovered automatically across the longitudinal four-dimensional cohort.

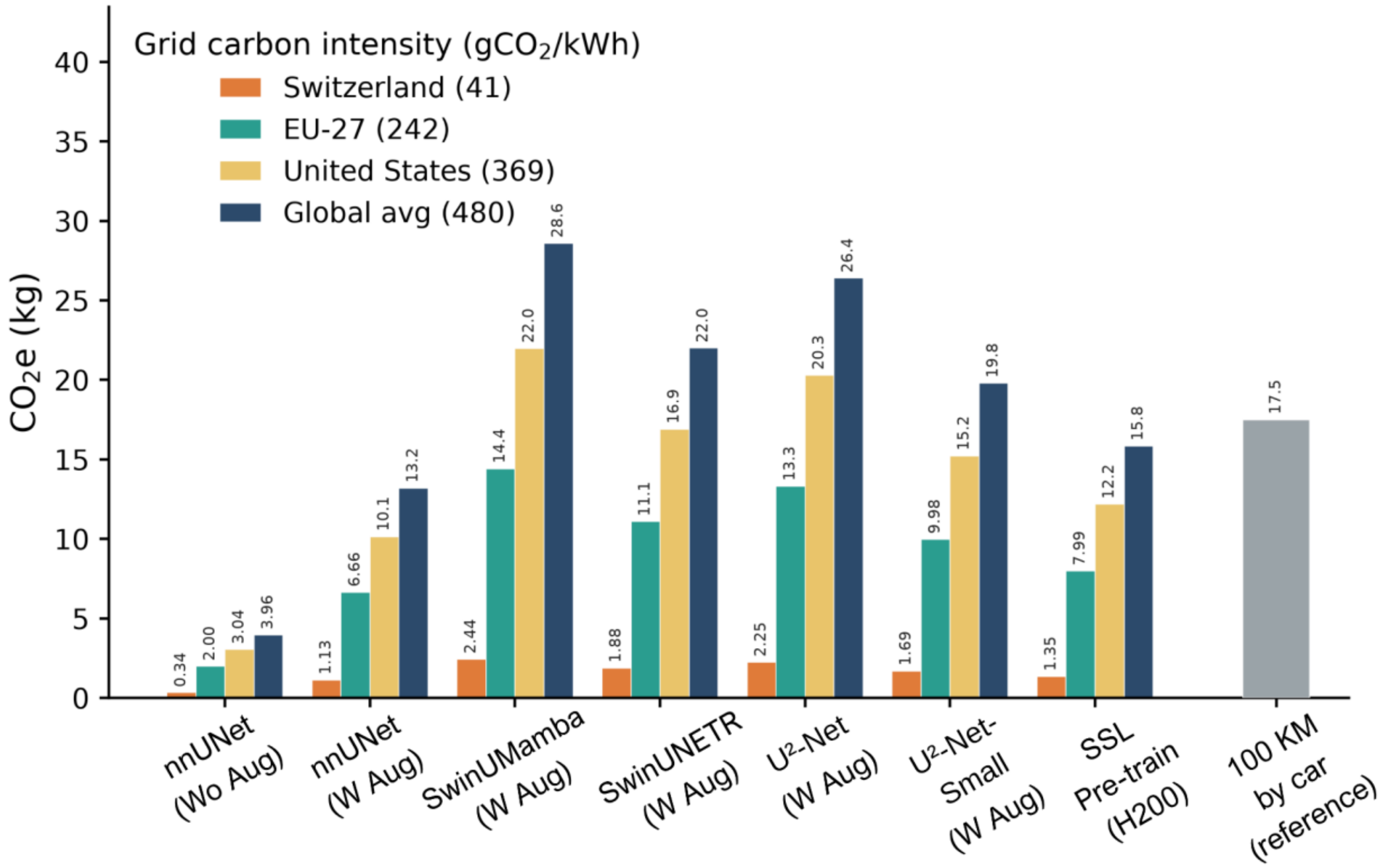

**Supplementary Figure S42.** Estimated carbon footprint of model training across electricity-grid scenarios. Bars show one-time $CO_2$-equivalent emissions for self-supervised pre-training (≈50 GPU-hours on a single NVIDIA H200 NVL 140GB, 600 W) and for training each of the six benchmark segmentation models (1,000 epochs on a single NVIDIA A100 80 GB PCIe, 300 W), estimated with the Green Algorithms method ($CO_2e$ = runtime × n_GPU × TDP × PUE × carbon intensity; PUE = 1.1; GPU operational Scope-2 energy only). Within each task the four bars correspond to grid carbon intensities from a single source (Ember, 2023): Switzerland (41 $gCO_2$/kWh, the grid on which the models were trained), the EU-27 (242), the United States (369), and the global average (480); values in kg $CO_2e$ are printed above each bar. Because emissions scale linearly with grid carbon intensity, absolute footprints rise ≈12-fold from the Swiss grid to the global average while the model ranking is unchanged, with nnU-Net the most efficient and the transformer- and Mamba-based models the most carbon-intensive. A single grey reference bar shows the $CO_2e$ of driving 100 km in an average passenger car (17.5 kg; ≈175 $gCO_2e$/km, Green Algorithms framework, Lannelongue et al. 2021).

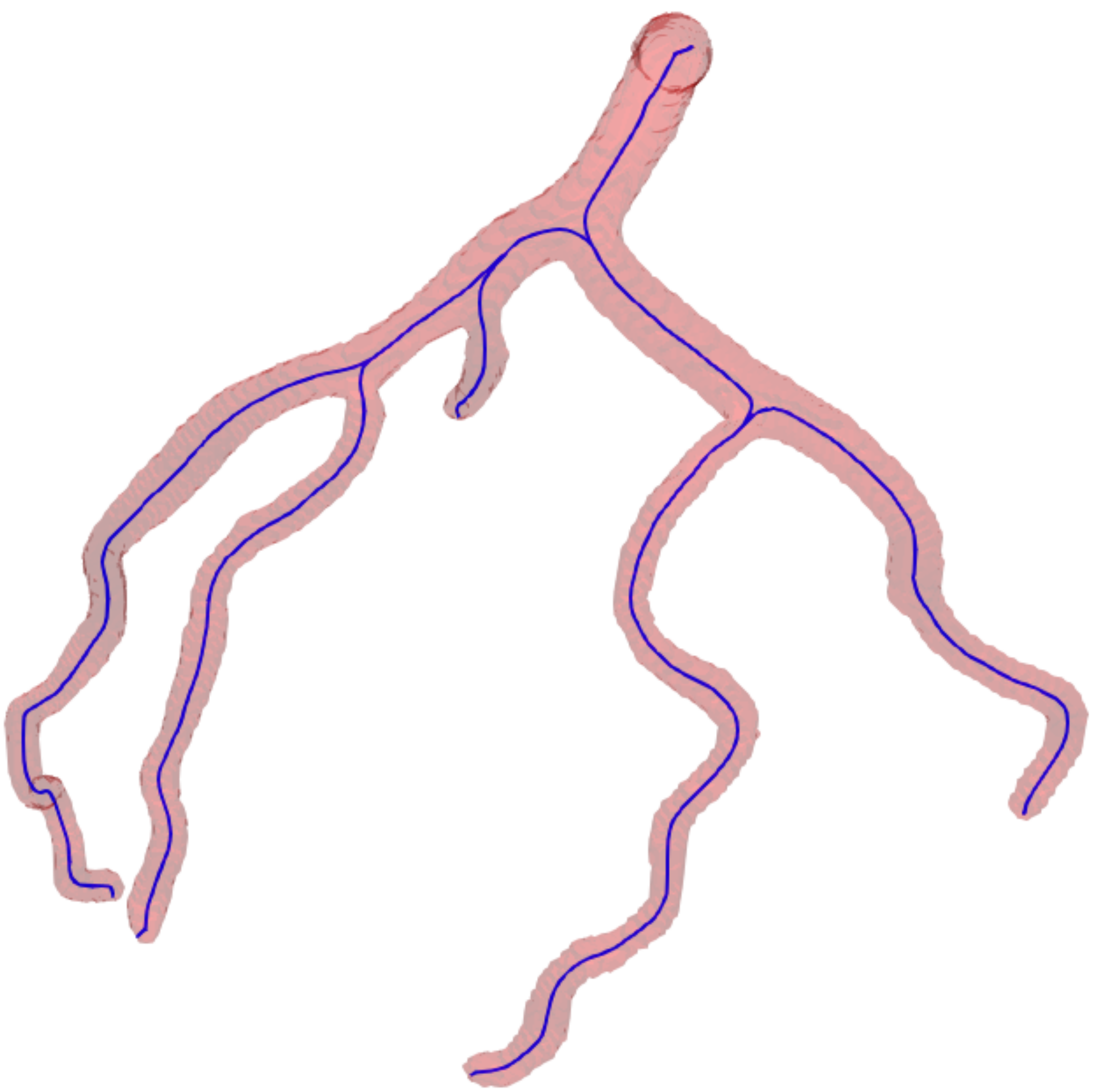

**Supplementary Figure S43. Coronary artery centerline extraction.** Coronary artery centerlines were subsequently extracted from the corrected coronary segmentations, separately for the left and right coronary arteries, using the Vascular Modeling Toolkit (VMTK)[9] and are released with the dataset. The extracted centerlines (blue) are shown rendered inside the semi-transparent coronary segmentation of the left ventricle.

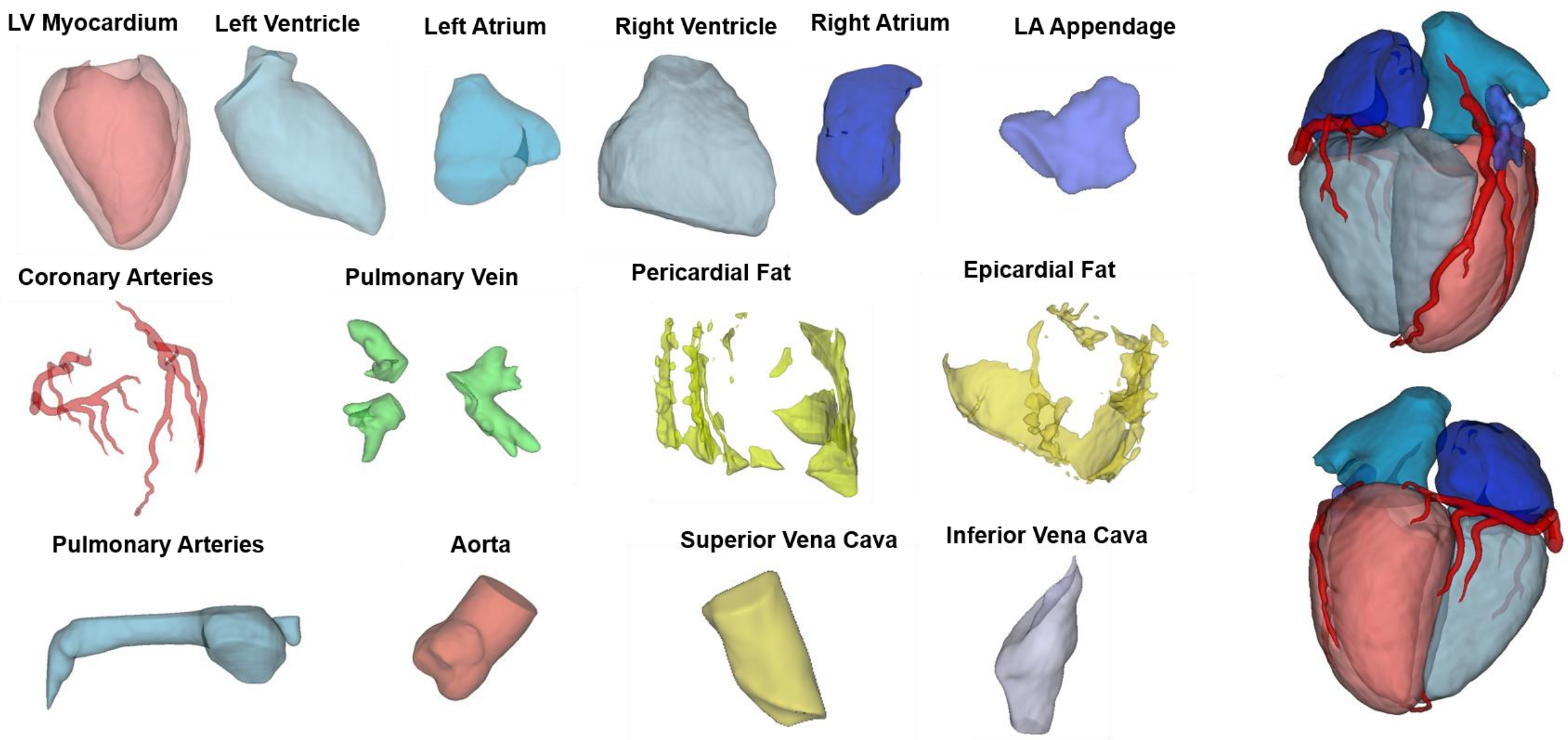

**Supplementary Figure S44. Released three-dimensional STL meshes of the segmented cardiac structures.** Surface renderings of the 14 individually segmented structures, LV myocardium, left ventricle, left atrium, right ventricle, right atrium, LA appendage, coronary arteries, pulmonary vein, pericardial fat, epicardial fat, pulmonary arteries, aorta, superior vena cava, and inferior vena cava, color-coded by anatomical category, alongside two composite whole-heart assemblies (right). A total of 14,280 STL meshes are released with the dataset to support development of shape-aware segmentation networks and downstream applications including cardiovascular simulation, 3D printing, and physical and digital twinning of the heart.

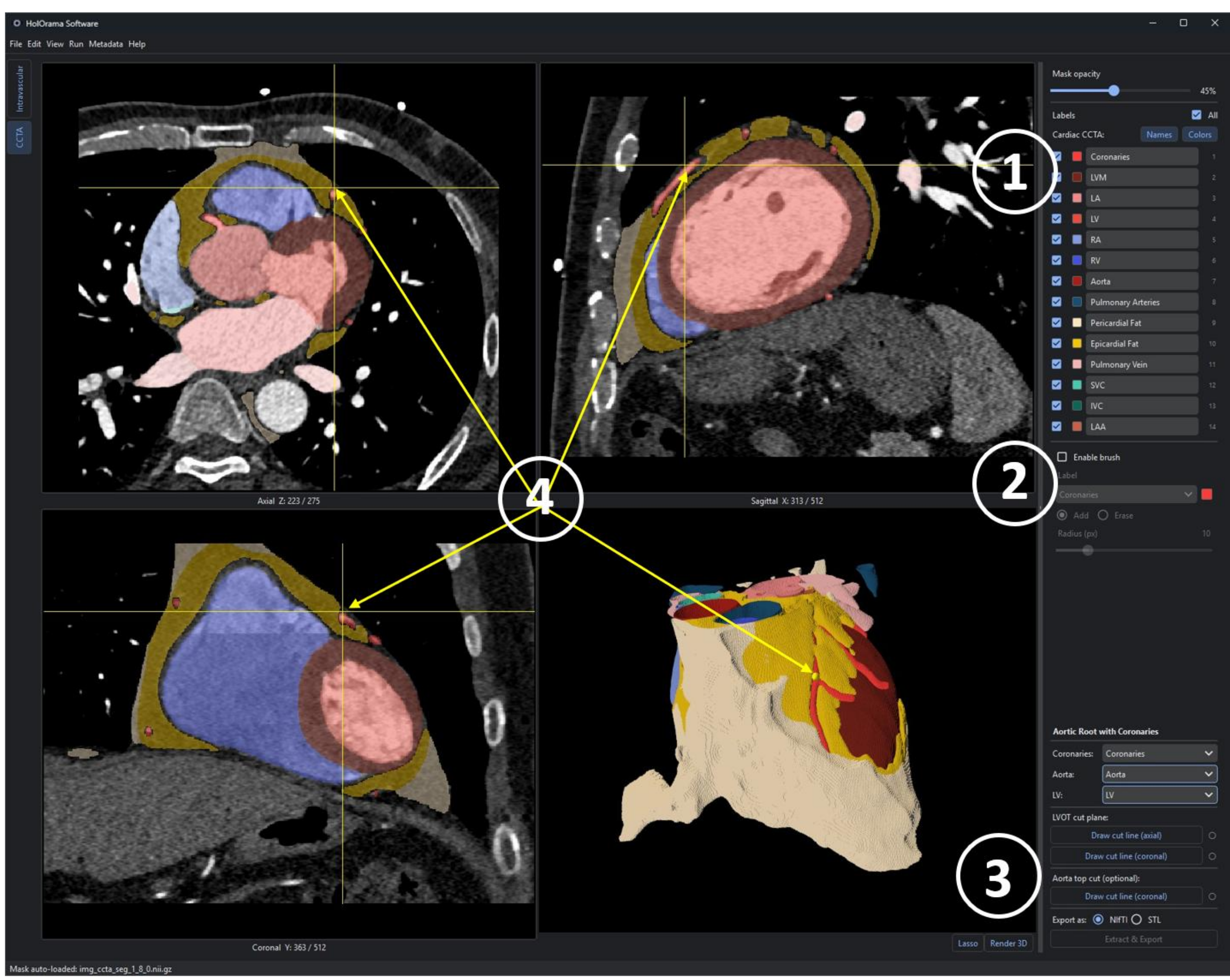

**Supplementary Figure S45.** Open-source software for visualization and use of CCT-FM. 1) Automatically segmented labels with their colors 2) brush tool 3) 3D-render 4) all views have a linked moveable coordinate indicator

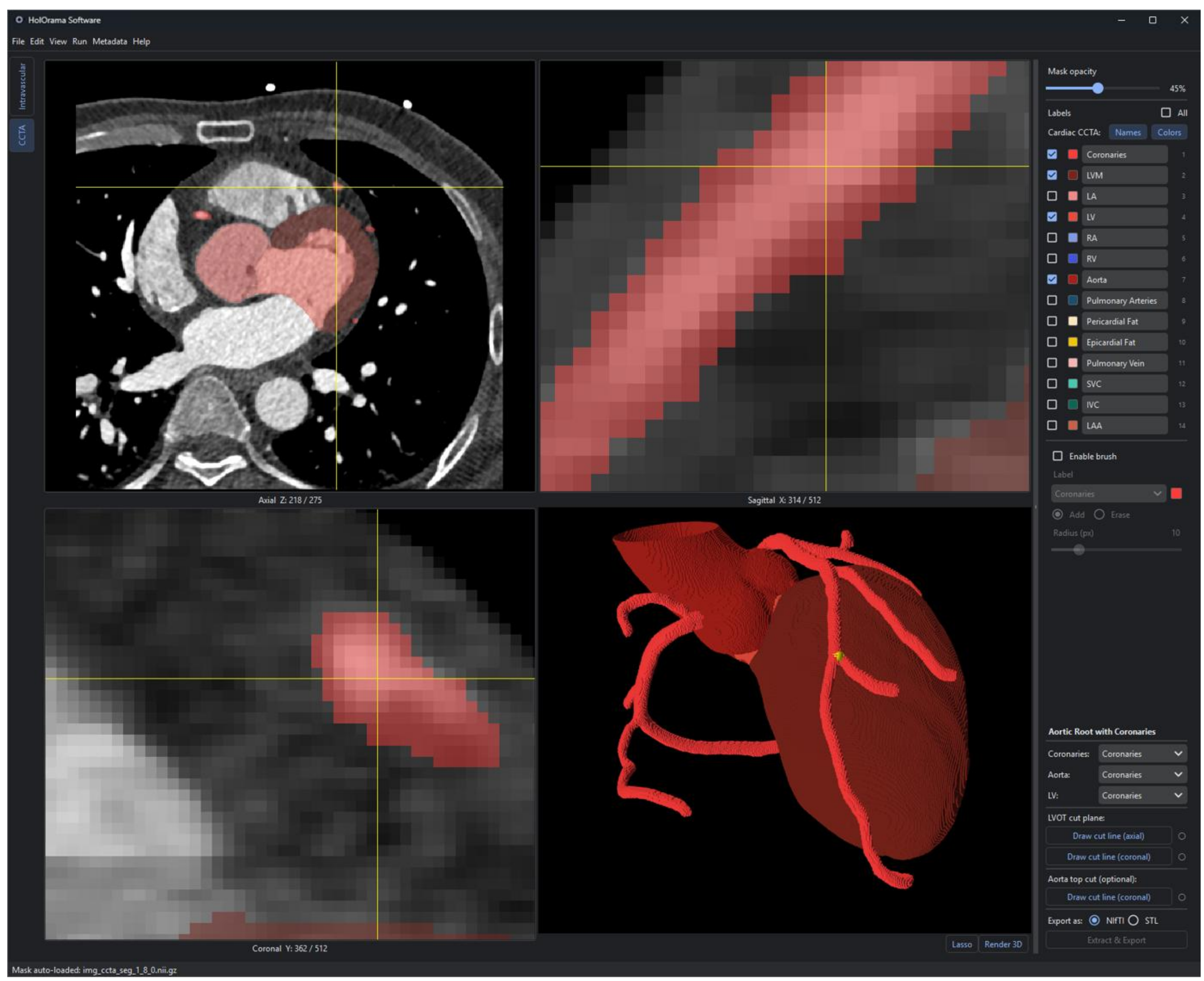

**Supplementary Figure S46.** Open-source software for visualization and use of CCT-FM. Adjust windowing and zoom for every slice individually, display only the contours you are interested in.

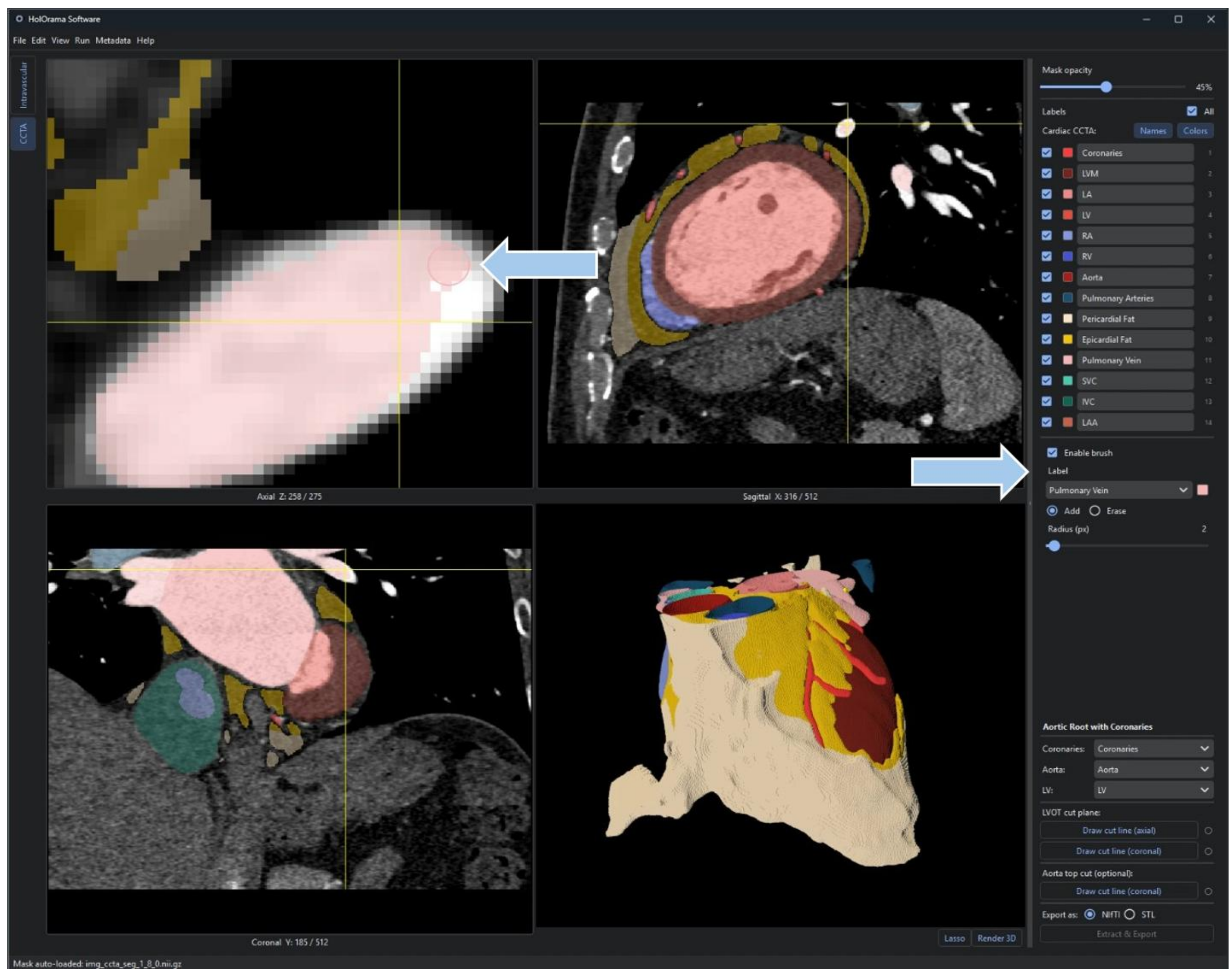

**Supplementary Figure S47.** Open-source software for visualization and use of CCT-FM. Brushtool to add or erase from a contour.

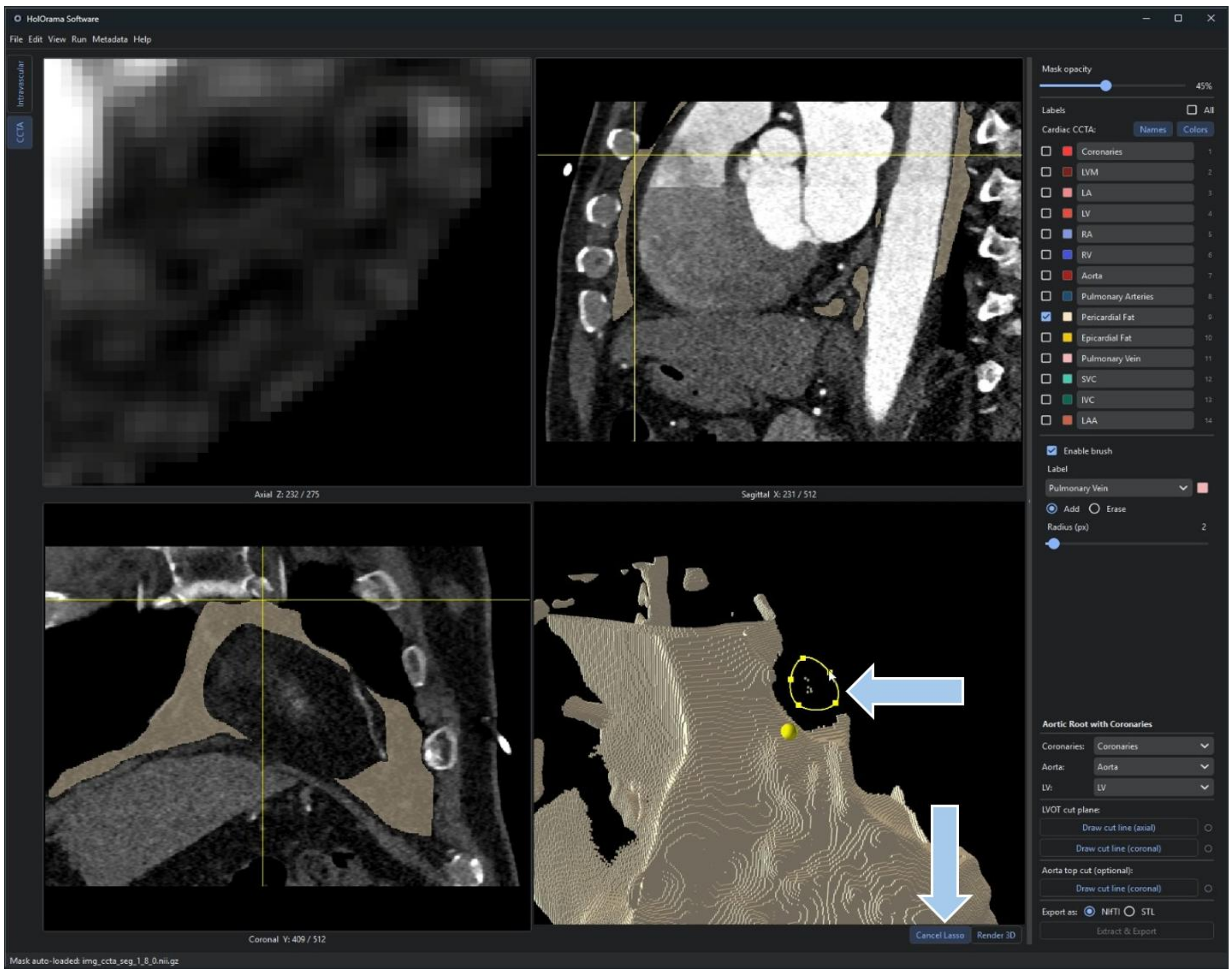

**Supplementary Figure S48.** Open-source software for visualization and use of CCT-FM. Remove outlier points with the lasso tool

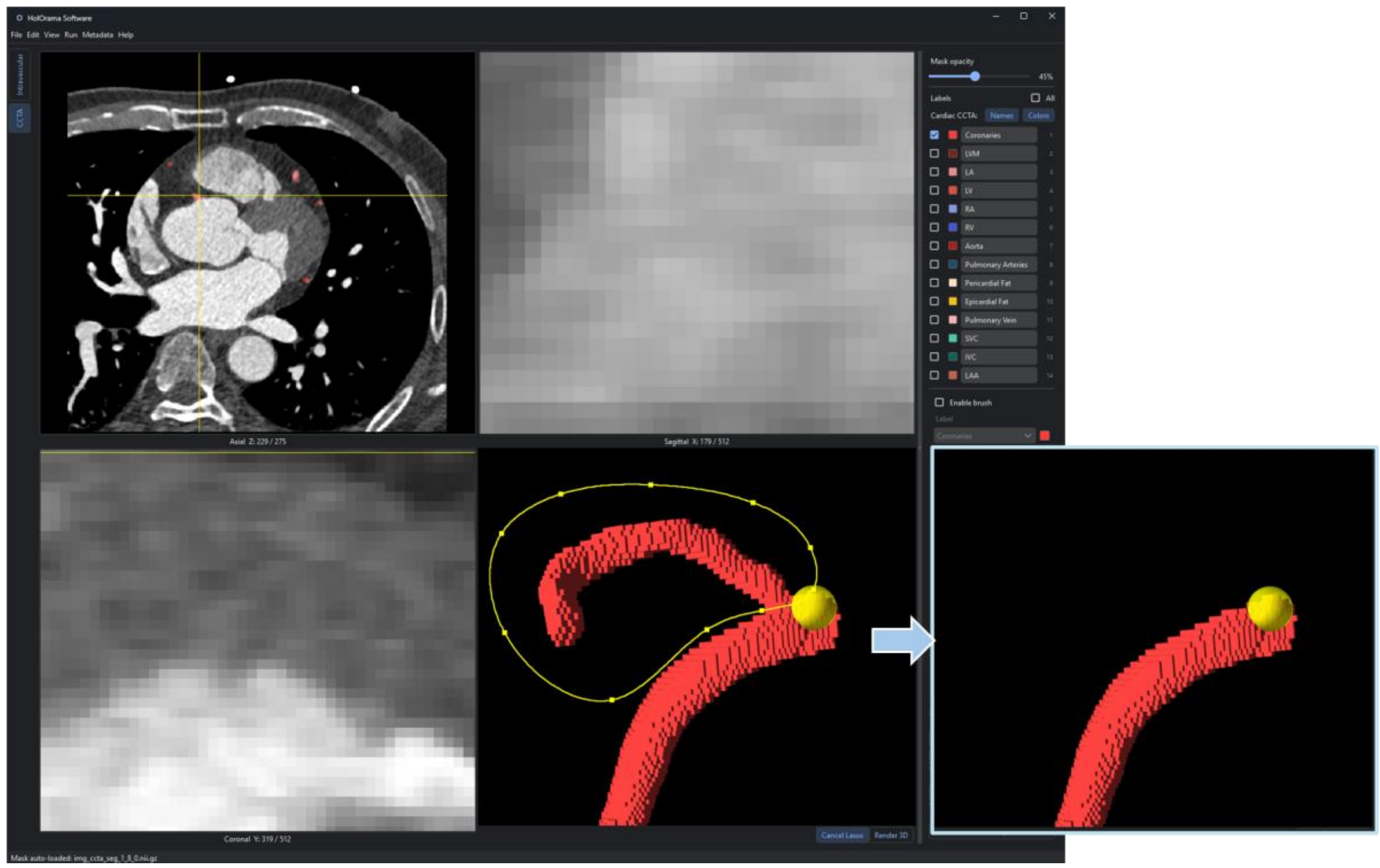

**Supplementary Figure S49.** Open-source software for visualization and use of CCT-FM. Cut away small coronary branches with the lasso tool.

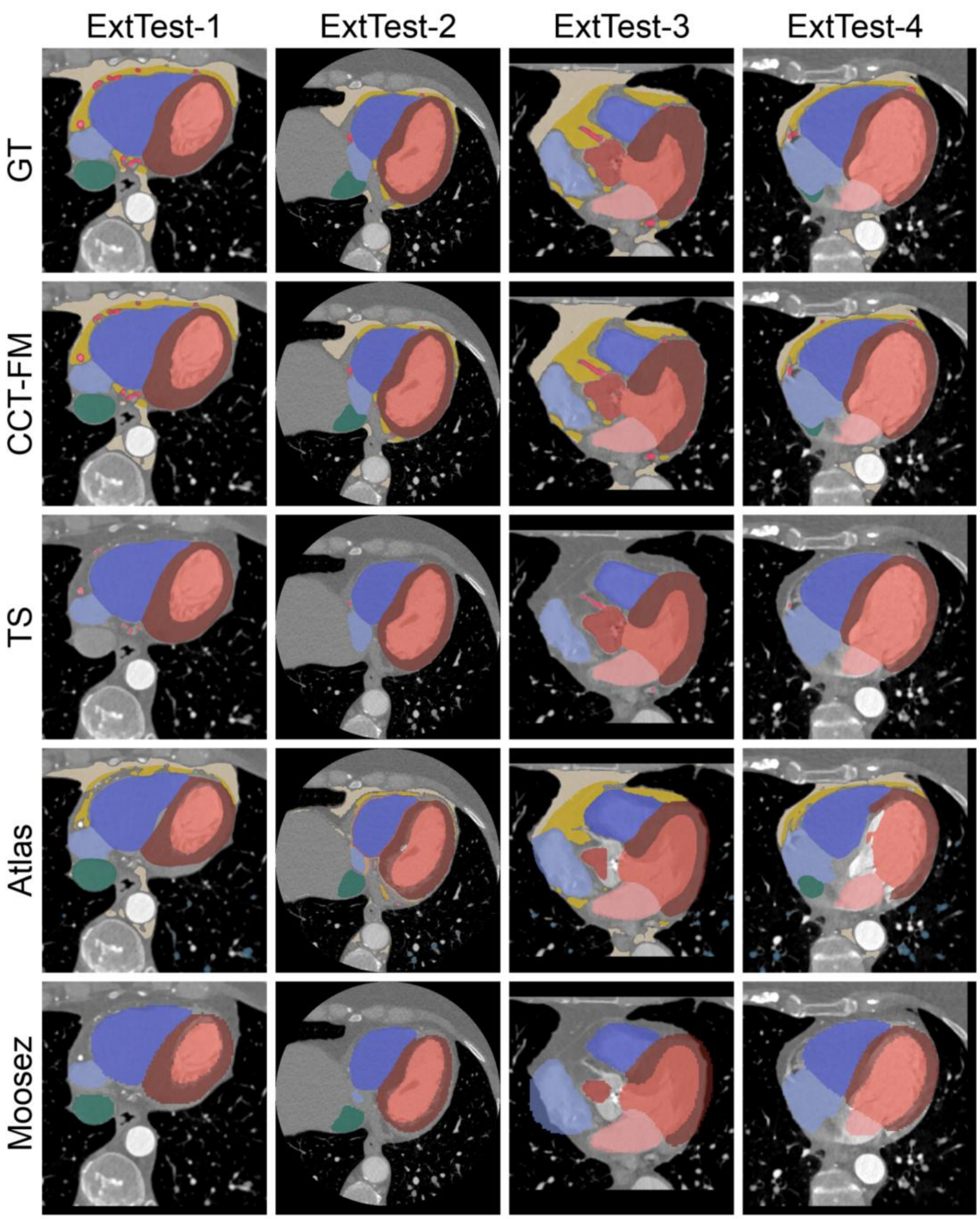

**Supplementary Figure S50 Qualitative comparison of segmentation outputs across models and external test sets.** Representative axial CCT slices from four external test sets (ExtTest-1 to -4; columns) overlaid with segmentations from the ground truth (GT) and each benchmarked model (CCT-FM, TotalSegmentator (TS), Atlas, and MOOSE; rows). Structures are color-coded by anatomical category. Relative to the other models, CCT-FM most closely reproduces the ground truth across structures and test sets, including the smaller structures and fat compartments that TS, Atlas, and MOOSE segment incompletely or omit.